\documentclass{article}

\usepackage[preprint,nonatbib]{neurips_2024}

\usepackage{natbib}
\usepackage{color}
\usepackage[normalem]{ulem}
\usepackage{tabularray}
\usepackage{soul}
\usepackage{pifont}
\usepackage{array}
\usepackage{graphicx}
\usepackage{amsmath}
\usepackage{amsthm}
\usepackage{amssymb}
\usepackage{booktabs}
\usepackage{tabularray}
\usepackage{graphicx}
\usepackage{fancyhdr}
\usepackage{float} 
\usepackage{ulem}
\usepackage{multicol}
\usepackage{multirow}
\usepackage{enumitem}
\usepackage{makecell}
\usepackage{xcolor}
\usepackage{bm}
\usepackage[pagebackref=true,breaklinks=true,letterpaper=true,colorlinks,citecolor=citecolor,bookmarks=false]{hyperref}
\usepackage{textpos}
\usepackage[table]{colortbl}
\definecolor{citecolor}{RGB}{16,78,139}
\DeclareMathOperator*{\argmax}{arg\,max}

\DeclareMathOperator*{\avgpool}{\texttt{Avg}\,\texttt{Pool}}

\usepackage[capitalize]{cleveref}
\crefname{section}{Sec.}{Secs.}
\Crefname{section}{Section}{Sections}
\Crefname{table}{Table}{Tables}
\crefname{table}{Tab.}{Tabs.}

\usepackage{caption}
\usepackage{subcaption}
\expandafter\def\csname ver@subfig.sty\endcsname{}
\usepackage{subfig}
\usepackage{wrapfig}

\title{AlignedCut: Visual Concepts Discovery on Brain-Guided Universal Feature Space}

\author{%
  \textbf{Huzheng Yang} \quad \textbf{{James Gee}*} \quad \textbf{{Jianbo Shi}*}\\
  University of Pennsylvania\\
  *: Equal advising\\
}

\begin{document}

\makeatletter
\DeclareRobustCommand{\pdot}{\mathbin{\mathpalette\pdot@\relax}}
\newcommand{\pdot@}[2]{%
  \ooalign{%
    $\m@th#1\circ$\cr
    \hidewidth$\m@th#1\cdot$\hidewidth\cr
  }%
}
\makeatother

\maketitle

\begin{abstract}
  We study the intriguing connection between visual data, deep networks, and the brain. Our method creates a universal channel alignment by using brain voxel fMRI response prediction as the training objective. We discover that deep networks, trained with different objectives, share common feature channels across various models.   These channels can be clustered into recurring sets, corresponding to distinct brain regions, indicating the formation of visual concepts. Tracing the clusters of channel responses onto the images, we see semantically meaningful object segments emerge, even without any supervised decoder.  Furthermore, the universal feature alignment and the clustering of channels produce a picture and quantification of how visual information is processed through the different network layers, which produces precise comparisons between the networks.

\end{abstract}

\newcommand\blfootnote[1]{\begingroup\renewcommand\thefootnote{}\footnote{#1}\addtocounter{footnote}{-1}\endgroup}

\section{Introduction}
Introducing a novel approach, \citet{yang_brain_2024} has successfully established a method of computing a mapping between the brain and deep-nets, effectively linking two black boxes. The brain fMRI prediction task allows for visualizing information flow from layer to layer, using the brain as an analysis tool.  




\begin{wrapfigure}{r}{0.5\textwidth}
    \centering
    \vspace{-4mm}
    \includegraphics[width=0.5\textwidth]{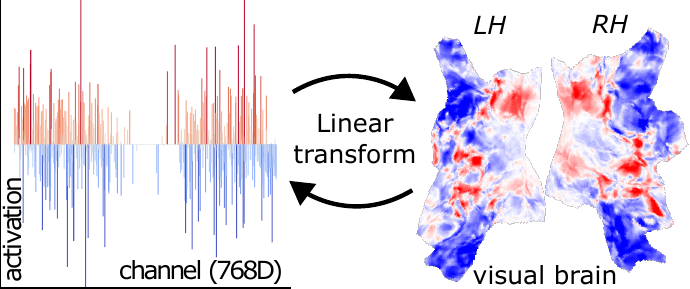}
    \caption{Transform the hidden channel activation of deep-nets into visual brain voxels' response.}
    \label{fig:linear_transform_link_channel_and_brain}
    \vspace{-2mm}
\end{wrapfigure}

If a picture is worth a thousand words, the main idea is that the brain's thousands of voxels can be thought of as alphabets for these words that describe an image. Just as alphabets must be combined to form words and phrases with meanings, we need to find the grouping of brain voxels and their network channel counterparts to understand their meaning (\Cref{fig:linear_transform_link_channel_and_brain}). 

Our main discovery is that while the network layer structure differs, channel feature correspondence exists across networks with a shared encoding of reoccurring visual concepts. This paper builds upon the idea of `Rosetta stone' neurons \citep{dravid_rosetta_2023}, which find channels across networks that share similar image responses in binary segmentation.   If channels are alphabets, `Rosetta stone' provides an alphabet-level translation between networks.

Individual channel-level analysis could miss feature correspondence across networks at finer and coarser levels. On a finer level, because the channels are invariant up to a linear transformation, we might miss a reconstituted feature constructed from a composition of existing channels. On a coarse level, the channels can be combined and clustered to form a bigger `Rosetta' concept.   

To address fine-level channel analysis,  we use brain voxel response as a reference signal and linearly transform channels for each network into a shared space sufficient for brain fMRI prediction. This process produces a universal feature space that aligns channel features across the layers and models.

\begin{figure}[h]
    \vspace{-1mm}
    \centering
    \includegraphics[width=0.9\linewidth]{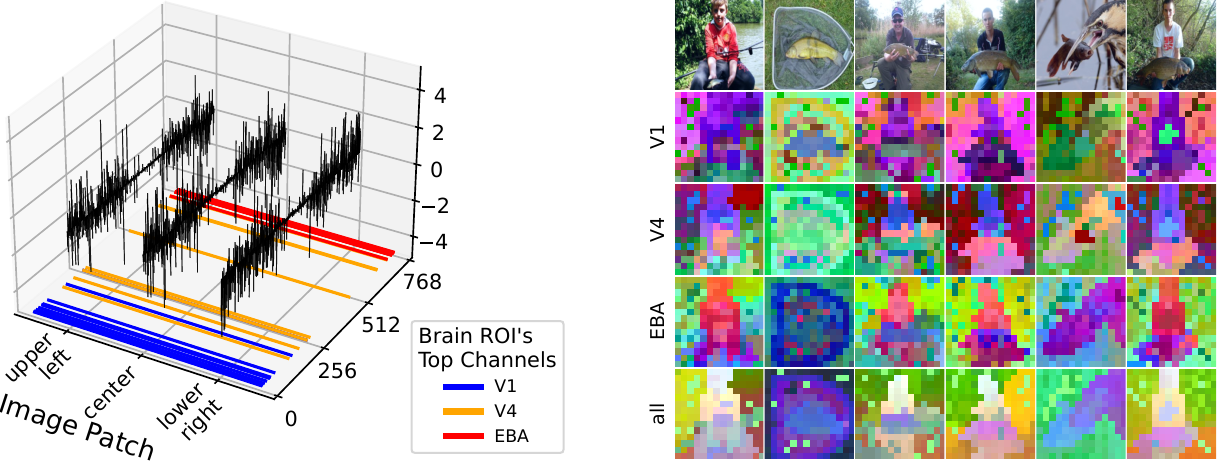}
    \vspace{-1mm}
    \caption{From the 768D feature on CLIP layer-6, we extract different levels of segmentation by restricting the use of a subset of channels. \textbf{\textit{Left:}} Channel activation on example image patches. The ordering of channels is sorted from the early brain to the late brain by their weights for brain voxels. \textbf{\textit{Right:}} Spectral clustering on each subset of channels filtered by each brain ROI (V1, V4, EBA), image pixels colored by 3D spectral-tSNE of top 10 eigenvectors.}
    \label{fig:ch_subset_segmentation}
    \vspace{-2mm}
\end{figure}

To find bigger visual concepts, one can start with Neuroscience knowledge of brain regions (ROIs) with specific brain functionality, i.e., V1, V4, and EBA.    While tracing the mapping of the ROIs to channels can produce visual concepts (\Cref{fig:ch_subset_segmentation}), brain regions don't function in isolation. 

Instead of searching through all possible channel grouping combinations, our first insight is that we can create a channel grouping hypothesis by examining channels from each pixel's perspective. Think of the pixels and channels forming a bipartite graph; each channel produces a per-pixel response (image activation map), defining the graph edge between the pixels and channels. Taking the perspective of pixels, one can collect graph edges incident on each pixel into a vector, which can be thresholded to produce a hypothesis grouping over channels.

Our second insight is that if a channel grouping hypothesis repeats across images, layers, and models, it is highly unlikely to be accidental and, therefore, signals meaningful visual concepts.

We formulate this clustering problem as a graph partition task. The graph nodes are the product space of pixels and layers. We apply spectral clustering to produce k-top eigenvectors.   We take advantage of two properties of spectral clustering: 
it makes 1) soft-cluster embedding space in the form of eigenvectors and 2) hierarchical clustering by varying the number of eigenvectors.

We made the following discoveries. First, shared channel sets, reoccurring across layers and models, predict response in distinct brain regions.   By tracing the channel activation to the known brain ROI properties, we observe that the channel cluster encodes visual concepts at various levels of visual abstraction.   

Second, meaningful object segments can emerge by tracing the channel cluster responses onto each image. We observed that some channel clusters produce figure/ground separation while others produce fine-grained category classification. Our image segmentation requires no additional segmentation decoder and uses only a simple distance measure over the eigenvectors.  


Finally, the universal feature alignment and the spectral clustering of channels produce a picture and quantification of how visual information is processed through the different network layers.


While these discoveries are promising, there are two main technical hurdles to overcome to verify them on a large scale. Our method rests upon a crucial assumption: the channels across the different layers and models can be mapped into a shared space. While brain prediction over thousands of voxels can provide strong guidance for this alignment, an additional constraint would be needed when the shared space has a large dimension (suitable for expressiveness).  We use clustering as a constraint, ensuring alignment linear transformation preserves spectral clustering eigenvectors.
Furthermore, the graph size is enormous as it is a product space over pixels, layers, images, and models; therefore, computing eigenvectors over their pairwise affinity matrix can be computationally infeasible. We developed a Nystrom-like approximation to ensure efficient computation.

In summary, our key contributions are:

1. We constructed a universal channel-aligned space using brain encoding as supervision and spectral clustering eigenvector constraints to ensure minimal channel signal loss. Brain encoding associates the aligned channel space to brain regions and gives them meanings.

2. Models trained with different objectives learned similar visual concepts: corresponding channel patterns exist across different models. The resulting visual concepts can be validated by unsupervised segmentation benchmarks on ImageNet-segmentation and PASCAL VOC.

3. Models show divergent computation paths over the visual concept space formed by the top-k spectral eigenvectors. Different models differ in trajectories and pace of movement layer-to-layer.

\section{Methods: AlignedCut}
\begin{wrapfigure}{r}{0.47\textwidth}
    \centering
    \vspace{-6mm}
    \includegraphics[width=0.47\textwidth]{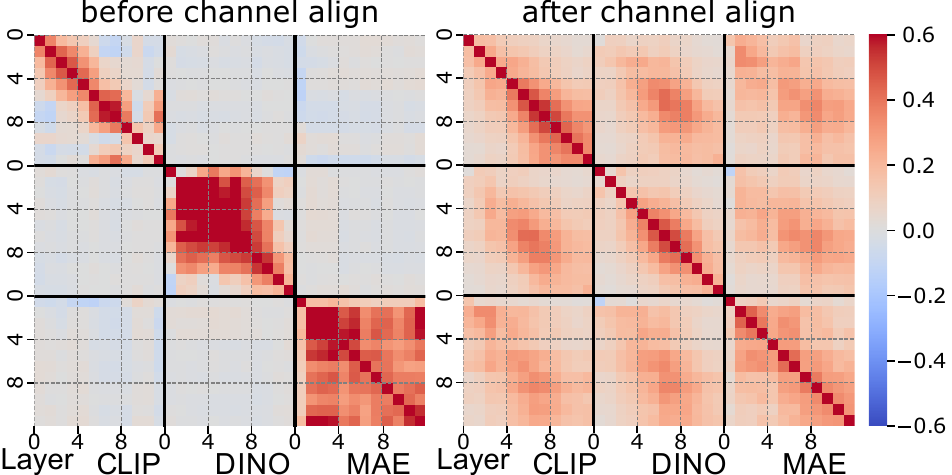}
    \caption{Cosine similarity of channel activation on the same image inputs.}

    \label{fig:model_similarity}
    \vspace{-4mm}
\end{wrapfigure}






Just as human languages might consist of distinct alphabets, features across different models appear superficially in embedding spaces as almost mutually orthogonal (\Cref{fig:model_similarity}). However, the underlying information that they represent can be similar. To jointly analyze features across models and layers, we proposed the \textbf{channel align transform} that linearly projects features to a universal space.

The learning signal for the channel align transform is provided by \textbf{brain response prediction}. Learning from brain prediction offers two advantages. First, brain response covers rich representations from all levels of semantics; the channel alignment removes irrelevant information while preserving the necessary and sufficient visual image features. Second, knowledge of brain regions provides an interpretable understanding of their corresponding channels derived from the alignment.

Our visual concept discovery is formulated as a graph partitioning task using \textbf{spectral clustering}. We term our approach for this channel align and graph partitioning as \textbf{AlignedCut}. Furthermore, a major challenge in applying spectral clustering to large graphs is the complexity scaling issue. To address this, we developed a \textbf{Nystrom-like approximation} to reduce the computational complexity.




\subsection{Brain-Guided Universal Channel Align}
\label{sec:channel_align}








\paragraph{Brain Dataset} 
We used the Algonauts 
competition \citep{gifford_algonauts_2023} release of Nature Scenes Dataset (NSD) \citep{allen_massive_2022}. Briefly, NSD provides an fMRI brain scan when watching COCO images. Each subject viewed 10,000 images over 40 hours of scanning. We used the first subject's publicly shared pre-processed and denoised \citep{prince_improving_2022} data.

\paragraph{Channel Align} Let $\mathcal{V} = \{ \bm{V}_1, \bm{V}_2, \cdots, \bm{V}_n | \bm{V}_i \in \mathbb{R}^{P \times D_i} \} $ be the set of image features, extracted from each layer of pre-trained ViT models, where $P = (H \times W + 1)$ is image patches and class token, $D_i$ is the hidden dimension. In particular, we used the attention layer output for each $\bm{V_i}$ without adding residual connections from previous layers.
Let $\mathcal{V'}$ be the channel-aligned features; the goal of channel alignment is to learn a set of linear transform $\mathcal{W} = \{ \bm{W}_1, \bm{W}_2, \cdots, \bm{W}_n | \bm{W}_i \in \mathbb{R}^{D_i \times D'} \}$. In the new $D'$ dimensional space, channels are aligned.
\begin{equation}
    \mathcal{V'} = \mathcal{V} \odot \mathcal{W} = \{ \bm{V}_1\bm{W}_1, \bm{V}_2\bm{W}_2, \cdots, \bm{V}_n\bm{W}_n | \bm{V}_i\bm{W}_i \in \mathbb{R}^{P \times D'}\}
\end{equation}
\paragraph{Brain Prediction} To produce a learning signal for channel align $\mathcal{W}$, features from $\mathcal{V'}$ are summed (\textit{not concatenated}) to do brain prediction. Let $\bm{Y} \in \mathbb{R}^{1 \times N}$ be the brain prediction target, where $N$ is the number of flattened 3D brain voxels, and $1$ indicates that each voxel's response is a scalar value. Let $F_\theta: \mathbb{R}^{P \times D'} \Rightarrow \mathbb{R}^{1 \times N}$ be the learned brain encoding model; without loss of generalizability, we set $F_\theta$ as global average pooling then linear weight $\bm{\beta}_\theta \in \mathbb{R}^{D' \times N}$ and bias $\bm{\epsilon}_\theta \in \mathbb{R}^{1 \times N}$:
\begin{equation}
    \left[ \avgpool_{p \in P}(\frac{1}{n} \sum_{i=1}^{n} (\bm{V}_i\bm{W}_i)) \times \bm{\beta}_\theta + \bm{\epsilon}_\theta \right] \Rightarrow \bm{Y}
\label{eq:brain_predict}
\end{equation}
\paragraph{Channel in the Brain's Space} Let $\mathcal{B} = \{ \bm{B}_1, \bm{B}_2, \cdots, \bm{B}_n | \bm{B}_i \in \mathbb{R}^{P \times N} \} $ be the set of channel activations in the brain's space. By defining $\bm{B}_i := \bm{V}_i\bm{W}_i\times\bm{\beta}_\theta$, we have the brain response prediction $\bm{Y} = \avgpool_{p \in P}(\frac{1}{n} \sum_{i=1}^{n} \bm{B}_i) + \bm{\epsilon}_\theta$ (\cref{eq:brain_predict}).  Intuitively, we linearly transformed the activation to the brain's space, such that the activation from all slots sum up to the brain response prediction.


\subsection{Graph Spectral Clustering}
\label{sec:spectral_clustering}




\paragraph{Spectral Clustering}  
We use spectral clustering for visual concepts discovery and image-channel analysis; it provides 1) soft-cluster embedding space and 2) unsupervised hierarchical image segmentation. Normalized Cut \citep{shi_normalized_2000} partitions the graph into sub-graphs with minimal cost of breaking edges. It embeds the graph into a lower dimensional eigenvector representation, where each eigenvector is a hierarchical sub-graph assignment.

Let $\bm{A} \in \mathbb{R}^{M \times M}$ be the symmetric affinity matrix, where $M$ denotes the total number of image patches. Given channel aligned features $\bm{V}' \in \mathbb{R}^{M \times D'}$, we define $\bm{A}_{ij} := \exp(\cos(\bm{V}'_i, \bm{V}'_j) - 1)$ such that $\bm{A}_{ij} > 0$ measures the similarity between data $i$ and $j$. The spectral clustering embedding $\bm{X} \in \mathbb{R}^{M \times C}$ is solved by the top $C$ eigenvectors of the following generalized eigenproblem:
\begin{equation}
    (\bm{D}^{-1/2} \bm{A} \bm{D}^{-1/2}) \bm{X} = \bm{X} \bm{\Lambda}
    \label{eq:ncut}
\end{equation}
where $\bm{D}$ is the diagonal degree matrix $\bm{D}_{ii} = \sum_j \bm{A}_{ij}$, $\bm{\Lambda}$ is diagonal eigenvalue matrix.



\paragraph{Nystrom-like Approximation}  Computing eigenvectors for $\bm{A} \in \mathbb{R}^{M \times M}$ is prohibitively expensive for enormous $M$ with a time complexity of $O(M^3)$. The original Nystrom approximation method \citep{fowlkes_spectral_2004} reduced the time complexity to $O(m^3 + m^2 M)$ by solving eigenvectors on sub-sampled graph $\bm{A}' \in \mathbb{R}^{m \times m}$, where $m \ll M$. In particular, the orthogonalization step of eigenvectors introduced the time complexity of $O(m^2 M)$. Because our Nystrom-like approximation trades the $O(m^2 M)$ orthogonalization term with the K-nearest neighbor, our Nystrom-like approximation reduced the time complexity to $O(m^3 + mM)$.  



Our Nystrom-like Approximation first solves the eigenvector $\bm{X}' \in \mathbb{R}^{m \times C}$ on a sub-sampled graph $\bm{A}' \in \mathbb{R}^{m \times m}$ using \Cref{eq:ncut}, then propagates the eigenvector from the sub-graph $m$ nodes to the full-graph $M$ nodes. Let $\bm{\Tilde{X}} \in \mathbb{R}^{M \times C}$ be the approximation $\bm{\Tilde{X}} \approx \bm{X}$. The eigenvector approximation $\bm{\Tilde{X}}_i$ of full-graph node $i \leq M$ is assigned by averaging the top K-nearest neighbors' eigenvector $\bm{X}'_k$ from the sub-graph nodes $k \leq m$:
\begin{equation}
\begin{aligned}
    \mathcal{K}_i &= KNN(\bm{A}_{*i}; m, K) = \argmax_{k \leq m} \sum_{k=1}^{K} \bm{A}_{ki}  \\
    \bm{\Tilde{X}}_i &= \frac{1}{\sum_{k\in \mathcal{K}_i} \bm{A}_{ki}} \sum_{k\in \mathcal{K}_i} \bm{A}_{ki} \bm{X}'_k 
\end{aligned}
\end{equation}
where $KNN(\bm{A}_{*i}; m, K)$ denotes KNN from full-graph node $i \leq M$ to sub-graph nodes $k \leq m$.



\subsection{Affinity Eigen-constraints as Regularization for Channel Align}
\label{sec:eigen_constraints}








\begin{wraptable}{r}{0.45\textwidth}
\centering
\vspace{-4mm}
\caption{Affinity eigen-constraints improved brain score ($R^2$: variance explained).}
\begin{tblr}{
  cell{1}{2} = {c=4}{c},
  cell{3}{1} = {c},
  cell{4}{1} = {c},
  cell{5}{1} = {c},
  hline{1} = {-}{0.1em},
  hline{6} = {-}{0.08em},
  hline{3} = {2-5}{0.05em},
}
       & ROI Brain Score $R^2$ ($\pm$ 0.001) &       &       &       \\
$\lambda_{eigen}$  & V1               & V4    & EBA   & all   \\
1.0    & \textbf{0.170}            & \textbf{0.181} & 0.295 & \textbf{0.196} \\
0.1    & 0.167            & 0.179 & 0.294 & 0.193 \\
0    & 0.155            & 0.166 & 0.296 & 0.188 
\end{tblr}
\label{tab:brain_score}
\end{wraptable}

While brain prediction can provide strong supervision for the learned channel align operation, we observed that the quality of unsupervised segmentation dropped after the channel alignment.  To address this issue, a regularization term is added:
\begin{equation}
\begin{aligned}
    \mathcal{L}_{eigen} &= \| \bm{X}_{b} \bm{X}_{b}^T - \bm{X}_{a} \bm{X}_{a}^T \|
\end{aligned}
\end{equation}
where $\bm{X}_{b}$ and $\bm{X}_{a} \in \mathbb{R}^{\tilde{m} \times c}$ are affinity matrix eigenvectors before and after channel alignment, respectively; $\tilde{m}=100$ are randomly sampled nodes in a mini-batch and $c=6$ are the top eigenvectors. 
The eigen-constraint preserves spectral clustering eigenvectors in dot-product space, invariant to random rotations in eigenvectors. We found adding eigen-constraints improved both the performance of segmentation (\Cref{fig:imgseg_pascal_score}) and the brain prediction score (\Cref{tab:brain_score}). 

\section{Results}

\vspace{-3mm}

Our spectral clustering analysis aims to discover visual concepts that share the same pattern of channel activation across different models and layers. 
However, implementing spectral clustering analysis comes with two main challenges. First, the models sit in different feature spaces, so direct clustering will not reveal their overlap and similarities. Second, when scaling up to a large graph, spectral clustering is computationally expensive. 


To address the first challenge, we developed our channel align transform to align features into a universal space. We extracted features from all 12 layers of the CLIP (ViT-B, OpenAI) \citep{radford_learning_2021}, DINOv2 (ViT-B with registers) \citep{darcet_vision_2024}, and MAE (ViT-B) \citep{he_masked_2022} and then transformed features from each layer into the universal feature space.

To address the second challenge, we developed our Nystrom-like approximation to reduce the computational complexity. We extracted features from 1000 ImageNet \citep{deng_imagenet_2009} images, with each image consisting of 197 patches per layer. The entire product space of all images and features totaled $M = 7\mathrm{e}{+6}$ nodes, from which we applied our Nystrom-like approximation with sub-sampled $m = 5\mathrm{e}{+4}$ nodes and KNN $K=100$, computing the top 20 eigenvectors.


To visualize the affinity eigenvectors, the top 20 eigenvectors were reduced to a 3-dimensional space by t-SNE, and a color value was assigned to each node by the RGB cube. We call this approach AlignedCut color.

\begin{figure}[h]
    \centering
    \includegraphics[width=\linewidth]{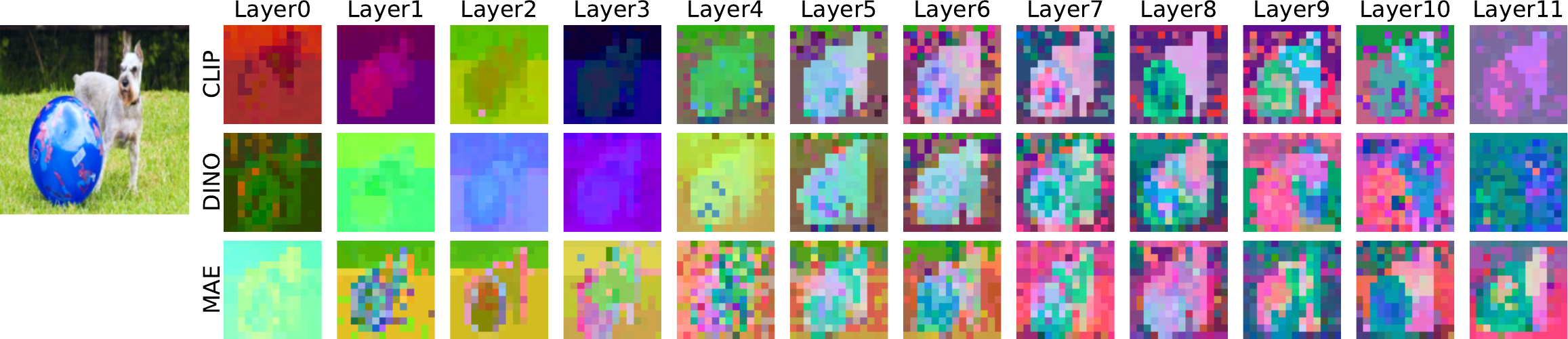}
    \includegraphics[width=\linewidth]{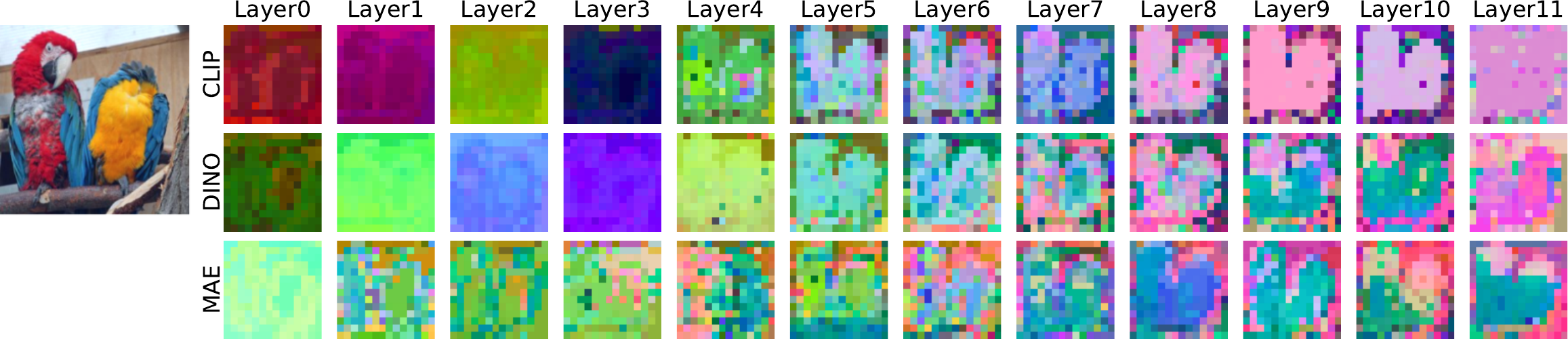}
    \includegraphics[width=\linewidth]{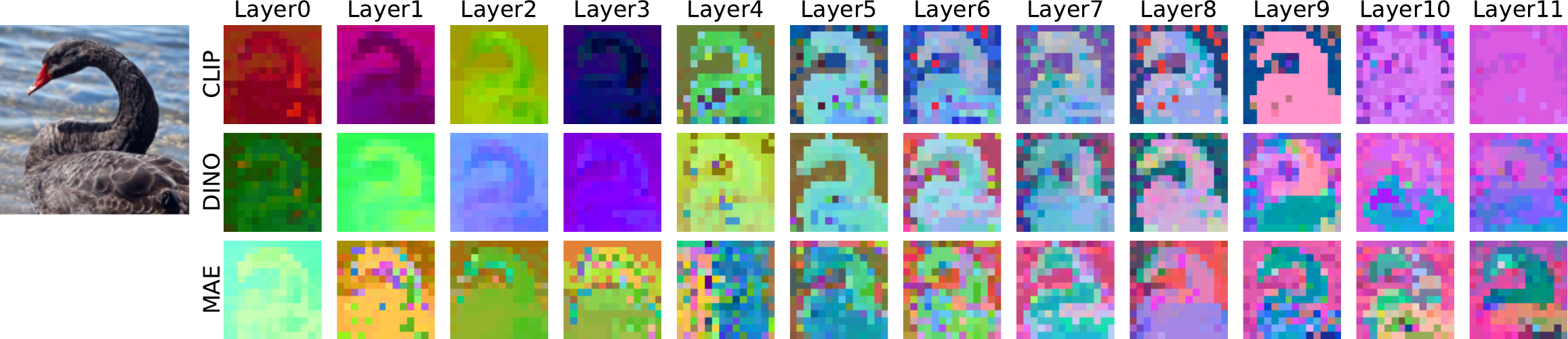}
    \caption{Spectral clustering in the universal channel aligned feature space. The image pixels are colored by our approach AlignedCut, the pixel RGB value is assigned by the 3D spectral-tSNE of the top 20 eigenvectors. The coloring is consistent across all images, layers, and models.}
    \label{fig:eig_tsne_images}
\end{figure}

In \Cref{fig:eig_tsne_images}, we displayed the analysis, AlignedCut color, and made the following observations:


 \quad  1. In CLIP layer-5, DINO layer-6, and MAE layer-8, there is class-agnostic figure-ground separation, with foreground objects from different categories grouped into the same AlignedCut color.
    
 \quad 2.  In CLIP layer-9, there is a class-specific separation of foreground objects, with foreground objects grouped into AlignedCut colors with associated semantic categories.
    
 \quad 3.  Before layer-3, CLIP and DINO produce the same AlignedCut color regardless of the image input. From layer-4 onwards, the AlignedCut color smoothly changes over layers.



\vspace{-4mm}
\subsection{Figure-ground representation emerge before categories}

In this section, we benchmark each layer in CLIP with unsupervised segmentation. The key findings from this benchmarking are: \textbf{1)} The figure-ground representation emerges at CLIP layer-4 and is preserved in subsequent layers; \textbf{2)} Categories emerge over layers, peaking at layer-9 and layer-10.

\begin{figure}[h]
    \centering
    \vspace{-2mm}
    \includegraphics[width=0.85\linewidth]{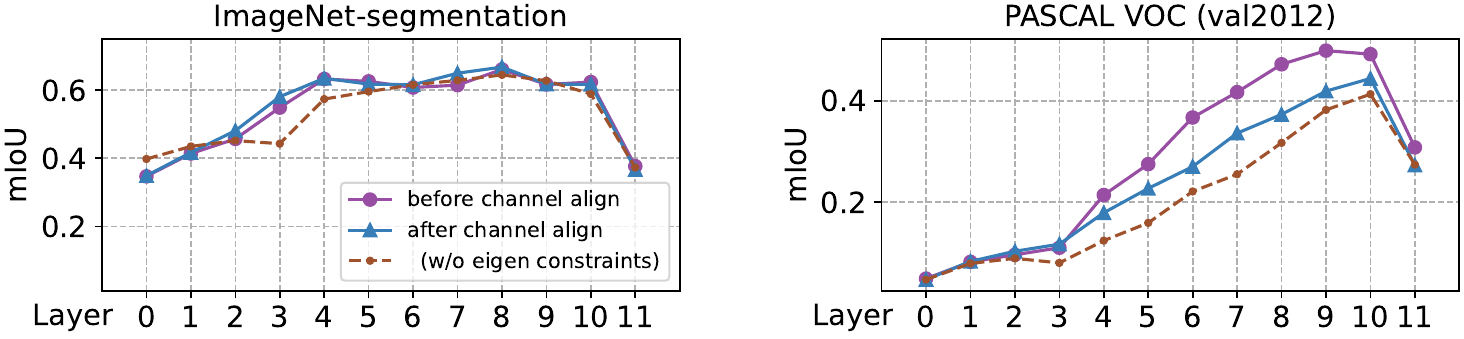}
    \hspace{0.025\linewidth}
    \vspace{-2mm}
    \caption{Unsupervised segmentation scores from spectral clustering on each CLIP layer. ImageNet-segmentation dataset is used with binary figure-ground labels, and the mIoU score peaks plateau from layer-4 to layer-10. In PASCAL VOC with 20 class labels, the mIoU score peaks at layer-9.}

        
    \label{fig:imgseg_pascal_score}
    \vspace{-2mm}
\end{figure}

\textbf{\textit{From which layers did the figure-ground and category representations emerge?}}
We conducted experiments that compared the unsupervised segmentation scores across layers, tracing how well each representation is encoded at each layer. We used two datasets: a) ImageNet-segmentation \citep{guillaumin_imagenet_2014} with binary figure-ground labels, and b) PASCAL VOC \citep{everingham_pascal_2010} with 20 category labels.
The results are presented in \Cref{fig:imgseg_pascal_score}. On the ImageNet-segmentation benchmark, the score peaks at layer-4 (mIoU=0.6) and plateaus in subsequent layers, suggesting that the figure-ground representation is encoded and preserved from layer-4 onwards. On the PASCAL VOC benchmark, the score peaks at layer-9 and layer-10 (mIoU=0.5) even though it is low at layer-4 (mIoU=0.2), indicating that category information is encoded at layer-9 and layer-10. Overall, we conclude that the figure-ground representation emerges before the category representation.



\subsection{Visual concepts: class-agnostic figure-ground}

In this section, we use brain activation heatmaps and image similarity heatmaps to describe figure-ground visual concepts. The key findings from these heatmaps are: \textbf{1)} 
The figure vs. ground pixels activate different channels; \textbf{2)} The figure-ground visual concept is class-agnostic; \textbf{3)} The figure-ground visual concept is consistent across models.

\begin{figure}[h]
    \centering
    \vspace{-4mm}
    \includegraphics[width=\linewidth]{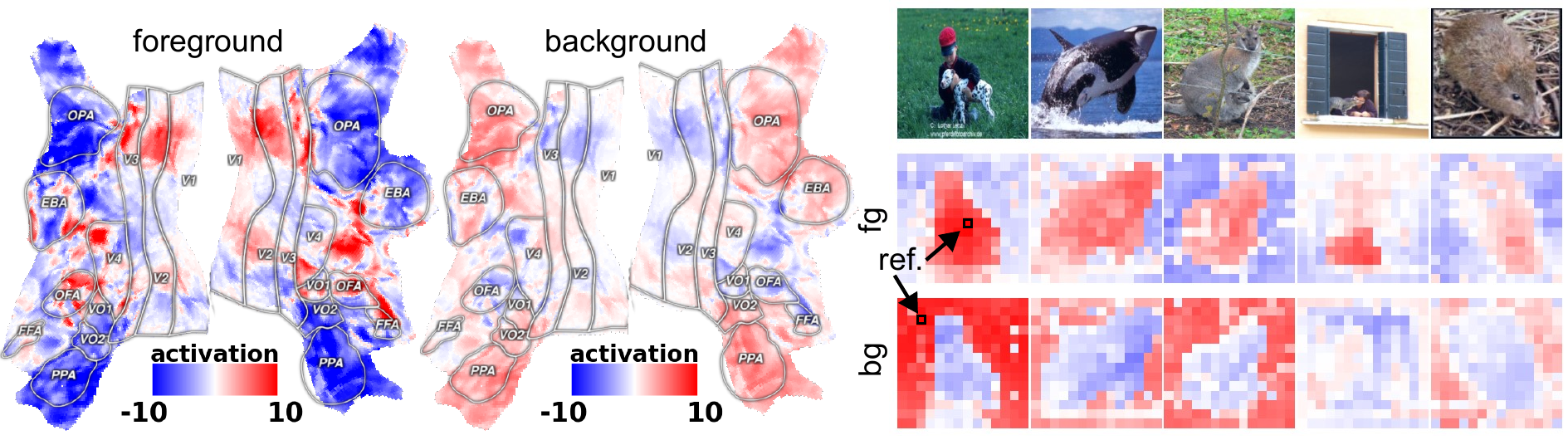}
    \vspace{-6mm}
    \caption{
    The figure-ground visual concepts in CLIP layer-5. \textbf{\textit{Left:}} Mean activation of foreground or background pixels, linearly transformed to the brain's space. \textbf{\textit{Right:}} Cosine similarity from \textit{\textbf{one}} reference pixel marked. The figure-ground visual concepts are agnostic to image categories.
    }
    \label{fig:fgbg_one_pixel_seg}
    \vspace{-2mm}
\end{figure}

\textbf{\textit{How can the channel activation patterns of the figure-ground visual concept be described?}}
We averaged the channel activations from foreground and background pixels, using the ground-truth labels from the ImageNet-segmentation dataset. The averaged channel activations were transformed into the brain's space. In \Cref{fig:fgbg_one_pixel_seg}, foreground pixels exhibit positive activation in early visual brain ROIs (V1 to V4) and the face-selective ROI (FFA), while negatively activating place-selective ROIs (OPA and PPA). Interestingly, background pixels activate the reverse pattern compared to foreground pixels. Overall, the figure and ground pixels 
activate distinct brain ROIs.

\textbf{\textit{Is the figure-ground visual concept class-agnostic?}}
We manually selected \textit{one} pixel and computed the cosine similarity to all of the other image pixels. In \Cref{fig:fgbg_one_pixel_seg}, the results demonstrate that one pixel (on the human) could segment out foreground objects from all other classes (shark, dog, cat, rabbit). The same result holds true for one background pixel. We conclude that the figure-ground visual concept is class-agnostic.

\begin{figure}
    \centering
    \vspace{-6mm}
    \includegraphics[width=\linewidth]{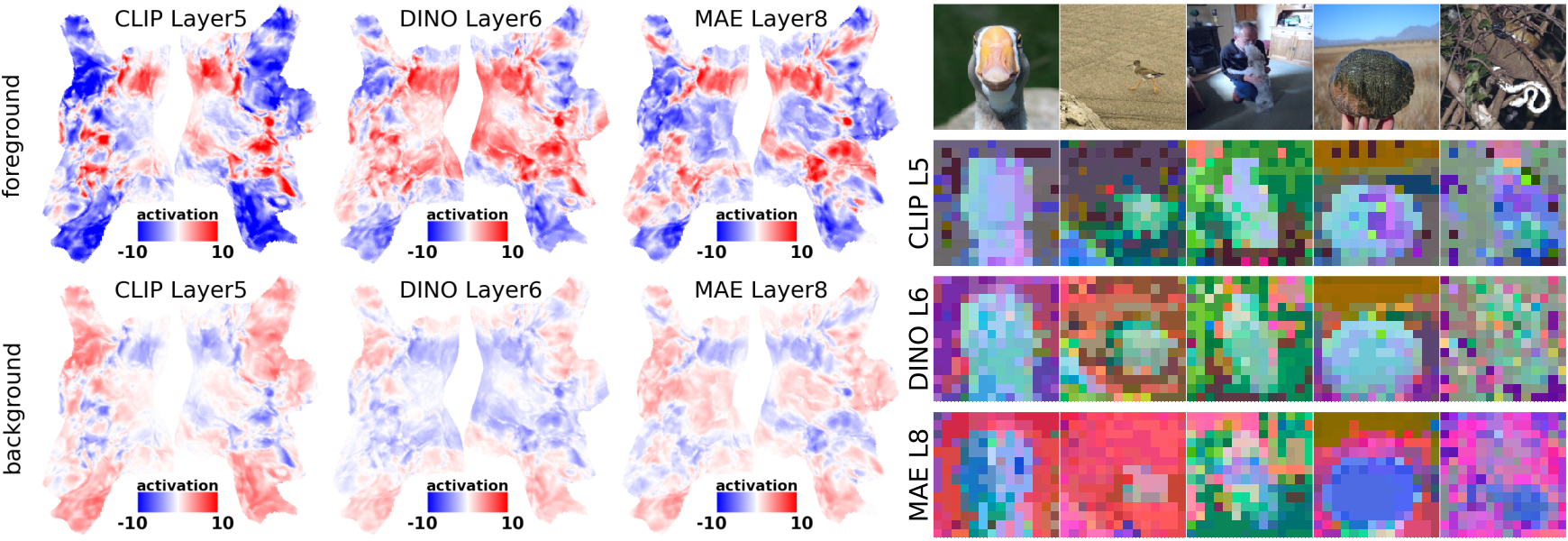}
    \vspace{-4mm}
    \caption{The same figure-ground visual concepts are found in CLIP, DINO and MAE. \textbf{\textit{Left:}} Mean activation of all foreground (top) and background (bottom) pixels; the three models exhibit similar activation patterns. \textbf{\textit{Right:}} AlignedCut, pixels colored by 3D spectral-tSNE of the top 20 eigenvectors; the three models show similar grouping colors for foreground pixels.}
    
    \label{fig:fgbg_consistent_across_models}
    \vspace{-4mm}
\end{figure}

\textbf{\textit{Is the figure-ground visual concept consistent across models?}}
We performed the channel analysis for CLIP, DINO, and MAE. In \Cref{fig:fgbg_consistent_across_models}, the foreground or background pixels activates similar brain ROIs across the three models. Additionally, spectral clustering grouped the representations of foreground objects into similar colors for CLIP and DINO (light blue), the grouping for MAE is less similar (dark blue). Overall, the figure-ground visual concept is consistent across models.

\vspace{-1mm}
\subsection{Visual concepts: categories}

In this section we use AlignedCut to discover category visual concepts. The key findings from the category visual concepts are: \textbf{1)} Class-specific visual concepts activate diverse brain regions; \textbf{2)} Visual concepts with higher channel activation values are more consistent.

\begin{figure}[h]
    \centering
    \vspace{-2mm}
    \includegraphics[width=\linewidth]{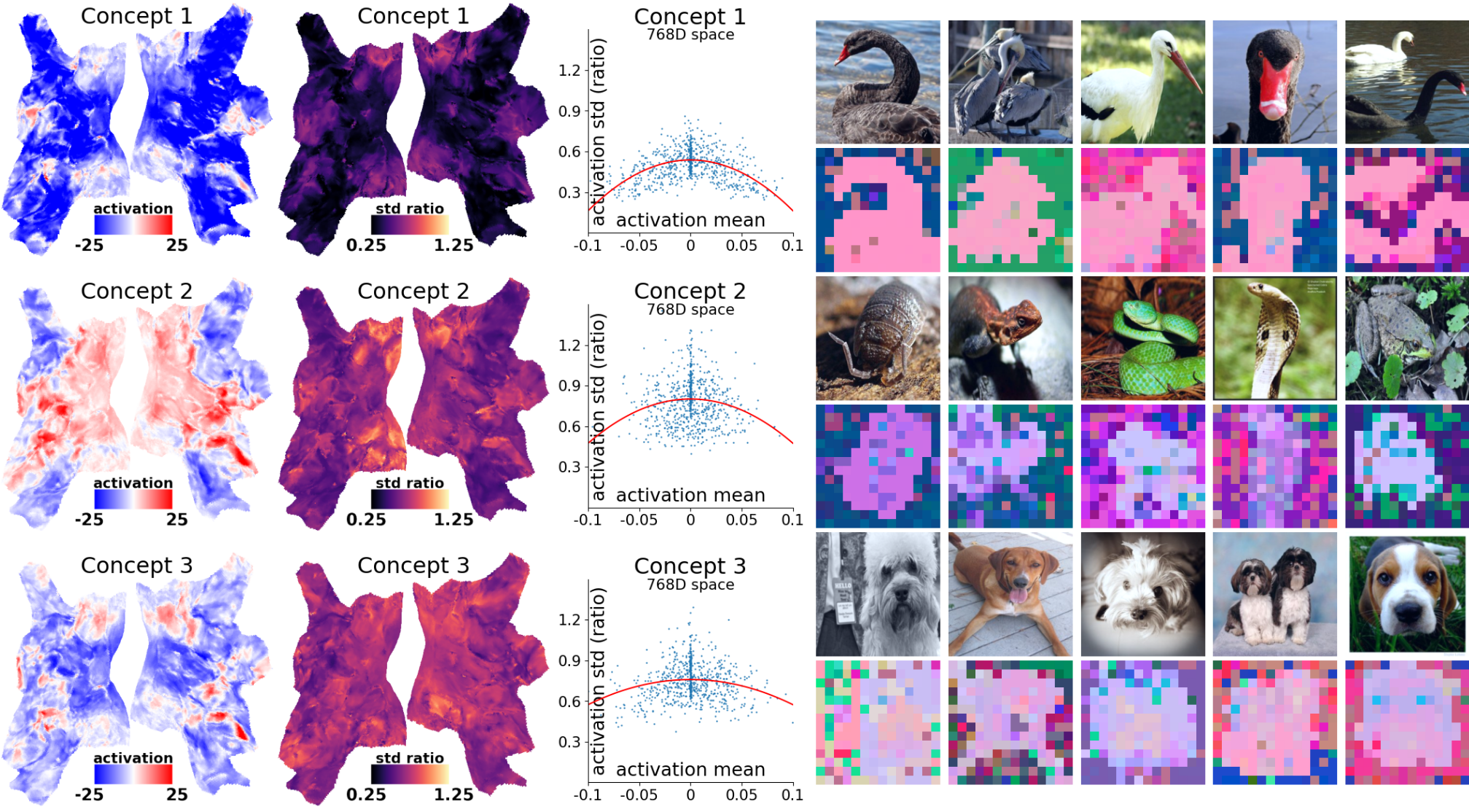}
    \vspace{-4mm}
    \caption{ Category visual concepts in CLIP layer-9. \textbf{\textit{Left:}} Mean activation of all pixels within an Euclidean sphere centered at the visual concept in the 3D spectral-tSNE space; the concepts activate different brain regions. \textbf{\textit{Middle:}} The standard deviation negatively correlates with absolute mean activations. \textbf{\textit{Right:}} AlignedCut, pixels colored by 3D spectral-tSNE of the top 20 eigenvectors.
    }
    
    \label{fig:category_concepts_mean_and_std_and_tsne}
    \vspace{-2mm}
\end{figure}

\textbf{\textit{How does each class-specific concept activate the channels?}} To answer this question, we sampled class-specific concepts from CLIP layer-9. First, we used farthest point sampling to identify candidate centers in the 3D spectral-tSNE space. Then, each candidate center was grouped with its neighboring pixels within an Euclidean sphere in the spectral-tSNE space. Finally, the channel activations of the grouped pixels were averaged to produce the mean channel activation for each visual concept. In \Cref{fig:category_concepts_mean_and_std_and_tsne}, Concept 1 (duck, goose) negatively activates late brain regions; Concept 2 (snake, turtle) positively activates early brain regions and also FFA; Concept 3 (dog) negatively activates early brain regions. Overall, category-specific visual concepts activate diverse brain regions.

\textbf{\textit{How do we quantify the consistency of each visual concept?}} Qualitatively, Concept 1 exhibits more consistent coloring (\Cref{fig:category_concepts_mean_and_std_and_tsne}, pink) than Concept 3 (purple). To further quantify this observation, we computed the mean and standard deviation of channel activations for each Euclidean sphere centered on a concept. In \Cref{fig:category_concepts_mean_and_std_and_tsne}, there is a reverse U-shape relation between magnitude and standard deviation. The reverse U-shape implies that larger absolute mean channel activation corresponds to lower standard deviation. Overall, higher channel activation magnitudes suggest more consistent visual concepts.




\subsection{Transition of visual concepts over layers}












In this section, instead of using 3D spectral-tSNE, we use 2D spectral-tSNE to trace the layer-to-layer feature computation. The key findings of spectral-tSNE in 2D are: \textbf{1)} The figure vs. ground pixels are encoded in separate spaces in late layers; \textbf{2)} The representations for foreground and background bifurcate at CLIP layer-4 and DINO layer-5.

\begin{figure}[h]
    \centering
    \vspace{-2mm}
    \includegraphics[width=\linewidth]{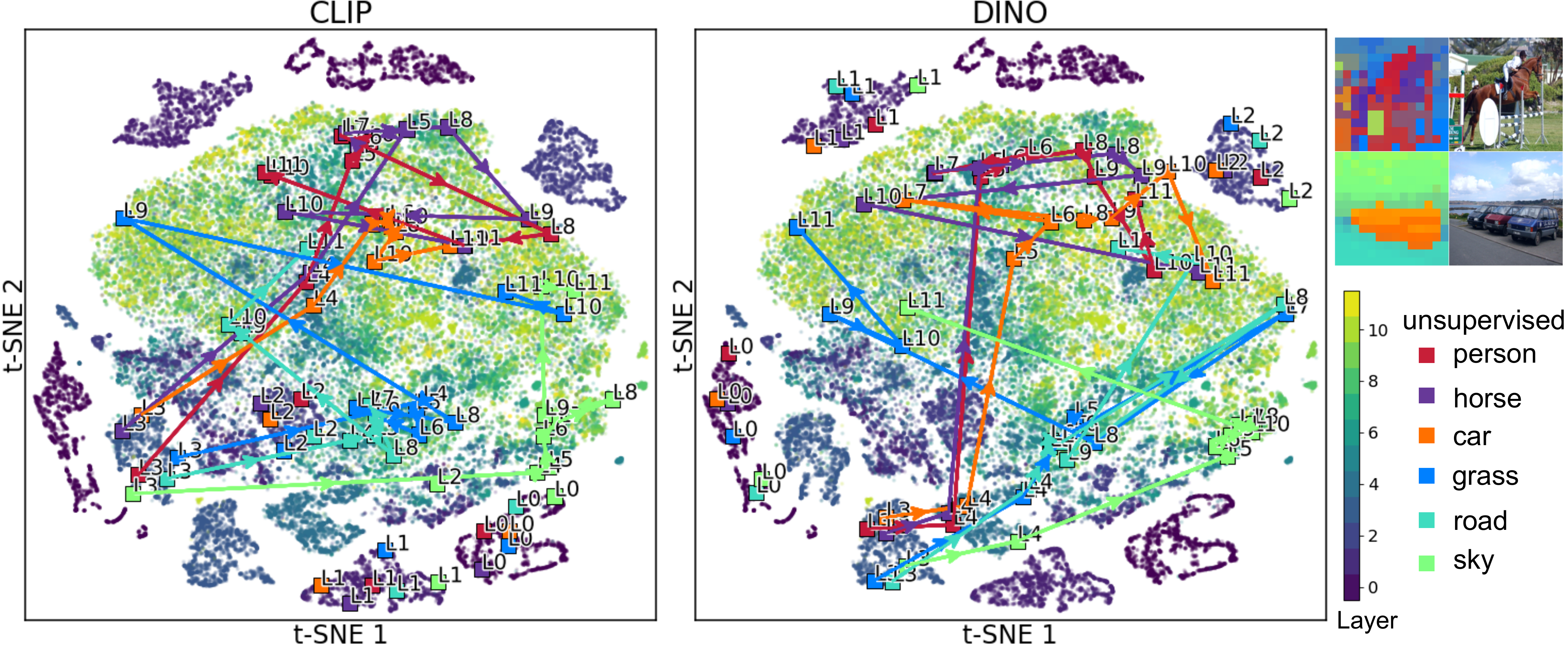}
    \vspace{-4mm}
    \caption{Trajectory of feature progression in layers for six example pixels. \textbf{\textit{Left:}} 2D spectral-tSNE plot of the top 20 eigenvectors, jointly clustered across all models; the foreground and background pixels bifurcate at CLIP layer-4 and DINO layer-5. \textbf{\textit{Right:}} Pixels colored by unsupervised segmentation.}
    
    \label{fig:trajectory_at_tsne_motion_flow}
    \vspace{-2mm}
\end{figure}

\textbf{\textit{How does the network encode figure and ground pixels in each layer?}} We performed spectral clustering and 2D t-SNE on the top 20 eigenvectors to project all layers into a 2D spectral-tSNE space. In \Cref{fig:trajectory_at_tsne_motion_flow}, we found that all foreground and background pixels are grouped together in each early layer. Each early layer (dark dots) forms an isolated cluster separate from other layers, while late layers (bright dots) are grouped in the center. In the late layers, there is a separation where foreground pixels occupy the upper part of 2D spectral-tSNE space, while background pixels occupy the middle part. Overall, foreground and background pixels are encoded in separate spaces in late layers.

\begin{figure}
    \centering
    \vspace{-7mm}
    \includegraphics[width=\linewidth]{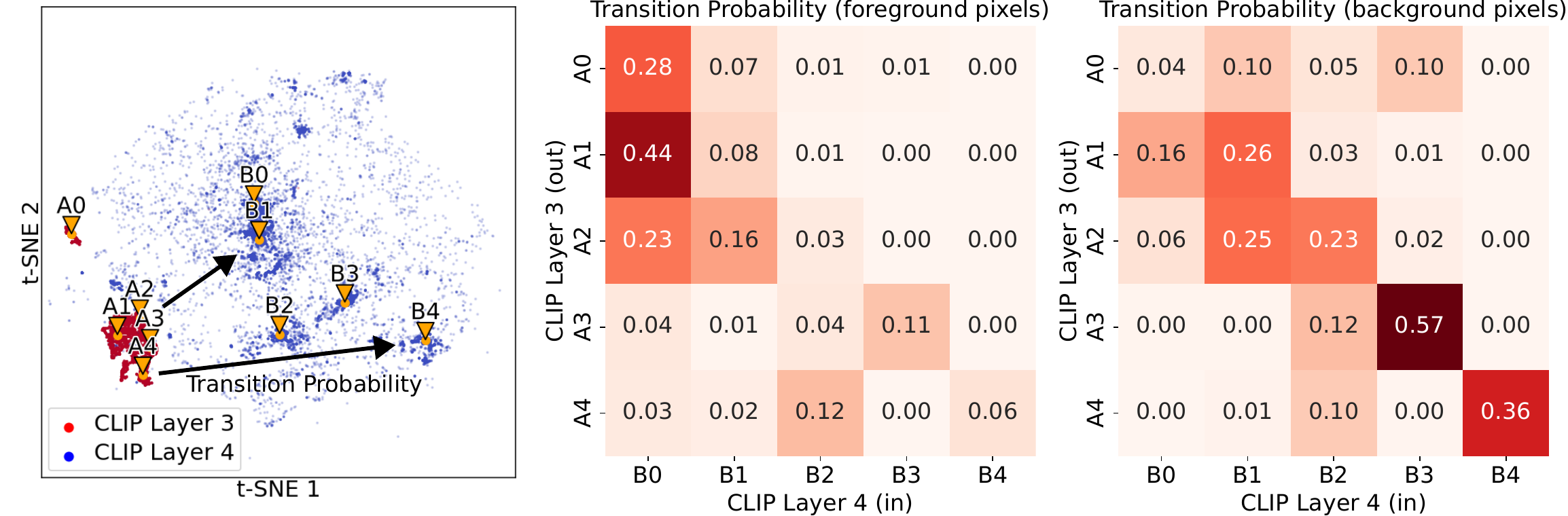}
    \vspace{-5mm}
    \caption{Transition probability of visual concepts from CLIP layer-3 to layer-4. \textbf{\textit{Left:}} Five visual concepts sampled from CLIP layer-3 and layer-4. \textbf{\textit{Right:}} Transition probability measured separately for foreground and background pixels; a bifurcation occurs where foreground pixels have more traffic to concept B0, while background pixels have more traffic to concepts B3 and B4.    }
    \label{fig:transition_matrix_at_l3_l4}
    \vspace{-4mm}
\end{figure}

\textbf{\textit{How does the network process each pixel from layer to layer?}} In the 2D spectral-tSNE plot, we traced the trajectory for each pixel from layer-3 to the last layer. In \Cref{fig:trajectory_at_tsne_motion_flow}, we found that the trajectories for foreground and background pixels bifurcate: foreground pixels (person, horse, car) traverse to the upper side and remain within the upper side; background pixels (grass, road, sky) jump between the middle right and left sides. The same bifurcation is consistently observed for CLIP from layer-3 to layer-4 and DINO from layer-4 to layer-5. Furthermore, to quantify the bifurcation for foreground and background pixels, we first sampled 5 visual concepts from CLIP layer-3 and layer-4. Then, we measured the transition probability between visual concepts, defined as the proportion of pixels that transited from an Euclidean circle around concept A to a circle around concept B. In \Cref{fig:transition_matrix_at_l3_l4}, the transition probability of foreground pixels to the upper side (A1 to B0) is higher than that of background pixels (0.44 vs. 0.16), while the transition probability of background pixels to the right side (A4 to B4) is higher than that of foreground pixels (0.36 vs. 0.06). Overall, this suggests a bifurcation of figure and ground pixel representations at the middle layers of both CLIP and DINO.

\section{Related Work}

\textbf{Mechanistic Interpretability} is a field of study that intends to understand and explain the inner working mechanisms of deep networks. One approach is to interpret individual neurons \citep{bau_network_2017, dravid_rosetta_2023} and circuit connections between neurons \citep{olah_zoom_2020}. Another approach is to interpret transformer attention heads \citep{gandelsman_interpreting_2024} and circuit connections between attention heads \citep{wang_interpretability_2023}. Other approaches also looked into the role of patch tokens \citep{sun_massive_2024}. These approaches made the assumption that channels are aligned within the same model; we compare across models by actively aligning the channels to a universal space.

\textbf{Spectral Clustering} is a graphical method to analyze data grouping in the eigenvector space. Spectral methods have been widely used for unsupervised image segmentation \citep{shi_normalized_2000, von_luxburg_tutorial_2007, wu_unsupervised_2018, wang_cut_2023}. One major challenge for applying spectral clustering to large graphs is the complexity scaling issue. To solve the scaling issue, the Nystrom approximation \citep{fowlkes_spectral_2004} approaches solve eigenvectors on sub-sampled graphs and then propagate to the full graph. Another approach is the gradient-based eigenvector solver \citep{zhang_deciphering_2023}, which solves the eigenvectors in mini-batches. Our proposed Nystrom-like approximation achieves a computational speedup over the original Nystrom approximation, albeit at the expense of weakened orthogonality of the eigenvectors.

\textbf{Brain Encoding Model} is widely used by the computational neuroscience community \citep{kriegeskorte_cognitive_2018}. They have been using deep nets to explain the brain's function. One approach is to use the gradient of the brain encoding model to find the most salient image features \citep{sarch_brain_2023}. Another approach generate text caption for brain activation \citep{luo_brainscuba_2024}. Other approaches compare brain prediction performance for different models \citep{schrimpf_integrative_2020}. The field focused on using deep nets as a tool to explain the brain's function; we go in the opposite direction by using the brain to explain deep nets.

\section{Conclusion and Limitations}
We present a novel approach to interpreting deep neural networks by leveraging brain data. Our fundamental innovation is twofold: First, we use brain prediction as guidance to align channels from different models into a universal feature space; Second, we developed a Nystrom-like approximation to scale up the spectral clustering analysis. Our key discovery is that recurring visual concepts exist across networks and layers; such concepts correspond to different levels of objects, ranging from figure-ground to categories. Additionally, we quantified the information flow from layer to layer, where we found a bifurcation of figure-ground visual concepts.

\textbf{Limitations.} While the learned channel align transformation projects all features onto a universal feature space, the nature of learned transformation does not preserve all the information. There is a small drop in unsupervised segmentation performance after channel alignment, which is not fully addressed by our proposed eigen-constraint regularization. Secondly, as a trade-off for faster computation, our Nystrom-like approximation does not produce strictly orthogonal eigenvectors. To produce expressive eigenvectors, our approximation relies on using larger sub-sample sizes than the original Nystrom method.

\noindent\textbf{Acknowledgements.}
This work is supported by funds provided by the National Science Foundation and by DoD OUSD (R\&E) under Cooperative Agreement PHY-2229929 (The NSF AI Institute for Artificial and Natural Intelligence).

\clearpage
\newpage

\bibliographystyle{apalike}
\bibliography{main}


\newpage
\appendix


\section{Appendix overview}

\begin{enumerate}
\item \Cref{sec:app_brain_roi} summarizes background of brain ROIs. 
\item \Cref{sec:app_implementation} is implementation details 
  \begin{enumerate}[label*=\arabic*.]
    \item Additional regularization terms
    \item Brain encoding model training loss function
    \item Unsupervised segmentation evaluation pipeline
    \item Nystrom-like approximation for t-SNE
  \end{enumerate}
\item \Cref{sec:app_3d} lists more image examples from the 3D spectral-tSNE.
\item \Cref{sec:fgbg} lists figure-ground channel activation for every model and layer.
\item \Cref{sec:cate} lists more example category-specific visual concepts.
\item \Cref{sec:2d} lists more example pixels from the 2D spectral-tSNE information flow.
\end{enumerate}

\newpage

\section{Brain Region Background Knowledge}
\label{sec:app_brain_roi}

\begin{figure}[ht]

    \centering
    \captionsetup{type=figure}
    \includegraphics[width=\linewidth]{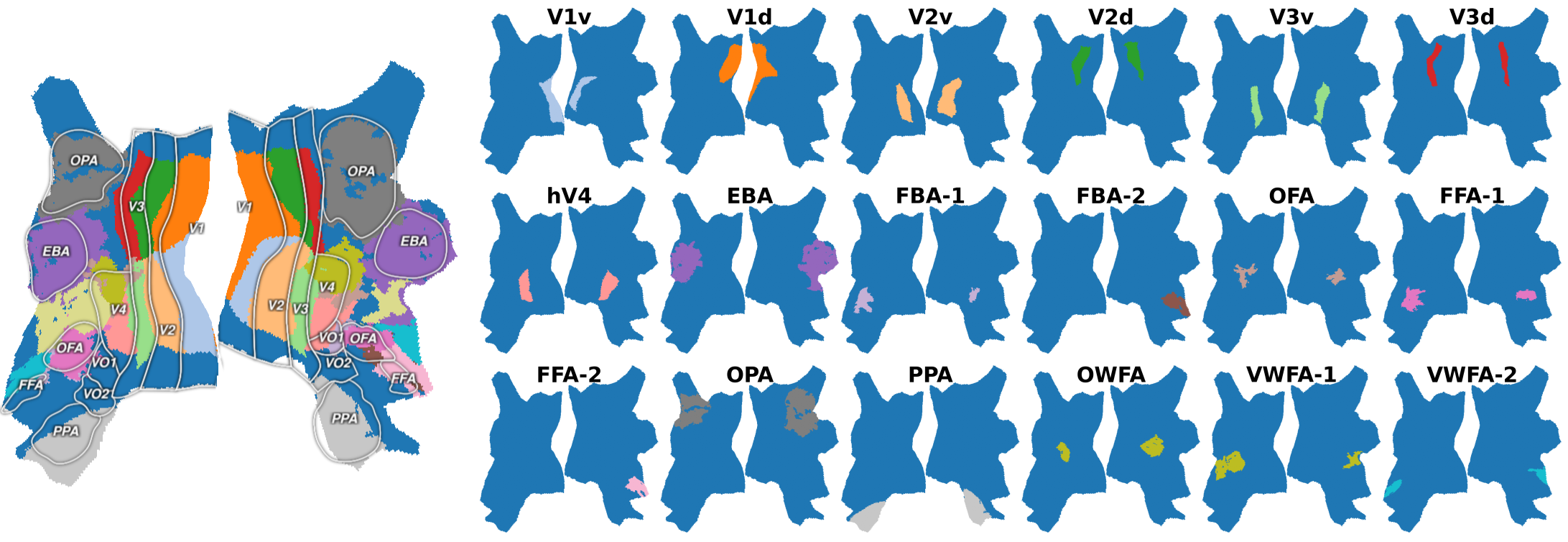}
    \captionof{figure}{\textbf{Brain Region of Interests (ROIs)}. V1v: ventral stream, V1d: dorsal stream.}
    \label{fig:supp_rois}

\vspace{4mm}

    \captionof{table}{Known function and selectivity of brain region of interests (ROIs).}
    \resizebox{0.99\linewidth}{!}{
    \begin{tblr}{
      hline{1,3} = {-}{0.08em},
    }
    \textbf{ROI name}  \quad\quad\quad & V1 V2 V3 & V4 \quad & EBA FBA & OFA FFA & OPA \quad & PPA \quad & OWFA VWFA \\
    \textbf{Known Function/Selectivity} \quad\quad & primary visual & mid-level & body & face & navigation & scene & words \\
    \end{tblr}
    }
    \label{tab:supp_rois}
\vspace{10mm}

\end{figure}

This section briefly summarizes the known functions of key brain regions of interest (ROIs). \Cref{fig:supp_rois} provides an overview of these brain ROIs. \Cref{tab:supp_rois} lists the known functions and selectivities for each ROI.

In brief, V1 to V3 are the primary visual stream, which is further divided into ventral (lower) and dorsal (upper) streams. V4 is a mid-level visual area. EBA (extrastriate body area) and FBA (fusiform body area) are selectively responsive to bodies, while FFA (fusiform face area) and OFA (occipital face area) show selectivity for faces. OWFA (occipital word form area) and VWFA (visual word form area) are selective for written words. PPA (parahippocampal place area) exhibits selectivity for scenes and places, and OPA (occipital place area) is involved in navigation and spatial reasoning.

Visual information processing in the brain follows a hierarchical, feedforward organization. Beginning in the primary visual cortex (V1) and progressing through higher visual areas like V2, V3, and V4, neurons exhibit increasingly large receptive fields and represent increasingly abstract visual concepts. While neurons in V1 encode low-level features like edges and orientations within a small portion of the visual field, neurons in V4 synthesize more complex patterns and object representations across a larger area of the visual input.

\newpage
\section{Implementation Details}
\label{sec:app_implementation}

\subsection{Additional Regularization for Channel Align Transformation}

Additional Regularization are added to the channel align transform to ensure good properties of the aligned features: 1) zero-centered, 2) small covariance between channels, and 3) focal loss.

\textbf{Zero-centered regularization.}
We did not apply z-score normalization to the extracted features; instead, we added a regularization term to ensure the transformed features are zero-centered. Recall that the channel-aligned transformed feature \(\bm{V}' \in \mathbb{R}^{M \times D'}\), where \(M\) is the number of data points and \(D'\) is the hidden dimension. The zero-center loss is defined as:
\begin{equation}
    \mathcal{L}_{\mathrm{zero}} = \frac{1}{D'} \frac{1}{M} \sum_{i\leq M, j\leq D'} v'_{ij}
\end{equation}
\textbf{Covariance regularization.}
We used the covariance loss to minimize the off-diagonal elements in the covariance matrix of the transformed feature $C(\bm{V}')$, aiming to bring them close to $\mathbf{0}$. Recall that channel align transformed feature $\bm{V}' \in \mathbb{R}^{M \times D'}$, where $M$ is number of data, $D'$ is the hidden dimension. The covariance loss is defined as:
\begin{equation}
    \mathcal{L}_{\mathrm{cov}} = \frac{1}{D'} \sum_{i \neq j}[C(\bm{V}')]_{i, j}^2, \text{ where } C(\bm{V}') = \frac{1}{M-1} \sum_{i=1}^M\left(v'_i-\bar{v'}\right)\left(v'_i-\bar{v'}\right)^T, \bar{v'}=\frac{1}{M} \sum_{i=1}^M v'_i.
\end{equation}
\textbf{Focal Loss.}
\citet{lin_focal_2017} introduced focal loss, which dynamically assigns smaller weights to the loss function for hard-to-classify classes. In our scenario, we apply spectral clustering on the affinity matrix \(\bm{A}_a \in \mathbb{R}^{M \times M}\) after performing the channel alignment transform, where \(M\) represents the number of data points. Due to the characteristics of spectral clustering, disconnected edges play a more critical role than connected edges. Adding an edge between disconnected clusters significantly reshapes the eigenvectors, while adding edges to connected clusters has only a minor impact. Therefore, we aim to assign larger weights to disconnected edges in the loss function:
\begin{equation}
\begin{aligned}
    \mathcal{L}_{eigen} &= \| ( \bm{X}_{b} \bm{X}_{b}^T - \bm{X}_{a} \bm{X}_{a}^T ) * \exp(-\bm{A}_b) \|
\end{aligned}
\end{equation}
where \(\bm{A}_b \in \mathbb{R}^{M \times M}\) is the affinity matrix before the channel alignment transform, element wise dot-product to \(\exp(-\bm{A}_b)\) assigned larger wights for disconnected edges. \(\bm{X}_{b} \in \mathbb{R}^{M\times C}, \bm{X}_{a} \in \mathbb{R}^{M\times C}\) are eigenvectors before and after channel align transform, respectively.

\subsection{Brain Encoding Model Training Loss}

Let \(\bm{Y} \in \mathbb{R}^{1 \times N}\) represent the brain prediction target, where \(N\) is the number of flattened 3D brain voxels, and the \(1\) indicates that each voxel's response is a scalar value. \(\bm{\hat{Y}}\) is the model's predicted brain response. The brain encoding model training loss is the L1 loss:
\begin{equation}
\begin{aligned}
    \mathcal{L}_{brain} &= \| \bm{Y} - \bm{\hat{Y}} \|
\end{aligned}
\end{equation}
\subsection{Total Training Loss}

The total training loss is a combination of the following components: 1) brain encoding model loss, 2) eigen-constraint regularization, 3) zero-centered regularization, and 4) covariance regularization:
\begin{equation}
\begin{aligned}
    \mathcal{L} &= \mathcal{L}_{brain} + \lambda_{eigen} \mathcal{L}_{eigen} + \lambda_{zero} \mathcal{L}_{zero} + \lambda_{cov} \mathcal{L}_{cov}
\end{aligned}
\end{equation}
where we set $\lambda_{eigen} = 1$, $\lambda_{zero} = 0.01$, $\lambda_{cov} = 0.01$.

\newpage
\subsection{Oracle-based Unsupervised Segmentation Evaluation Pipeline}

Our unsupervised segmentation pipeline aims to benchmark and compare the performance across each single layer of the CLIP model. The evaluation pipeline is oracle-based:

\quad 1. Apply spectral clustering jointly across all images, taking the top 10 eigenvectors.

\quad 2. For each class of object (plus one background class), use ground-truth labels from the dataset to mask out the pixels and their eigenvectors, and then use the mean of the eigenvectors to define a center for each class.

\quad 3. Compute the cosine similarity of each pixel to all class centers.

\quad 4. For each pixel, if the maximum similarity to all classes is less than a threshold value, assign this pixel to the background class.

\quad 5. Assign pixels (with a similarity greater than the threshold value) to the class with the maximum similarity.

There's one hyper-parameter, the threshold value that requires different optimal value for each layer of CLIP. To ensure a fair comparison across all layers, the threshold value is grid-searched from 10 evenly spaced values between 0 and 1, the maximum mIoU score in the grid search is taken for each layer.

\subsection{Nystrom-like approximation for t-SNE}
To visualize the eigenvectors, we applied t-SNE to the eigenvectors $\bm{X} \in \mathbb{R}^{M \times C}$, where the number of data points $M$ span the product space of models, layers, pixels, and images. Due to the enormous size of $M = 7\mathrm{e}{+6}$ nodes, t-SNE suffered from complexity scaling issues. We again applied our Nystrom-like approximation to t-SNE, with sub-sampled $m = 10\mathrm{e}{+4}$ nodes and KNN $K=1$.

It's worth noting that, since the non-linear distance adjustment in t-SNE, it's crucial to use only one nearest neighbor $K=1$ for t-SNE.

\subsection{Computation Resource}
All of our experiments are performed on one consumer-grade RTX 4090 GPU. The brain encoding model training took 3 hours on 4GB of VRAM, spectral clustering eigen-decomposition on large graph took 10 minutes on 10GB of VRAM and 60GB of CPU RAM.

\subsection{Code Release}
Our code will be publicly released upon publication.

\newpage

\clearpage
\pagestyle{fancy}
\fancyhead{}
\fancyhead[RO,LE]{\textbf{3D Spectral-tSNE}}

\section{3D spectral-tSNE}
\label{sec:app_3d}

\begin{figure}[h]
    \centering
    \includegraphics[width=\linewidth]{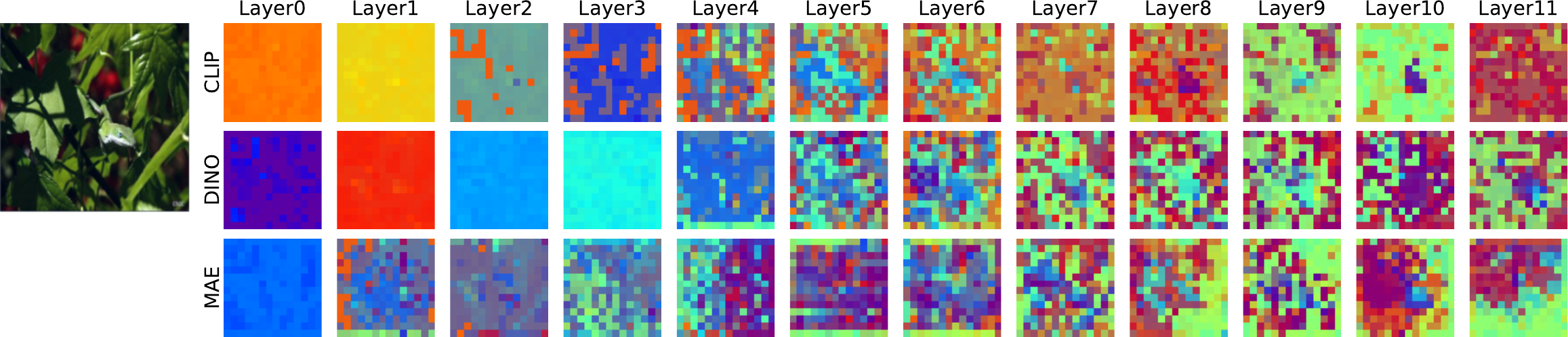}
    \includegraphics[width=\linewidth]{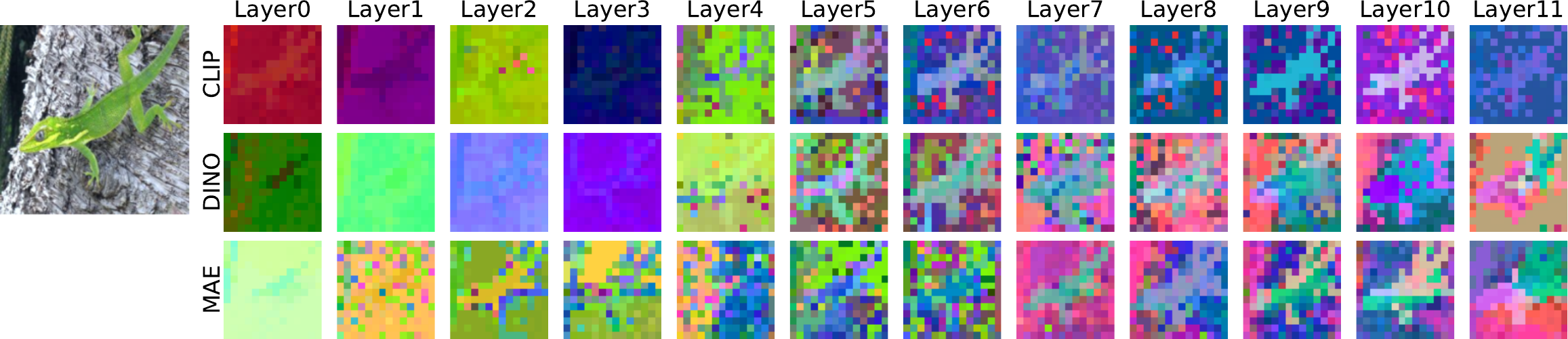}    \includegraphics[width=\linewidth]{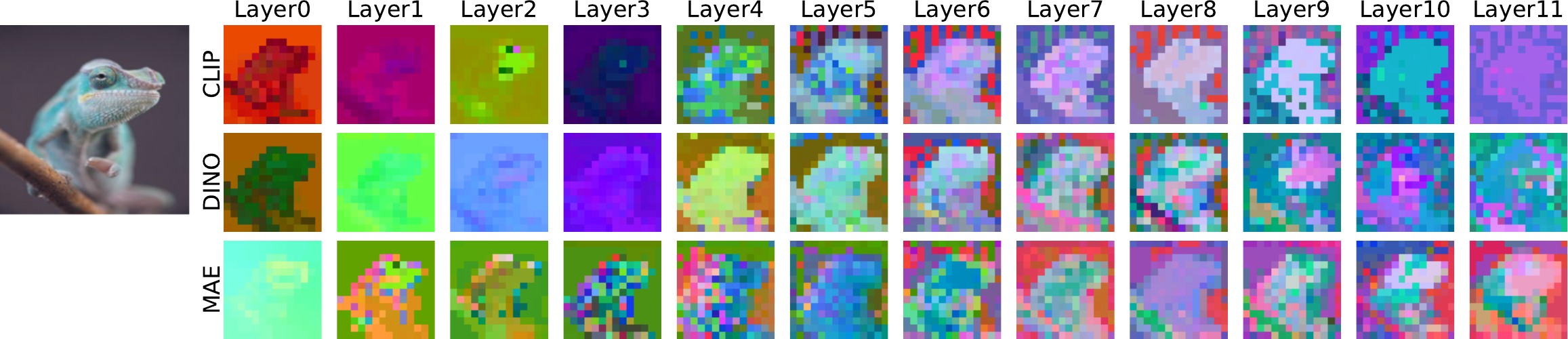}
    \includegraphics[width=\linewidth]{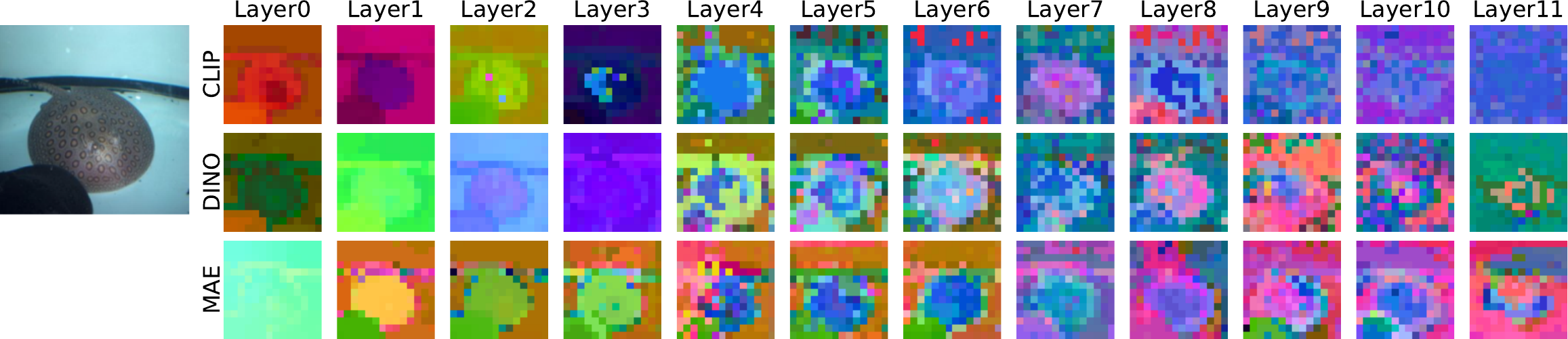}
    \includegraphics[width=\linewidth]{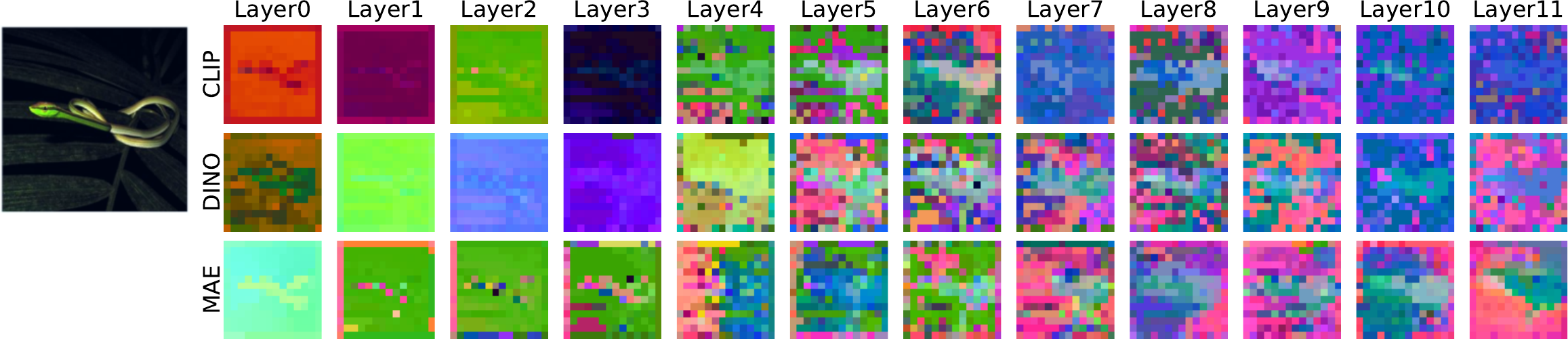}
    \includegraphics[width=\linewidth]{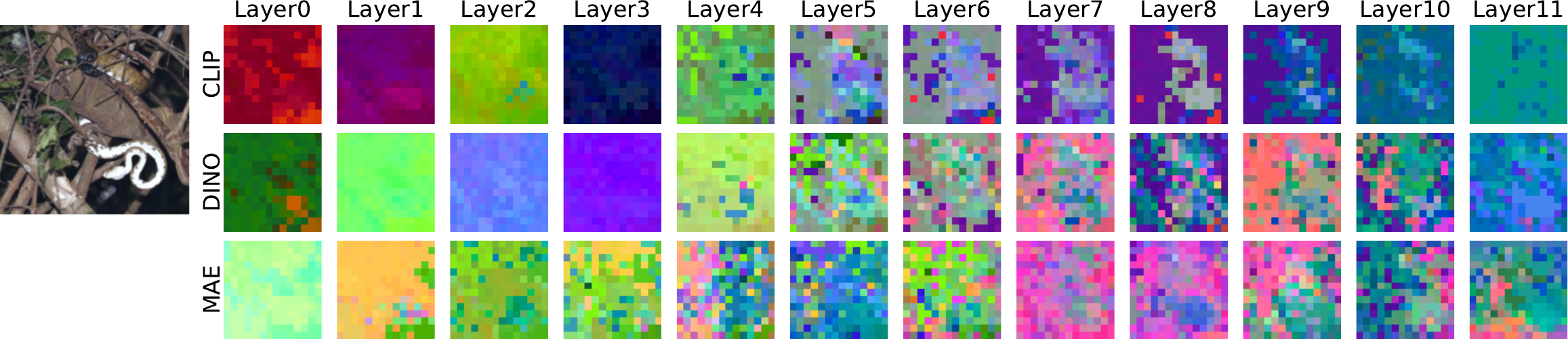}
    \caption{Spectral clustering in the universal channel aligned feature space. The image pixels are colored by our approach AlignedCut, the pixel RGB value is assigned by the 3D spectral-tSNE of the top 20 eigenvectors. The coloring is consistent across all images, layers, and models.}
    \label{fig:eig_tsne_images_app}
\end{figure}

\begin{figure}
    \centering
    \includegraphics[width=\linewidth]{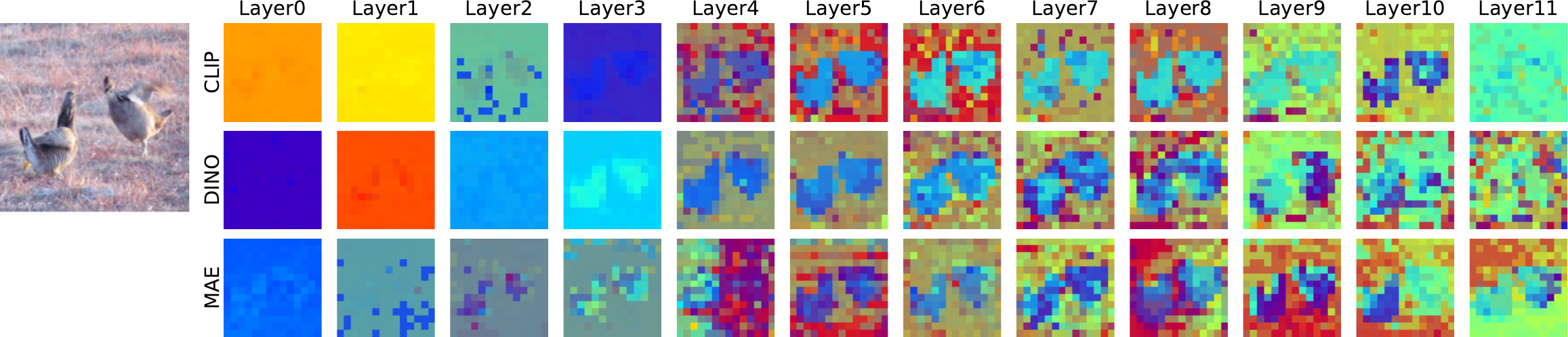}
    \includegraphics[width=\linewidth]{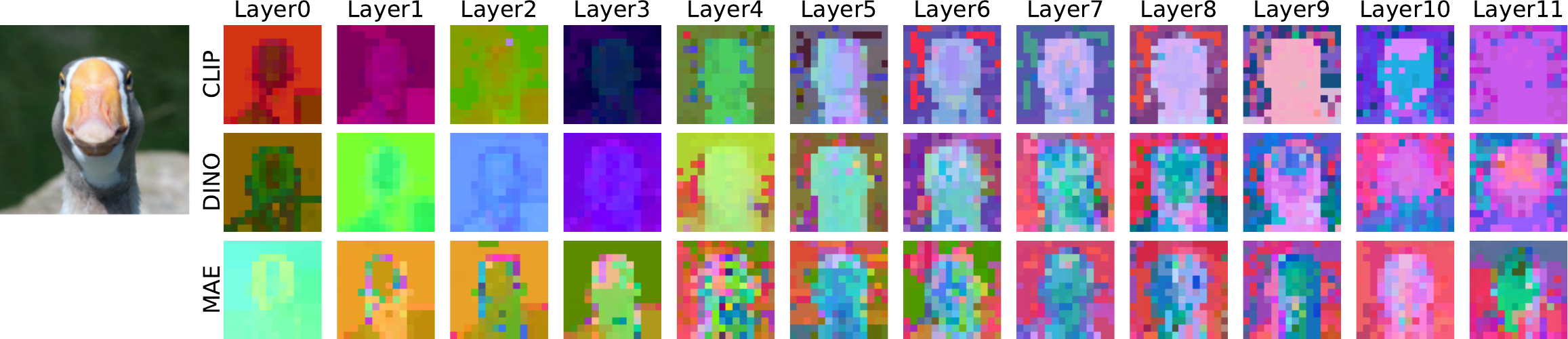}    \includegraphics[width=\linewidth]{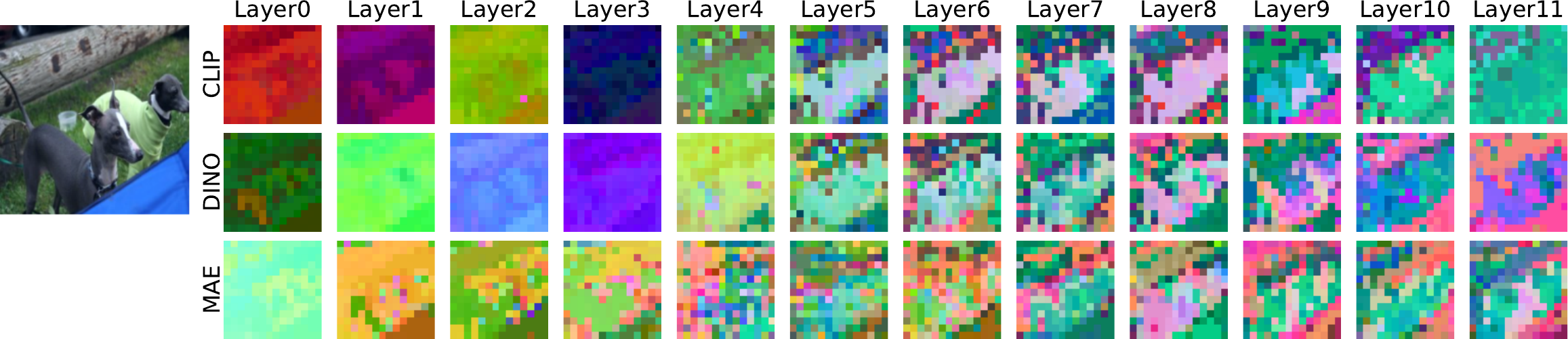}
    \includegraphics[width=\linewidth]{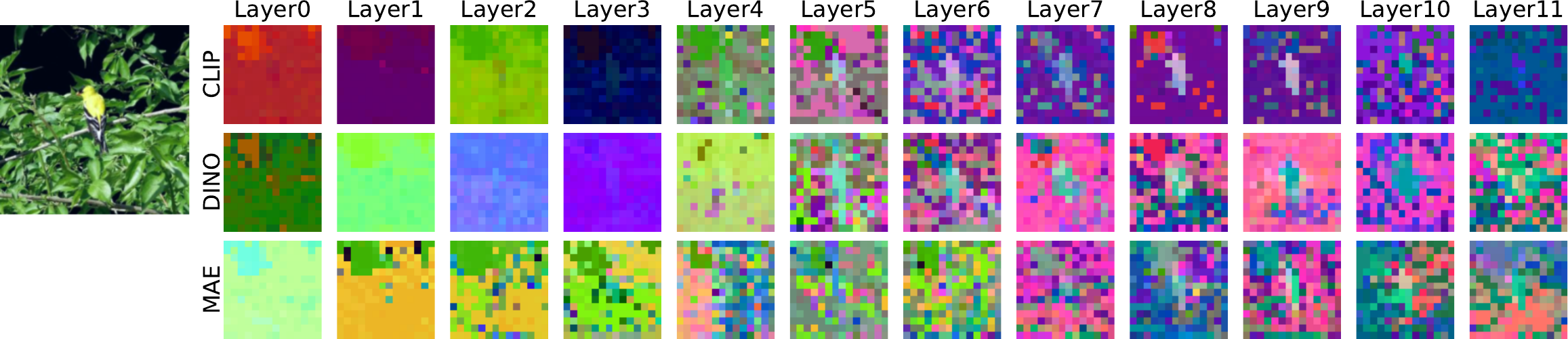}
    \includegraphics[width=\linewidth]{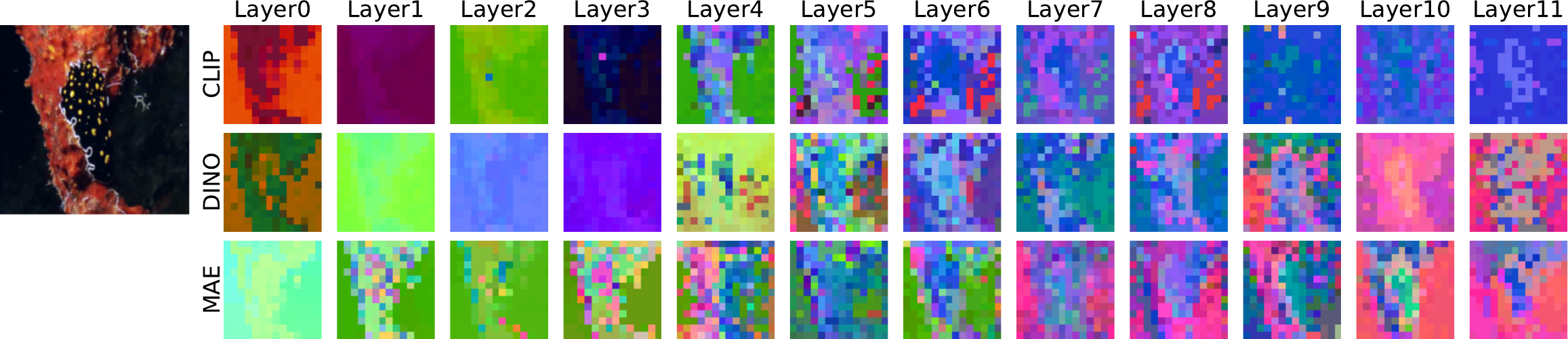}
    \includegraphics[width=\linewidth]{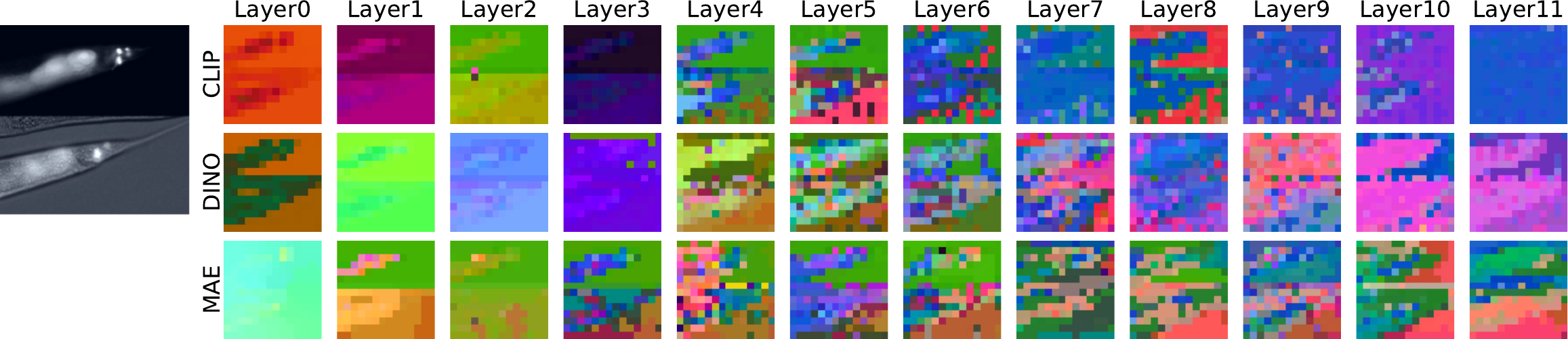}
    \caption{Spectral clustering in the universal channel aligned feature space. The image pixels are colored by our approach AlignedCut, the pixel RGB value is assigned by the 3D spectral-tSNE of the top 20 eigenvectors. The coloring is consistent across all images, layers, and models.}
    \label{fig:eig_tsne_images_app}
\end{figure}

\begin{figure}
    \centering
    \includegraphics[width=\linewidth]{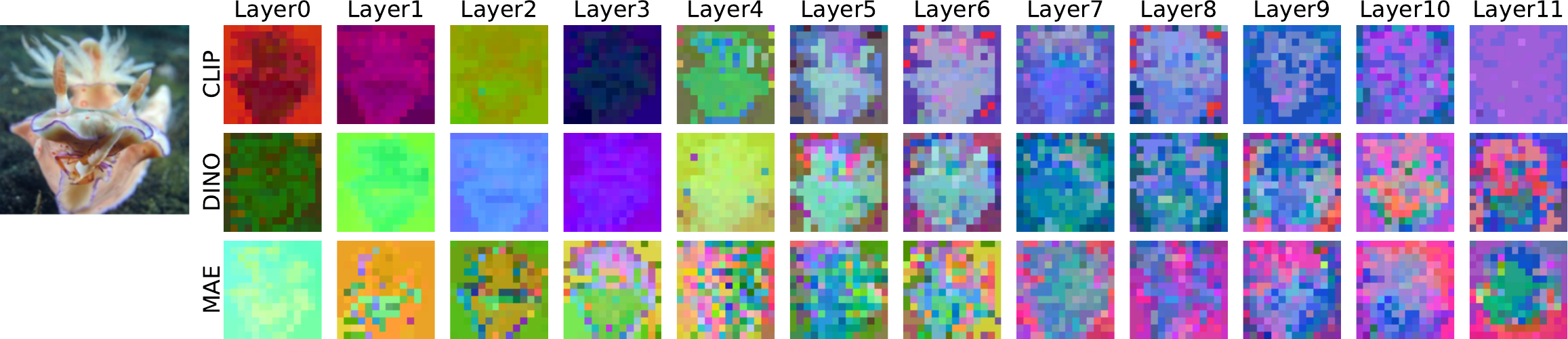}
    \includegraphics[width=\linewidth]{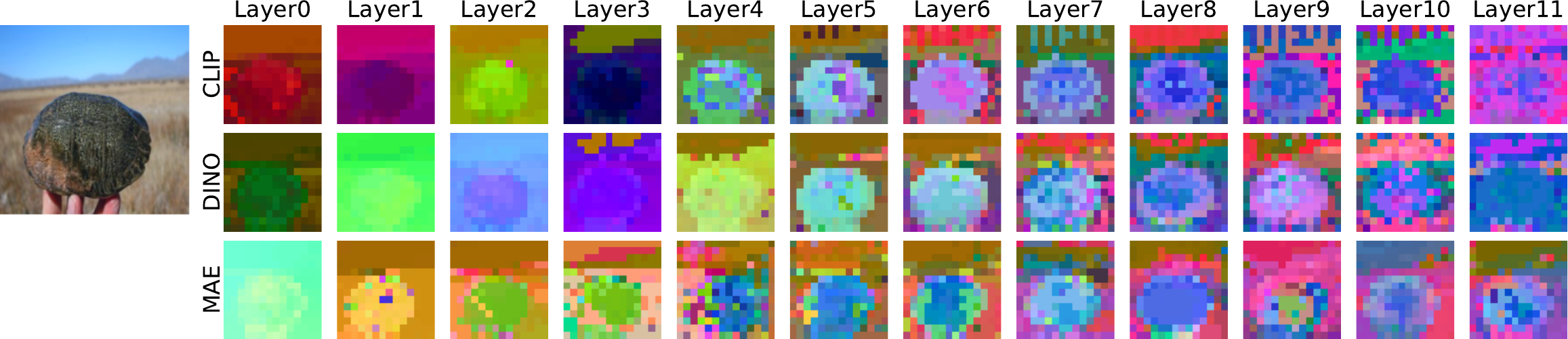}    \includegraphics[width=\linewidth]{appendix_figures/eig_tsne/eig_tsne_578.pdf}
    \includegraphics[width=\linewidth]{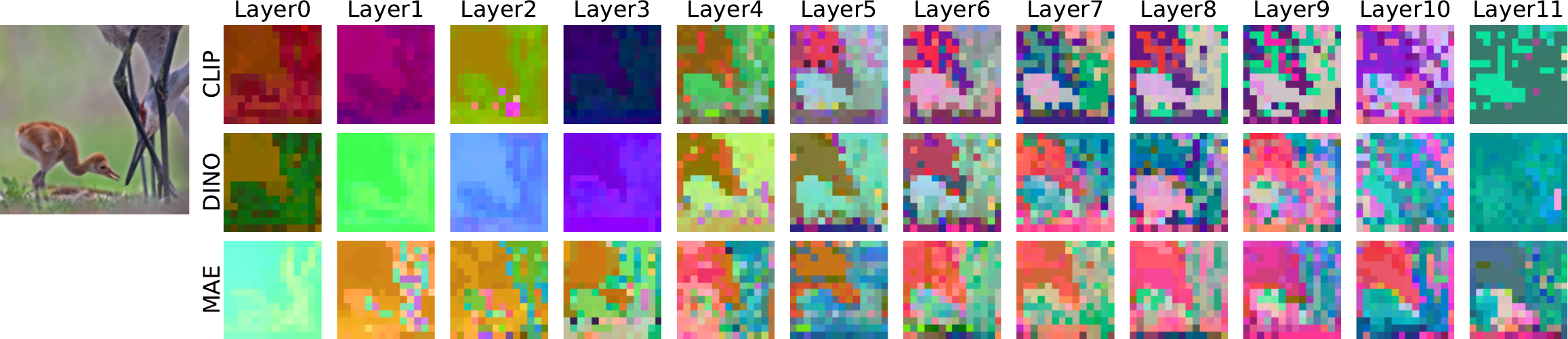}
    \includegraphics[width=\linewidth]{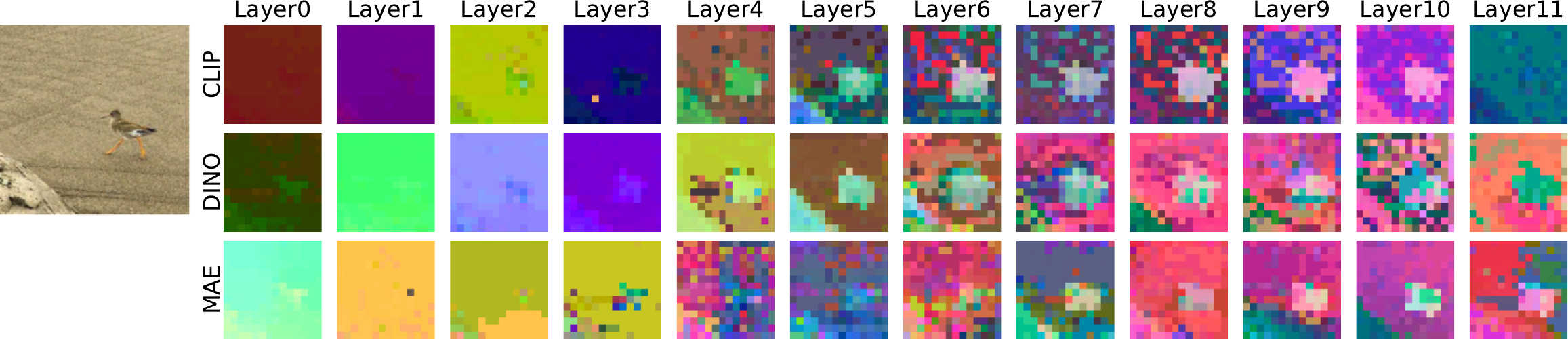}
    \includegraphics[width=\linewidth]{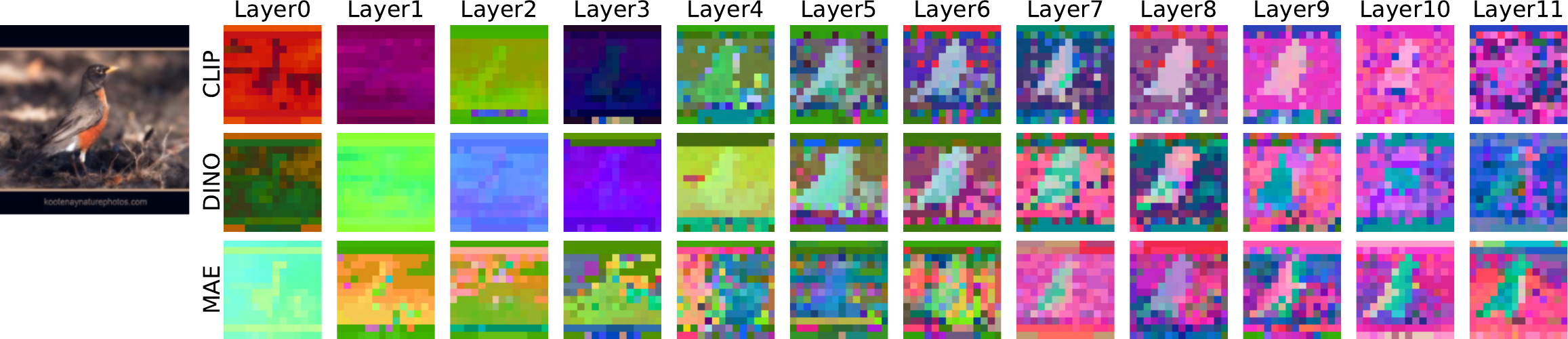}
    \caption{Spectral clustering in the universal channel aligned feature space. The image pixels are colored by our approach AlignedCut, the pixel RGB value is assigned by the 3D spectral-tSNE of the top 20 eigenvectors. The coloring is consistent across all images, layers, and models.}
    \label{fig:eig_tsne_images_app}
\end{figure}

\begin{figure}
    \centering
    \includegraphics[width=\linewidth]{appendix_figures/eig_tsne/eig_tsne_859.pdf}
    \includegraphics[width=\linewidth]{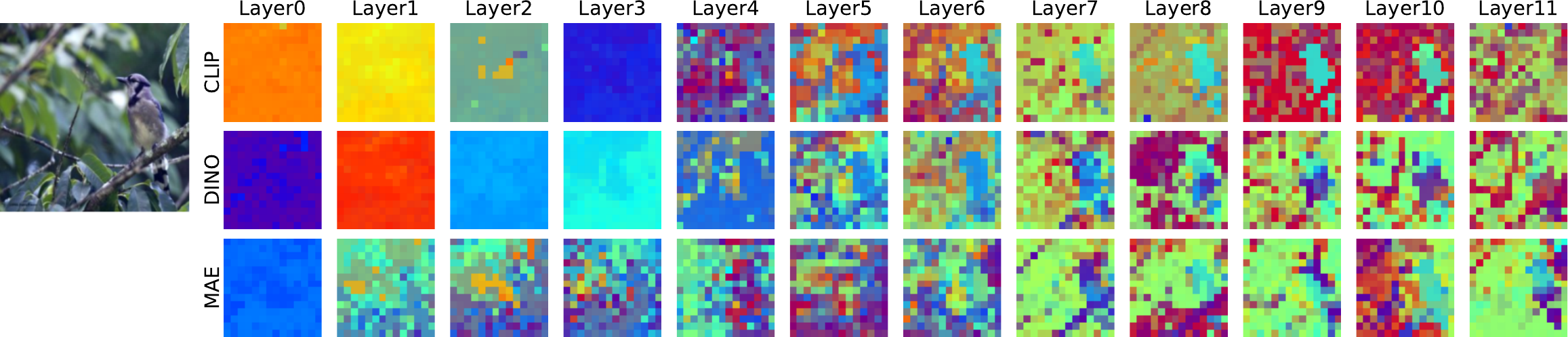}    \includegraphics[width=\linewidth]{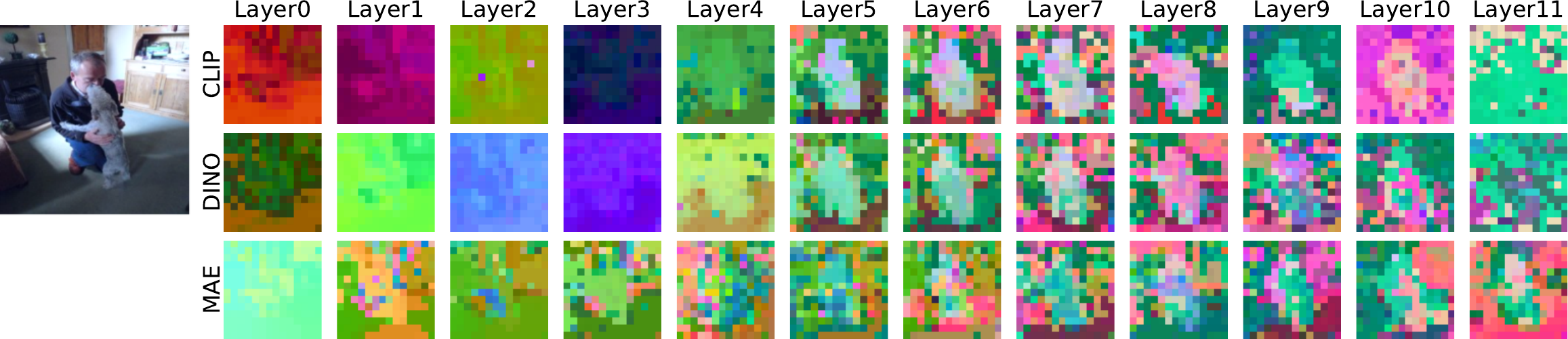}
    \includegraphics[width=\linewidth]{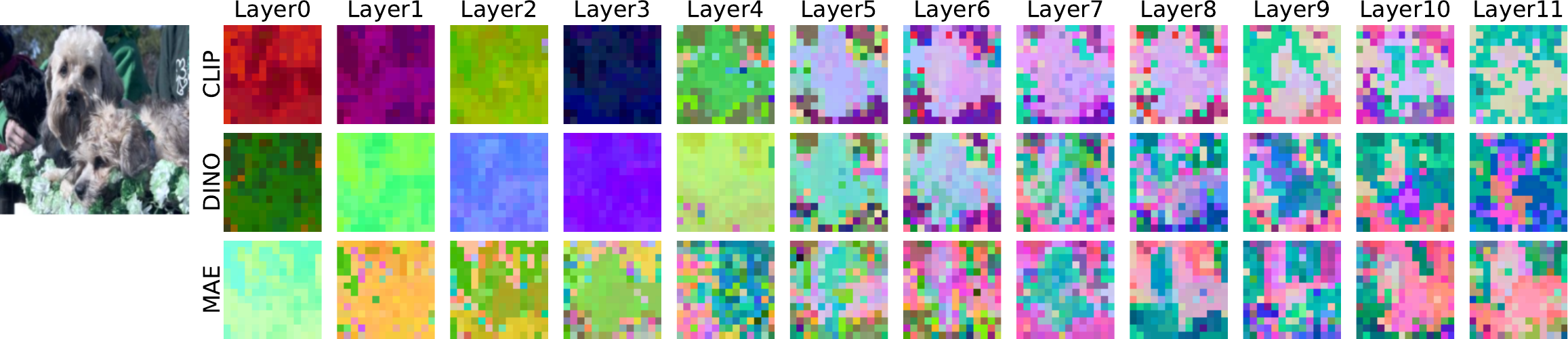}
    \includegraphics[width=\linewidth]{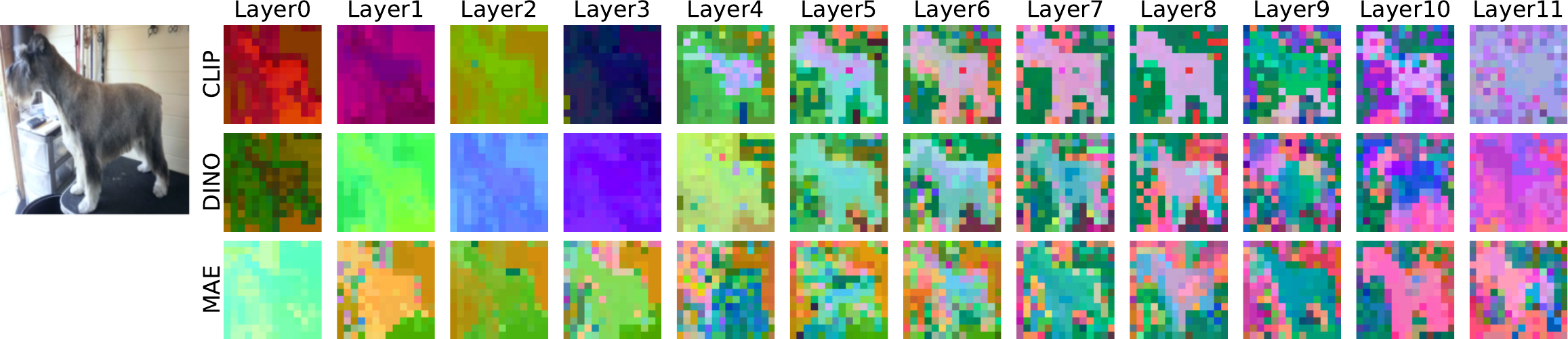}
    \includegraphics[width=\linewidth]{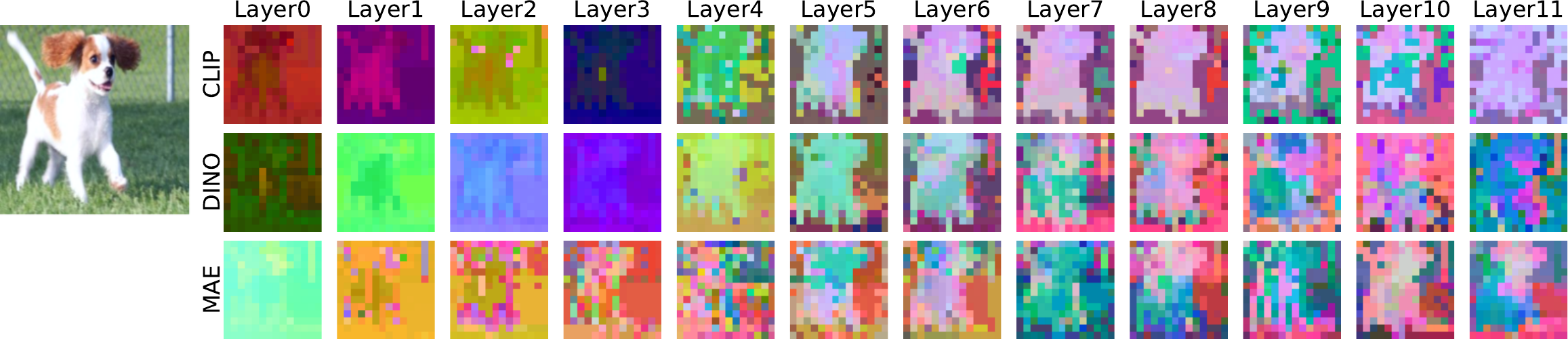}
    \caption{Spectral clustering in the universal channel aligned feature space. The image pixels are colored by our approach AlignedCut, the pixel RGB value is assigned by the 3D spectral-tSNE of the top 20 eigenvectors. The coloring is consistent across all images, layers, and models.}
    \label{fig:eig_tsne_images_app}
\end{figure}

\newpage

\clearpage
\pagestyle{fancy}
\fancyhead{}
\fancyhead[RO,LE]{\textbf{Figure-ground Channel Activation}}

\section{Figure-ground Channel Activation from All Layers and Models}
\label{sec:fgbg}

\begin{figure}[h]
    \centering
    \includegraphics[width=\linewidth]{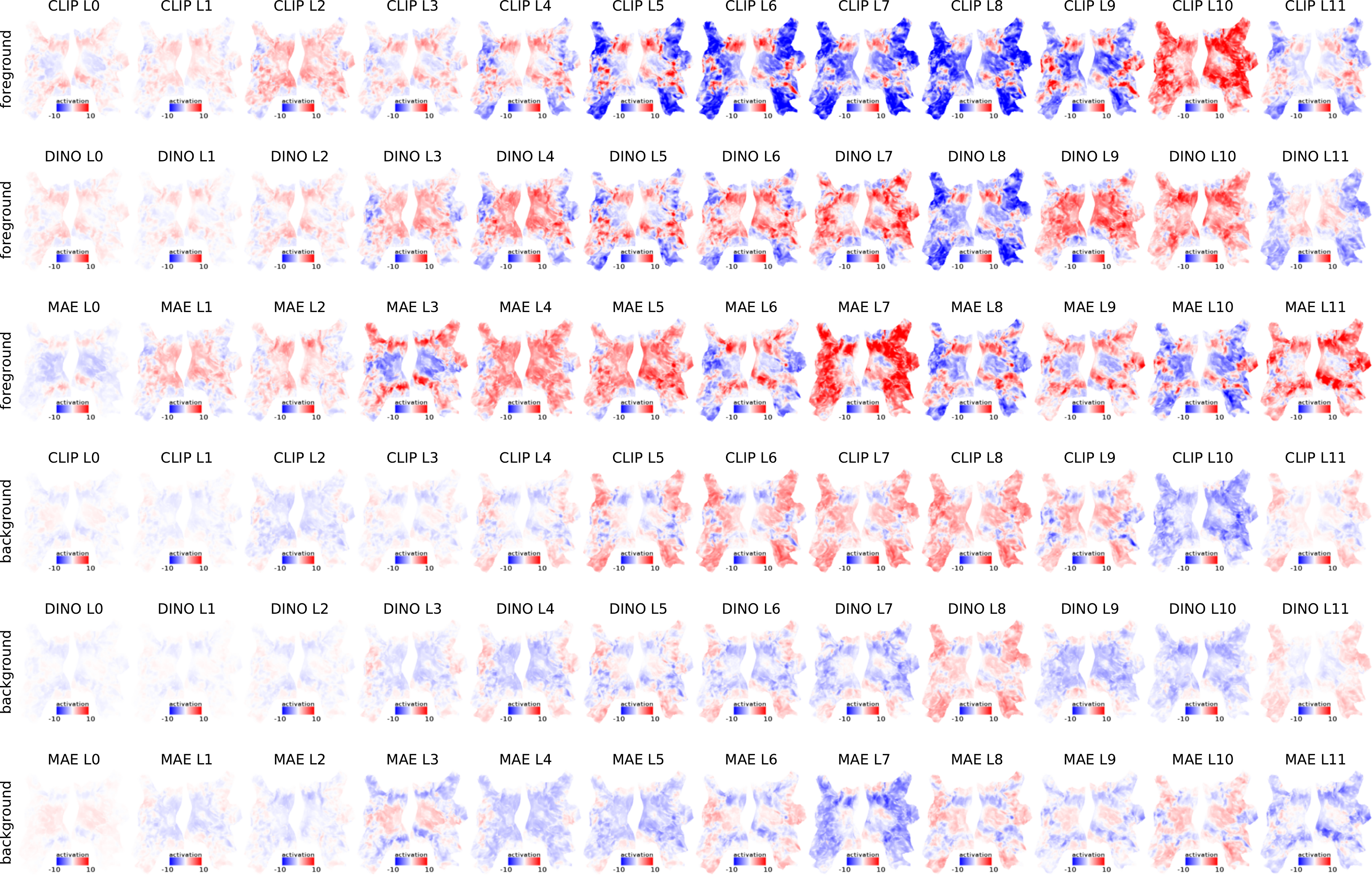}
    \caption{Mean activation of foreground or background pixels at each layer of CLIP, DINO and MAE. Channel activations are linearly transformed to the brain's space. Large absolute activation value means more consistent visual concepts.}
    \label{fig:enter-label}
\end{figure}

\newpage

\clearpage
\pagestyle{fancy}
\fancyhead{}
\fancyhead[RO,LE]{\textbf{Visual Concepts: Category-specific}}

\section{Visual Concepts: Categories }
\label{sec:cate}

\begin{figure}[h]
    \centering
    \includegraphics[width=\linewidth]{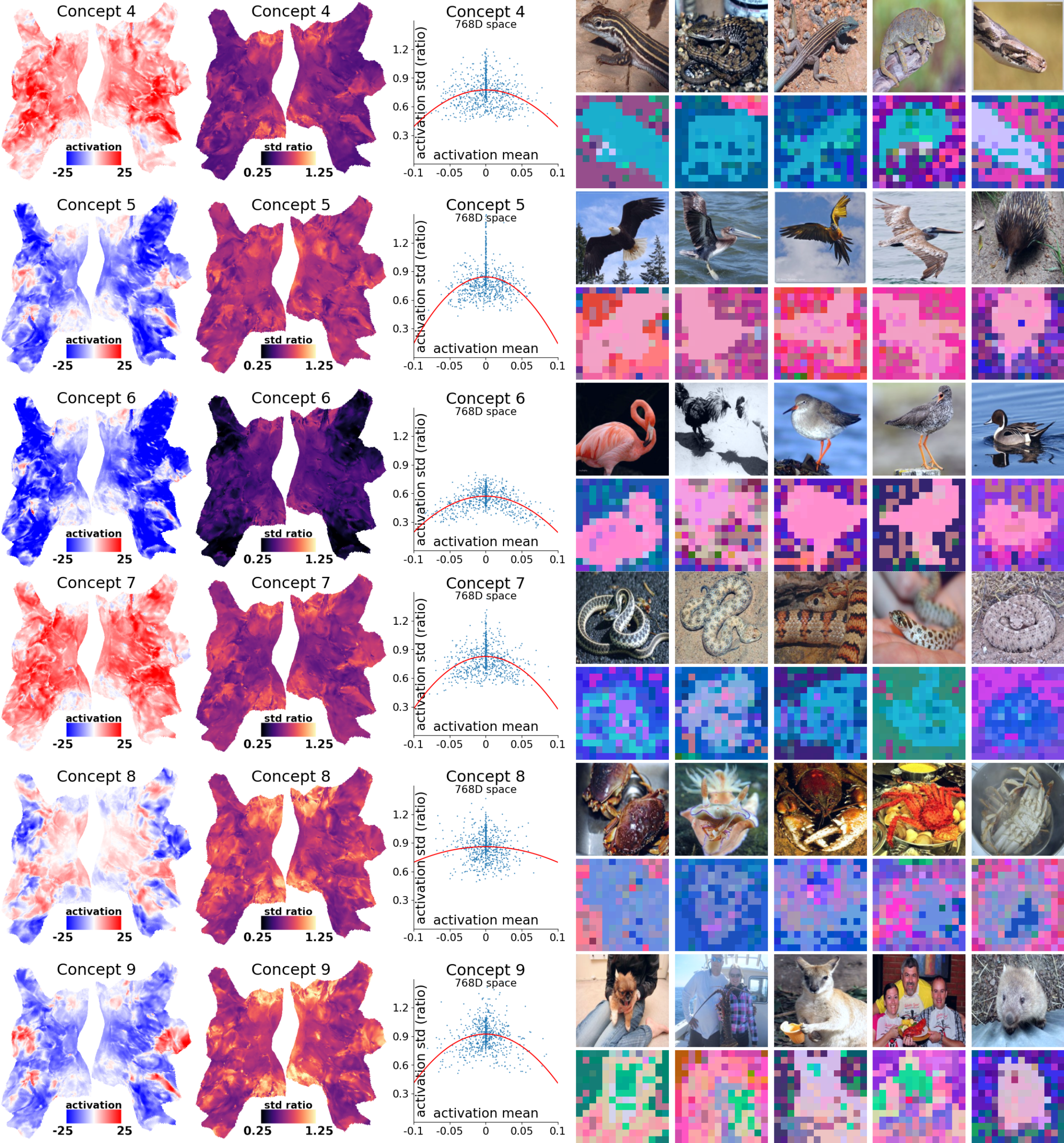}
    \caption{ Category visual concepts in CLIP Layer 9. \textbf{\textit{Left:}} Mean activation of all pixels within an Euclidean sphere centered at the visual concept in the 3D spectral-tSNE space; the concepts activate different brain regions. \textbf{\textit{Middle:}} The standard deviation negatively correlates with absolute mean activations. \textbf{\textit{Right:}} Spectral clustering, colored by 3D spectral-tSNE of the top 20 eigenvectors.
    }
    \label{fig:enter-label}
\end{figure}

\newpage

\clearpage
\pagestyle{fancy}
\fancyhead{}
\fancyhead[RO,LE]{\textbf{2D Spectral-tSNE Space Information Flow}}

\section{Layer-to-Layer Feature Computation Flow in 2D spectral-tSNE space}
\label{sec:2d}

\begin{figure}[h]
    \centering
    \includegraphics[width=\linewidth]{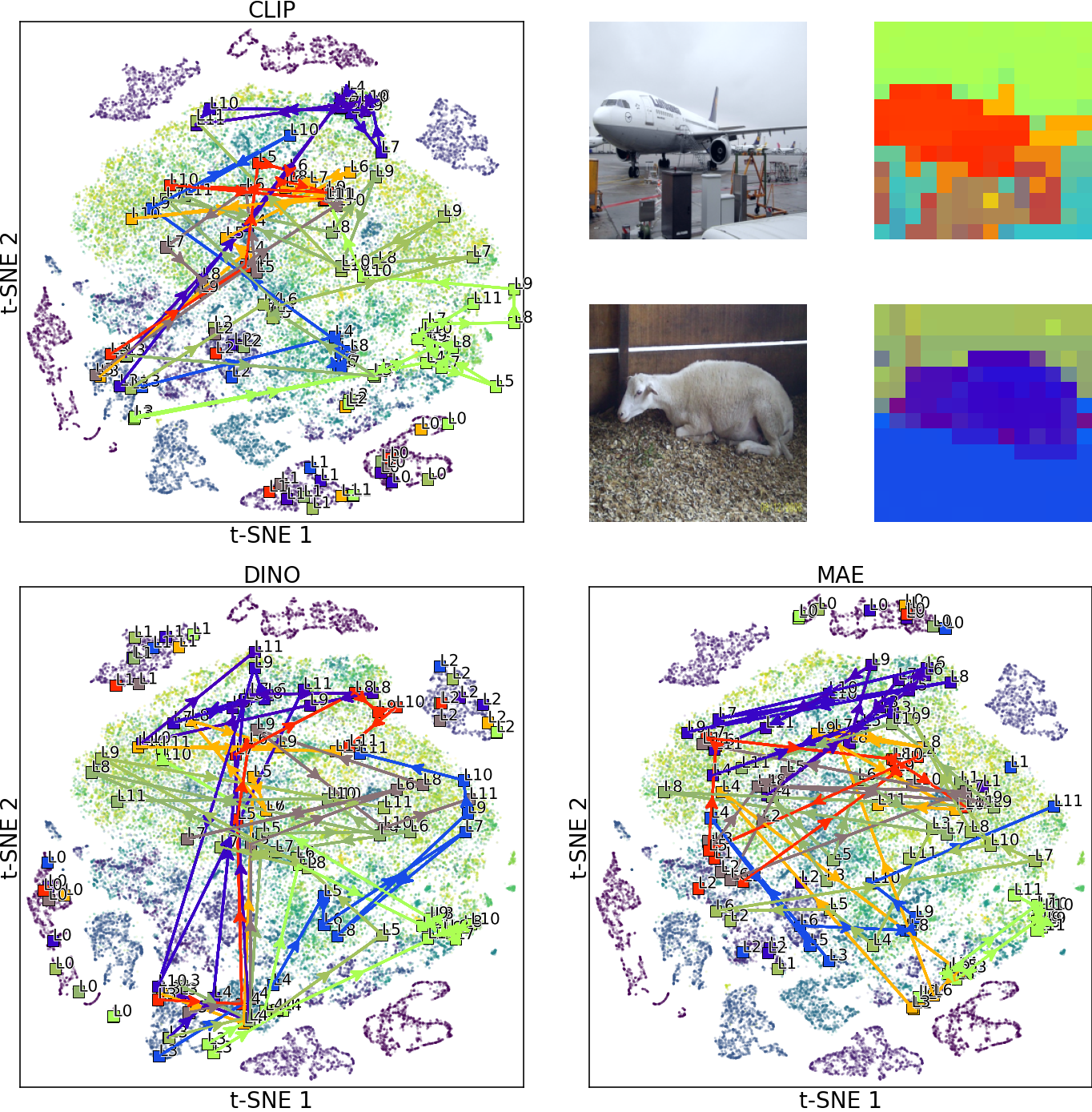}
    \caption{Trajectory of feature progression in from layer to layer, in the 2D spectral-tSNE space. Arrows displayed for 10 randomly sampled example pixels. \textbf{\textit{Top Right:}} Pixels are colored by unsupervised segmentation.}
    \label{fig:enter-label}
\end{figure}

\begin{figure}[t]
    \centering
    \includegraphics[width=\linewidth]{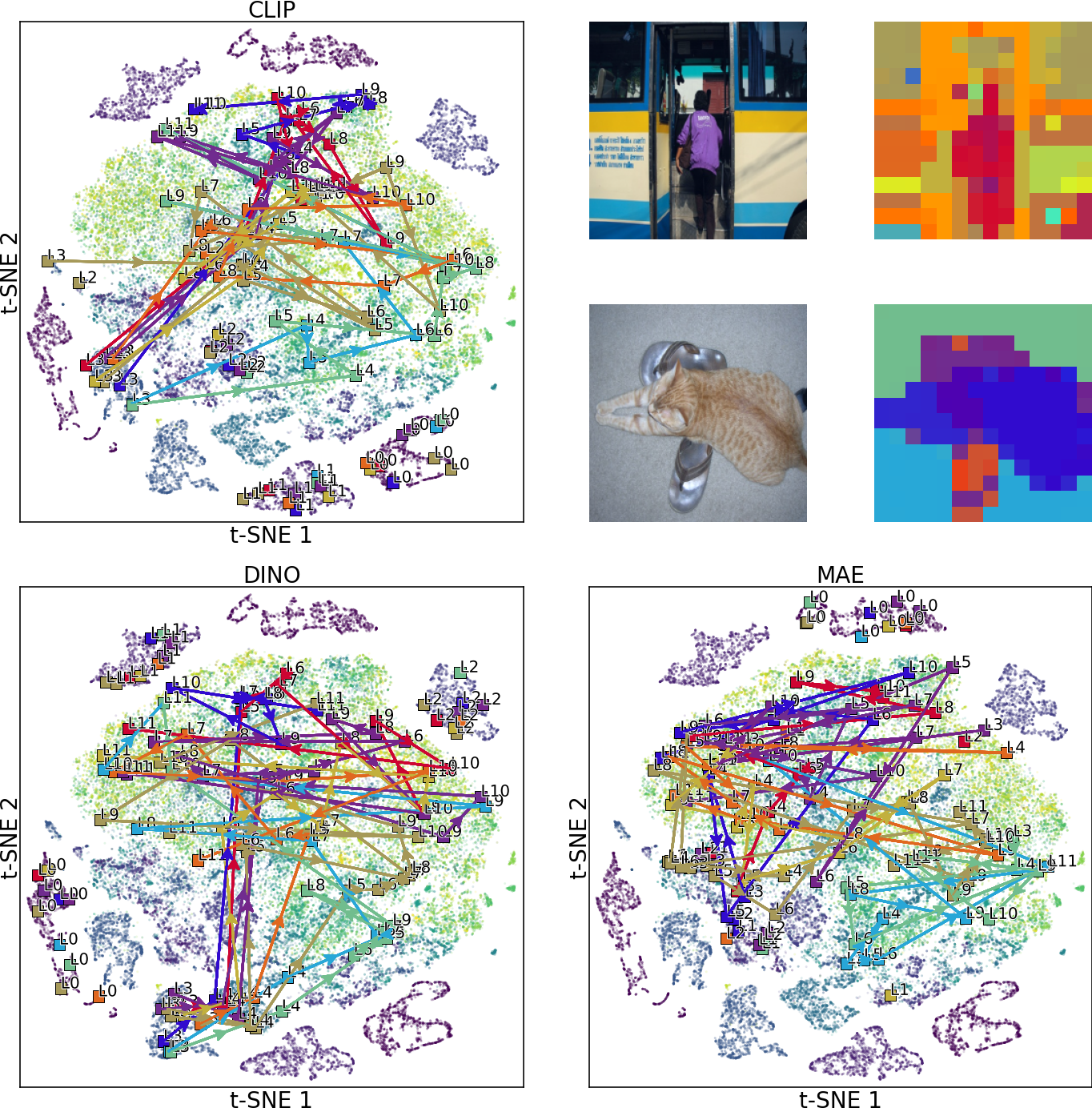}
    \caption{Trajectory of feature progression in from layer to layer, in the 2D spectral-tSNE space. Arrows displayed for 10 randomly sampled example pixels. \textbf{\textit{Top Right:}} Pixels are colored by unsupervised segmentation.}
    \label{fig:enter-label}
\end{figure}

\begin{figure}[t]
    \centering
    \includegraphics[width=\linewidth]{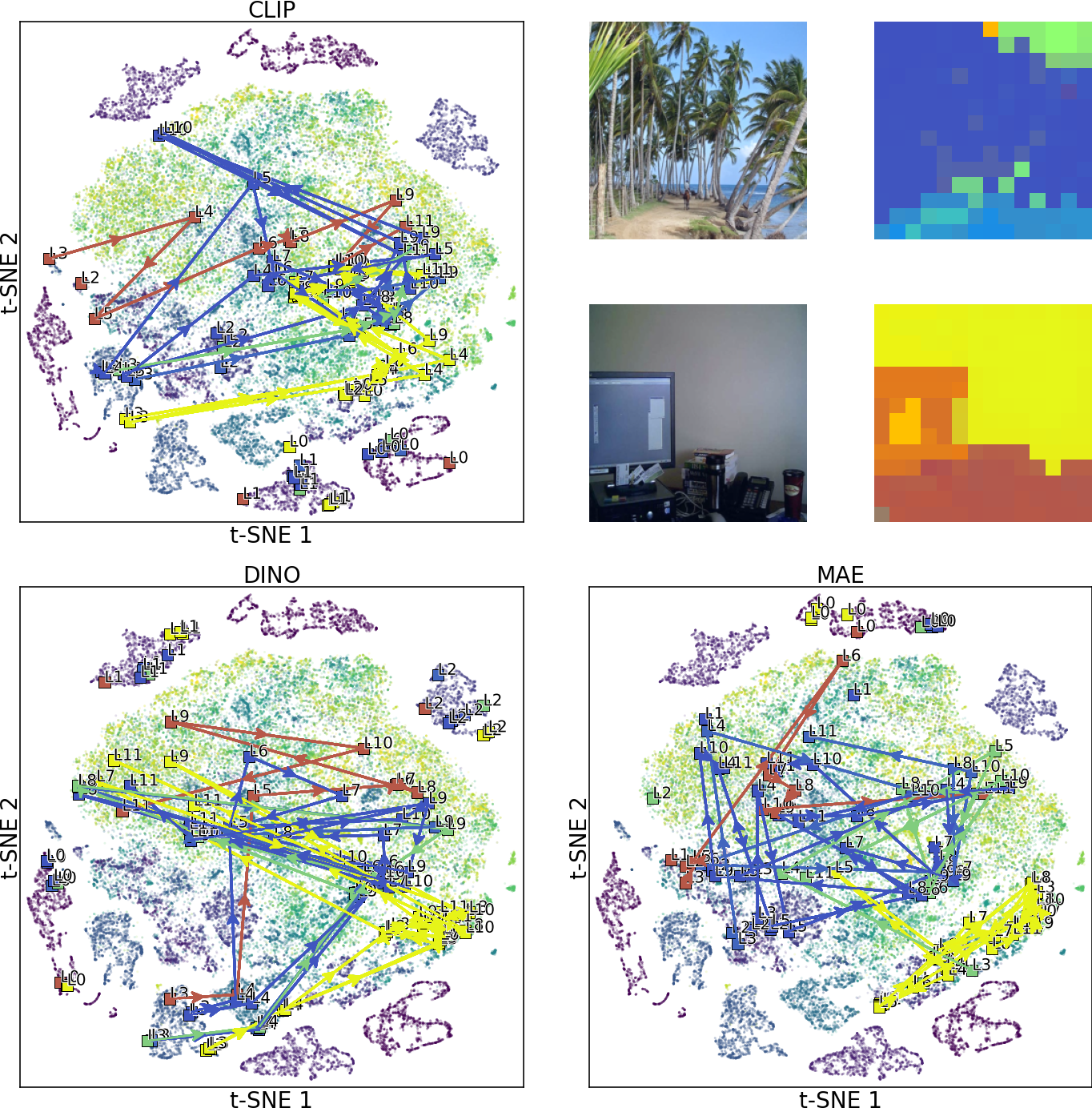}
    \caption{Trajectory of feature progression in from layer to layer, in the 2D spectral-tSNE space. Arrows displayed for 10 randomly sampled example pixels. \textbf{\textit{Top Right:}} Pixels are colored by unsupervised segmentation.}
    \label{fig:enter-label}
\end{figure}

\begin{figure}[t]
    \centering
    \includegraphics[width=\linewidth]{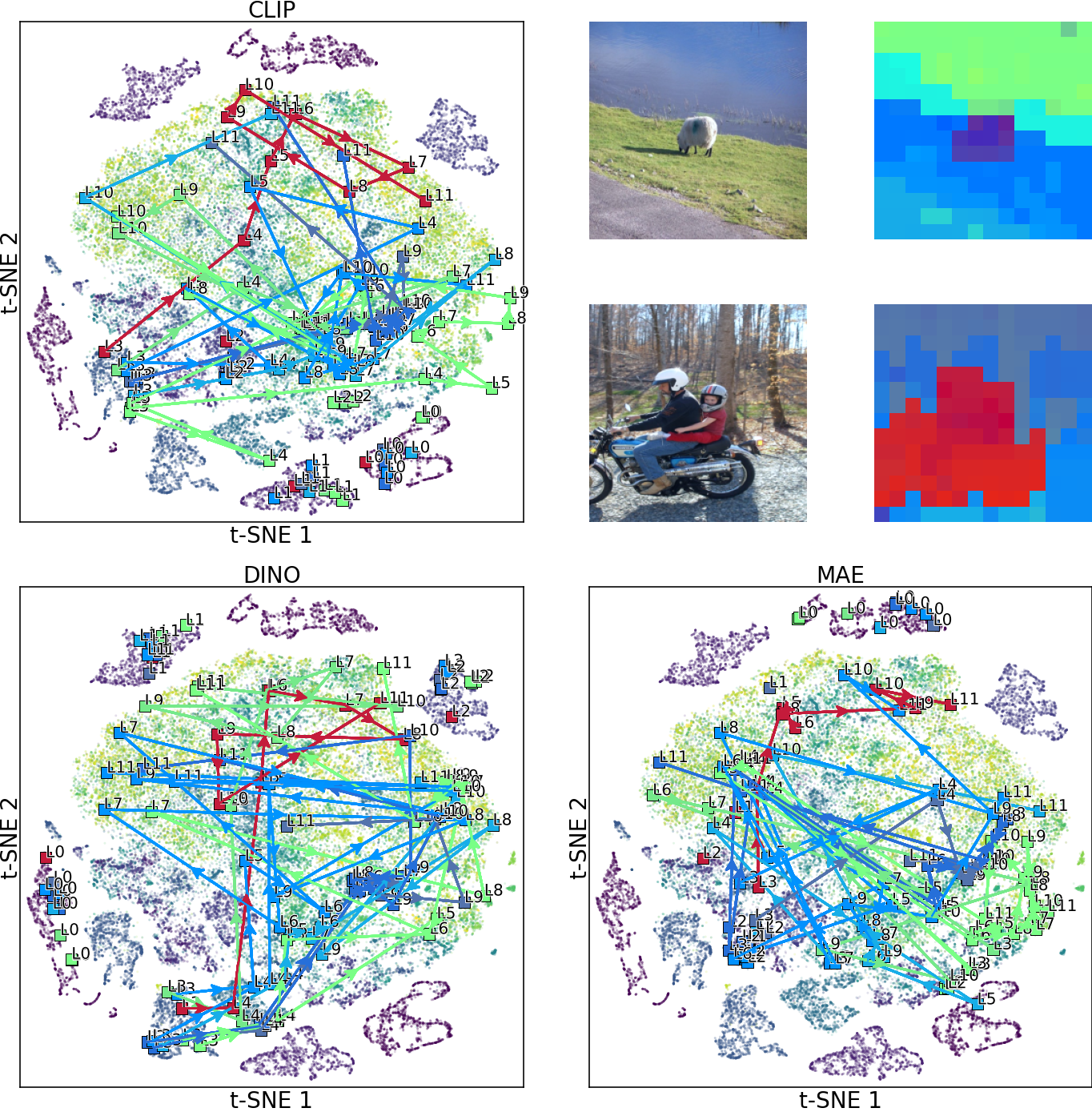}
    \caption{Trajectory of feature progression in from layer to layer, in the 2D spectral-tSNE space. Arrows displayed for 10 randomly sampled example pixels. \textbf{\textit{Top Right:}} Pixels are colored by unsupervised segmentation.}
    \label{fig:enter-label}
\end{figure}

\begin{figure}[t]
    \centering
    \includegraphics[width=\linewidth]{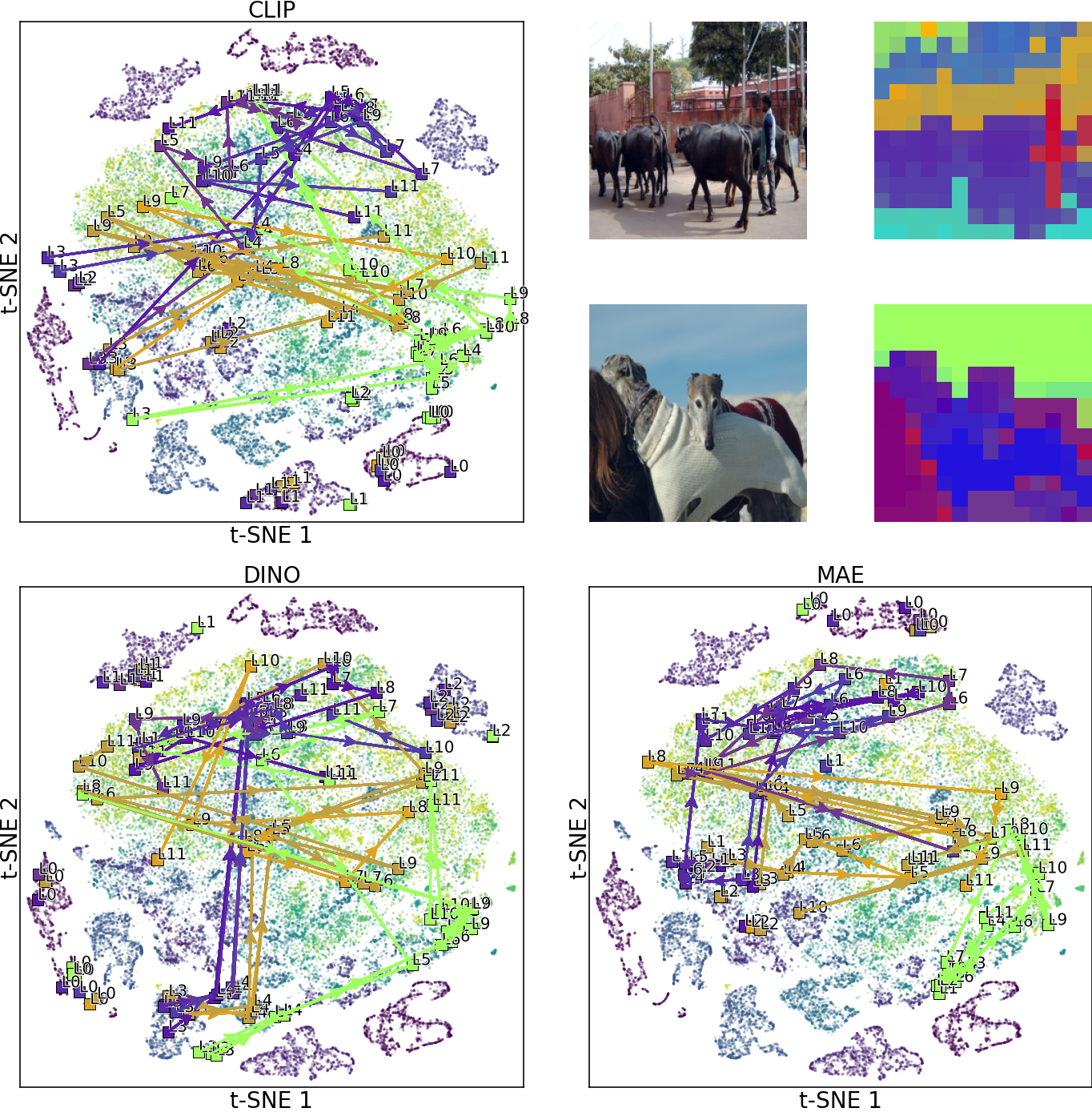}
    \caption{Trajectory of feature progression in from layer to layer, in the 2D spectral-tSNE space. Arrows displayed for 10 randomly sampled example pixels. \textbf{\textit{Top Right:}} Pixels are colored by unsupervised segmentation.}
    \label{fig:enter-label}
\end{figure}





\end{document}


\maketitle


\appendix


\section{Appendix overview}

\begin{enumerate}
\item \Cref{sec:app_brain_roi} summarizes background of brain ROIs. 
\item \Cref{sec:app_implementation} is implementation details 
  \begin{enumerate}[label*=\arabic*.]
    \item Additional regularization terms
    \item Brain encoding model training loss function
    \item Unsupervised segmentation evaluation pipeline
    \item Nystrom-like approximation for t-SNE
  \end{enumerate}
\item \Cref{sec:app_3d} lists more image examples from the 3D spectral-tSNE.
\item \Cref{sec:fgbg} lists figure-ground channel activation for every model and layer.
\item \Cref{sec:cate} lists more example category-specific visual concepts.
\item \Cref{sec:2d} lists more example pixels from the 2D spectral-tSNE information flow.
\end{enumerate}

\newpage

\section{Brain Region Background Knowledge}
\label{sec:app_brain_roi}

\begin{figure}[ht]

    \centering
    \captionsetup{type=figure}
    \includegraphics[width=\linewidth]{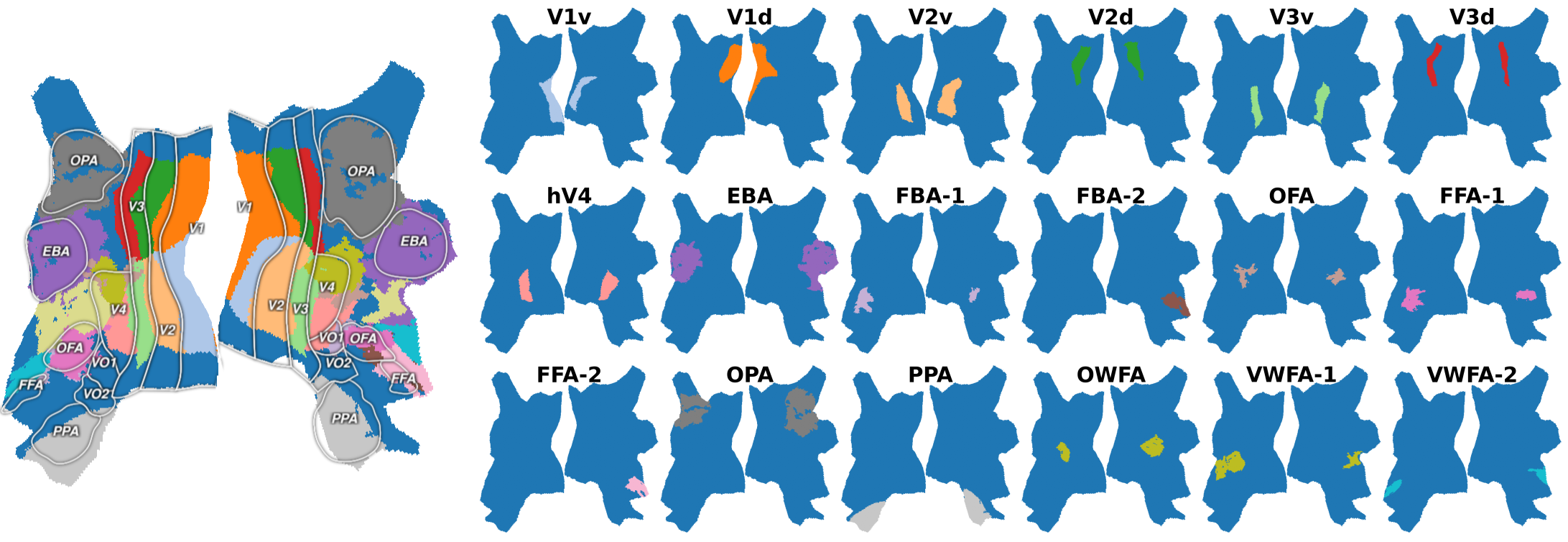}
    \captionof{figure}{\textbf{Brain Region of Interests (ROIs)}. V1v: ventral stream, V1d: dorsal stream.}
    \label{fig:supp_rois}

\vspace{4mm}

    \captionof{table}{Known function and selectivity of brain region of interests (ROIs).}
    \resizebox{0.99\linewidth}{!}{
    \begin{tblr}{
      hline{1,3} = {-}{0.08em},
    }
    \textbf{ROI name}  \quad\quad\quad & V1 V2 V3 & V4 \quad & EBA FBA & OFA FFA & OPA \quad & PPA \quad & OWFA VWFA \\
    \textbf{Known Function/Selectivity} \quad\quad & primary visual & mid-level & body & face & navigation & scene & words \\
    \end{tblr}
    }
    \label{tab:supp_rois}
\vspace{10mm}

\end{figure}

This section briefly summarizes the known functions of key brain regions of interest (ROIs). \Cref{fig:supp_rois} provides an overview of these brain ROIs. \Cref{tab:supp_rois} lists the known functions and selectivities for each ROI.

In brief, V1 to V3 are the primary visual stream, which is further divided into ventral (lower) and dorsal (upper) streams. V4 is a mid-level visual area. EBA (extrastriate body area) and FBA (fusiform body area) are selectively responsive to bodies, while FFA (fusiform face area) and OFA (occipital face area) show selectivity for faces. OWFA (occipital word form area) and VWFA (visual word form area) are selective for written words. PPA (parahippocampal place area) exhibits selectivity for scenes and places, and OPA (occipital place area) is involved in navigation and spatial reasoning.

Visual information processing in the brain follows a hierarchical, feedforward organization. Beginning in the primary visual cortex (V1) and progressing through higher visual areas like V2, V3, and V4, neurons exhibit increasingly large receptive fields and represent increasingly abstract visual concepts. While neurons in V1 encode low-level features like edges and orientations within a small portion of the visual field, neurons in V4 synthesize more complex patterns and object representations across a larger area of the visual input.

\newpage
\section{Implementation Details}
\label{sec:app_implementation}

\subsection{Additional Regularization for Channel Align Transformation}

Additional Regularization are added to the channel align transform to ensure good properties of the aligned features: 1) zero-centered, 2) small covariance between channels, and 3) focal loss.

\textbf{Zero-centered regularization.}
We did not apply z-score normalization to the extracted features; instead, we added a regularization term to ensure the transformed features are zero-centered. Recall that the channel-aligned transformed feature \(\bm{V}' \in \mathbb{R}^{M \times D'}\), where \(M\) is the number of data points and \(D'\) is the hidden dimension. The zero-center loss is defined as:
\begin{equation}
    \mathcal{L}_{\mathrm{zero}} = \frac{1}{D'} \frac{1}{M} \sum_{i\leq M, j\leq D'} v'_{ij}
\end{equation}
\textbf{Covariance regularization.}
We used the covariance loss to minimize the off-diagonal elements in the covariance matrix of the transformed feature $C(\bm{V}')$, aiming to bring them close to $\mathbf{0}$. Recall that channel align transformed feature $\bm{V}' \in \mathbb{R}^{M \times D'}$, where $M$ is number of data, $D'$ is the hidden dimension. The covariance loss is defined as:
\begin{equation}
    \mathcal{L}_{\mathrm{cov}} = \frac{1}{D'} \sum_{i \neq j}[C(\bm{V}')]_{i, j}^2, \text{ where } C(\bm{V}') = \frac{1}{M-1} \sum_{i=1}^M\left(v'_i-\bar{v'}\right)\left(v'_i-\bar{v'}\right)^T, \bar{v'}=\frac{1}{M} \sum_{i=1}^M v'_i.
\end{equation}
\textbf{Focal Loss.}
\citet{lin_focal_2017} introduced focal loss, which dynamically assigns smaller weights to the loss function for hard-to-classify classes. In our scenario, we apply spectral clustering on the affinity matrix \(\bm{A}_a \in \mathbb{R}^{M \times M}\) after performing the channel alignment transform, where \(M\) represents the number of data points. Due to the characteristics of spectral clustering, disconnected edges play a more critical role than connected edges. Adding an edge between disconnected clusters significantly reshapes the eigenvectors, while adding edges to connected clusters has only a minor impact. Therefore, we aim to assign larger weights to disconnected edges in the loss function:
\begin{equation}
\begin{aligned}
    \mathcal{L}_{eigen} &= \| ( \bm{X}_{b} \bm{X}_{b}^T - \bm{X}_{a} \bm{X}_{a}^T ) * \exp(-\bm{A}_b) \|
\end{aligned}
\end{equation}
where \(\bm{A}_b \in \mathbb{R}^{M \times M}\) is the affinity matrix before the channel alignment transform, element wise dot-product to \(\exp(-\bm{A}_b)\) assigned larger wights for disconnected edges. \(\bm{X}_{b} \in \mathbb{R}^{M\times C}, \bm{X}_{a} \in \mathbb{R}^{M\times C}\) are eigenvectors before and after channel align transform, respectively.

\subsection{Brain Encoding Model Training Loss}

Let \(\bm{Y} \in \mathbb{R}^{1 \times N}\) represent the brain prediction target, where \(N\) is the number of flattened 3D brain voxels, and the \(1\) indicates that each voxel's response is a scalar value. \(\bm{\hat{Y}}\) is the model's predicted brain response. The brain encoding model training loss is the L1 loss:
\begin{equation}
\begin{aligned}
    \mathcal{L}_{brain} &= \| \bm{Y} - \bm{\hat{Y}} \|
\end{aligned}
\end{equation}
\subsection{Total Training Loss}

The total training loss is a combination of the following components: 1) brain encoding model loss, 2) eigen-constraint regularization, 3) zero-centered regularization, and 4) covariance regularization:
\begin{equation}
\begin{aligned}
    \mathcal{L} &= \mathcal{L}_{brain} + \lambda_{eigen} \mathcal{L}_{eigen} + \lambda_{zero} \mathcal{L}_{zero} + \lambda_{cov} \mathcal{L}_{cov}
\end{aligned}
\end{equation}
where we set $\lambda_{eigen} = 1$, $\lambda_{zero} = 0.01$, $\lambda_{cov} = 0.01$.

\newpage
\subsection{Oracle-based Unsupervised Segmentation Evaluation Pipeline}

Our unsupervised segmentation pipeline aims to benchmark and compare the performance across each single layer of the CLIP model. The evaluation pipeline is oracle-based:

\quad 1. Apply spectral clustering jointly across all images, taking the top 10 eigenvectors.

\quad 2. For each class of object (plus one background class), use ground-truth labels from the dataset to mask out the pixels and their eigenvectors, and then use the mean of the eigenvectors to define a center for each class.

\quad 3. Compute the cosine similarity of each pixel to all class centers.

\quad 4. For each pixel, if the maximum similarity to all classes is less than a threshold value, assign this pixel to the background class.

\quad 5. Assign pixels (with a similarity greater than the threshold value) to the class with the maximum similarity.

There's one hyper-parameter, the threshold value that requires different optimal value for each layer of CLIP. To ensure a fair comparison across all layers, the threshold value is grid-searched from 10 evenly spaced values between 0 and 1, the maximum mIoU score in the grid search is taken for each layer.

\subsection{Nystrom-like approximation for t-SNE}
To visualize the eigenvectors, we applied t-SNE to the eigenvectors $\bm{X} \in \mathbb{R}^{M \times C}$, where the number of data points $M$ span the product space of models, layers, pixels, and images. Due to the enormous size of $M = 7\mathrm{e}{+6}$ nodes, t-SNE suffered from complexity scaling issues. We again applied our Nystrom-like approximation to t-SNE, with sub-sampled $m = 10\mathrm{e}{+4}$ nodes and KNN $K=1$.

It's worth noting that, since the non-linear distance adjustment in t-SNE, it's crucial to use only one nearest neighbor $K=1$ for t-SNE.

\subsection{Computation Resource}
All of our experiments are performed on one consumer-grade RTX 4090 GPU. The brain encoding model training took 3 hours on 4GB of VRAM, spectral clustering eigen-decomposition on large graph took 10 minutes on 10GB of VRAM and 60GB of CPU RAM.

\subsection{Code Release}
Our code will be publicly released upon publication.

\newpage

\clearpage
\pagestyle{fancy}
\fancyhead{}
\fancyhead[RO,LE]{\textbf{3D Spectral-tSNE}}

\section{3D spectral-tSNE}
\label{sec:app_3d}

\begin{figure}[h]
    \centering
    \includegraphics[width=\linewidth]{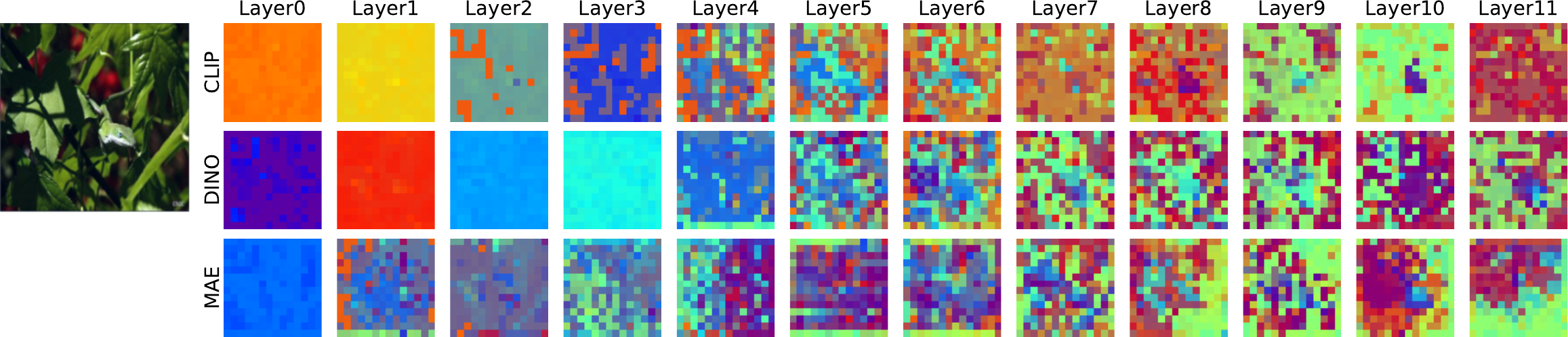}
    \includegraphics[width=\linewidth]{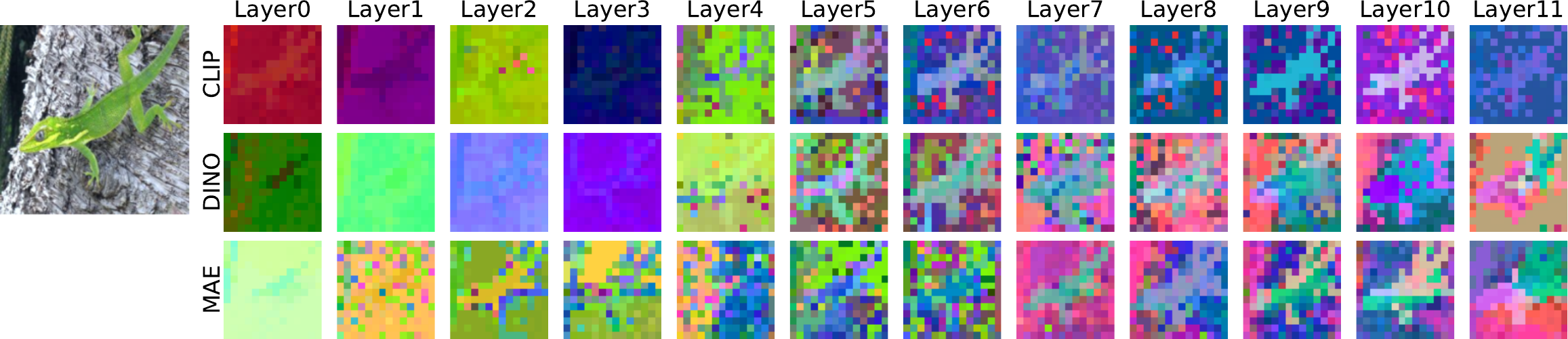}    \includegraphics[width=\linewidth]{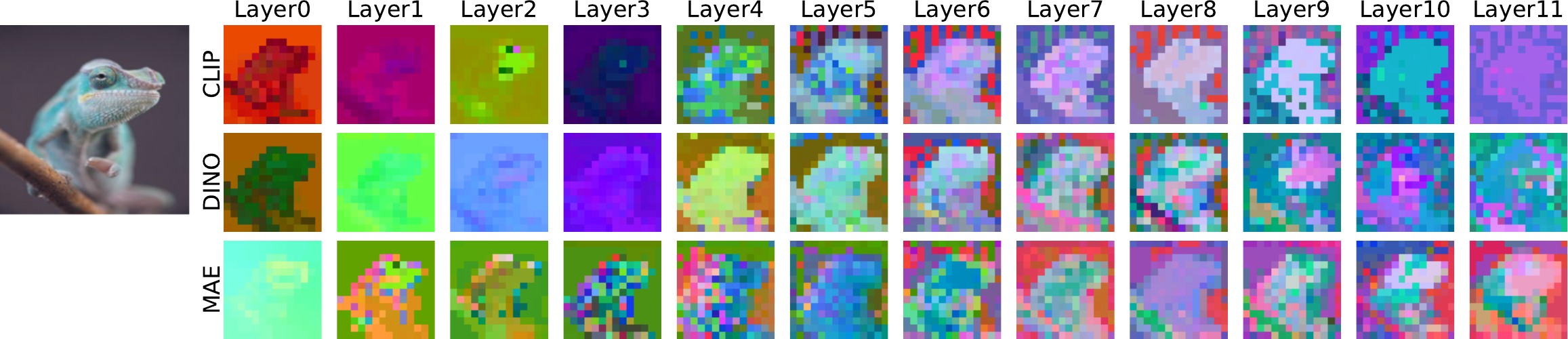}
    \includegraphics[width=\linewidth]{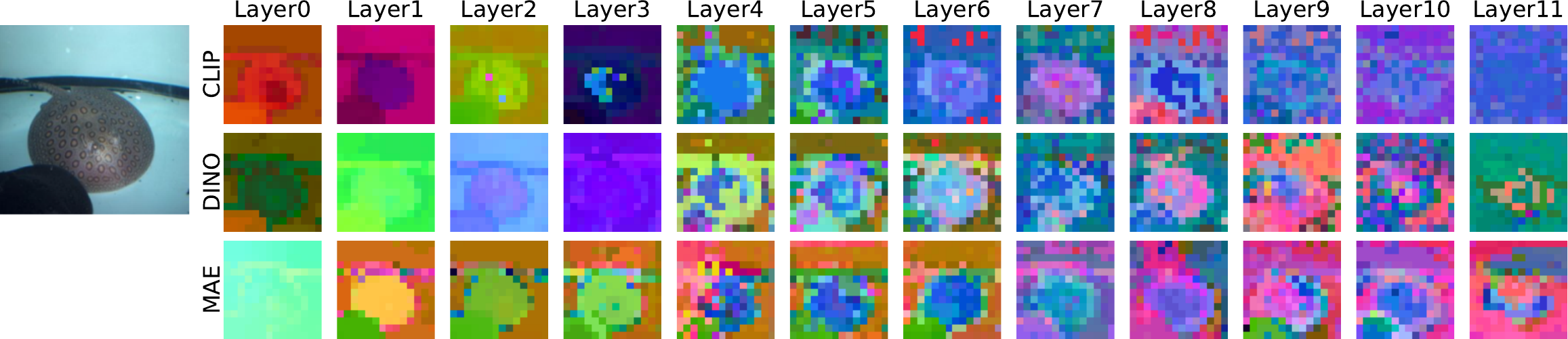}
    \includegraphics[width=\linewidth]{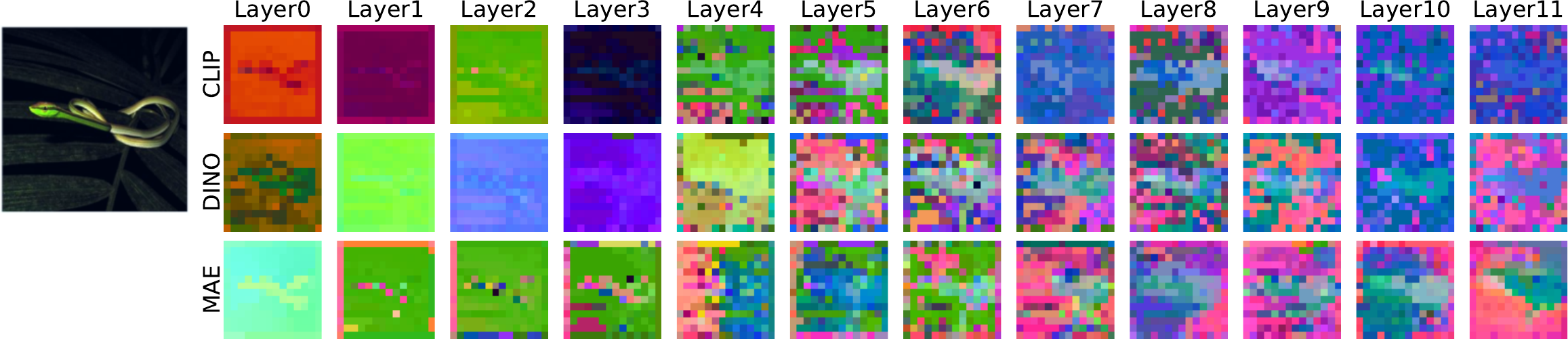}
    \includegraphics[width=\linewidth]{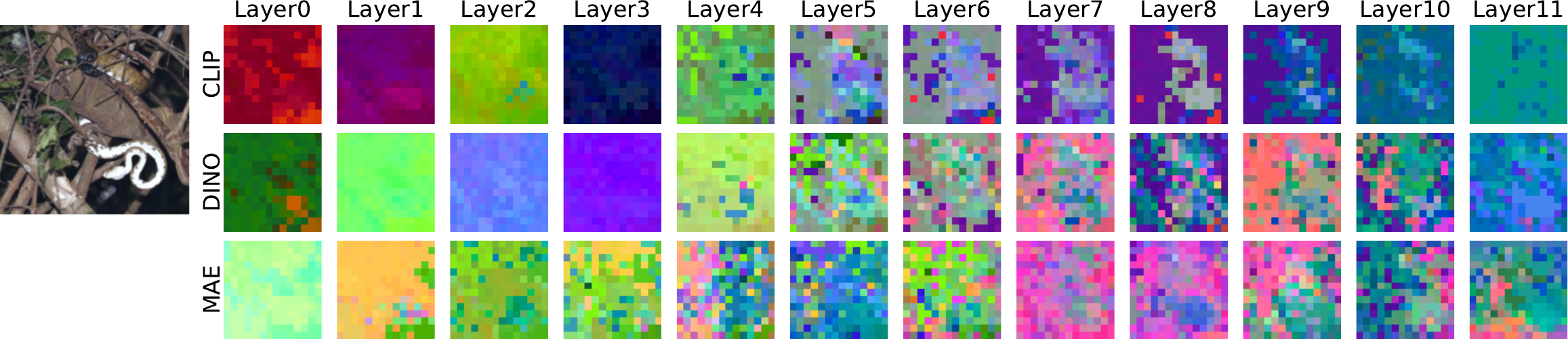}
    \caption{Spectral clustering in the universal channel aligned feature space. The image pixels are colored by our approach AlignedCut, the pixel RGB value is assigned by the 3D spectral-tSNE of the top 20 eigenvectors. The coloring is consistent across all images, layers, and models.}
    \label{fig:eig_tsne_images_app}
\end{figure}

\begin{figure}
    \centering
    \includegraphics[width=\linewidth]{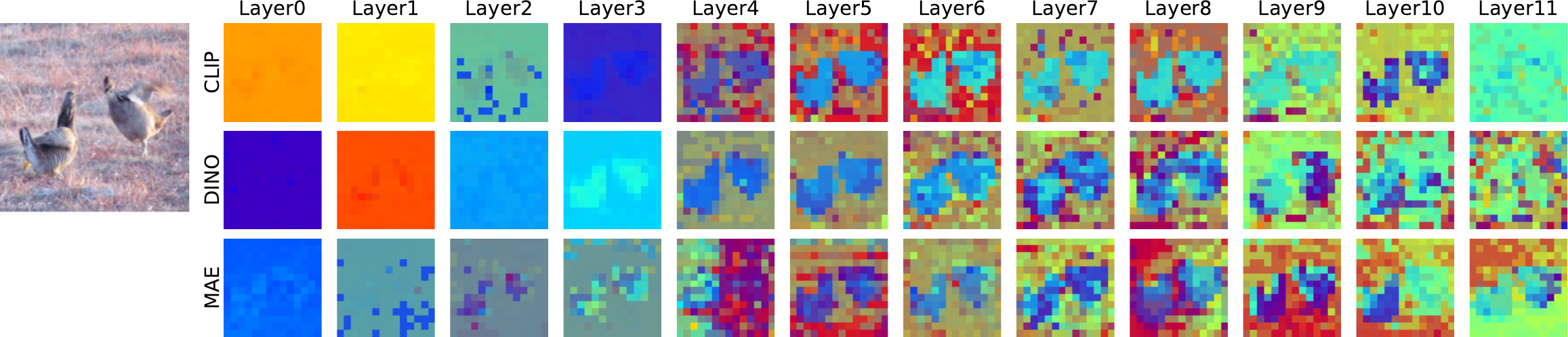}
    \includegraphics[width=\linewidth]{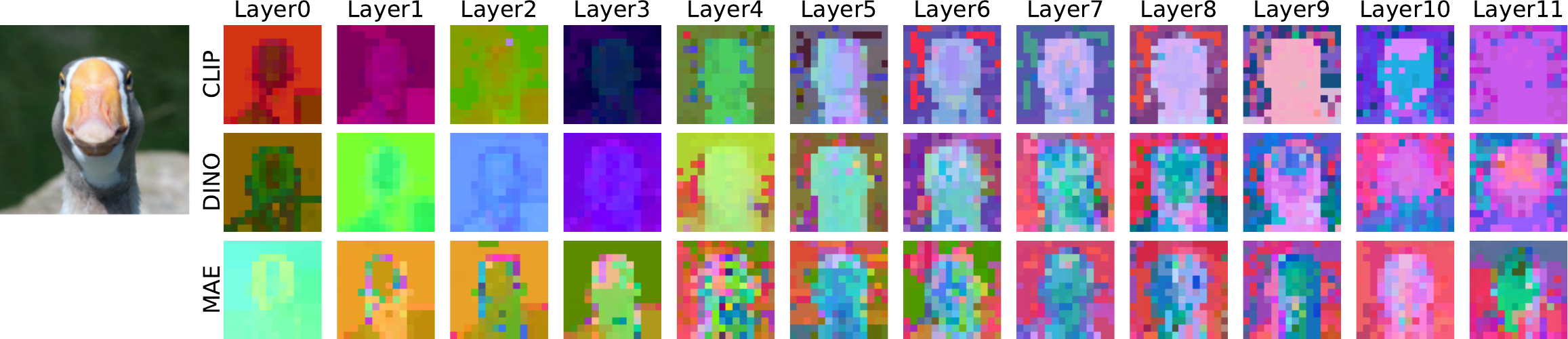}    \includegraphics[width=\linewidth]{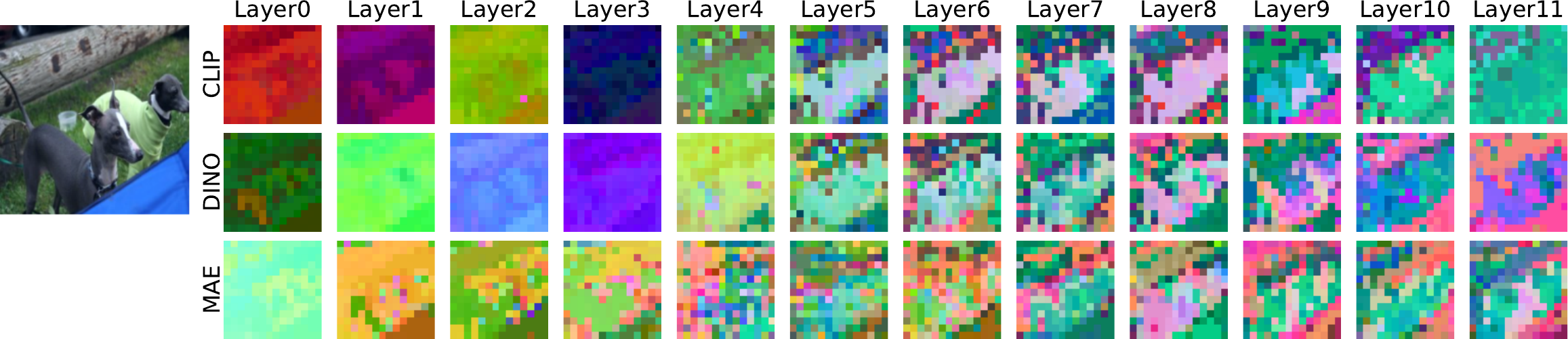}
    \includegraphics[width=\linewidth]{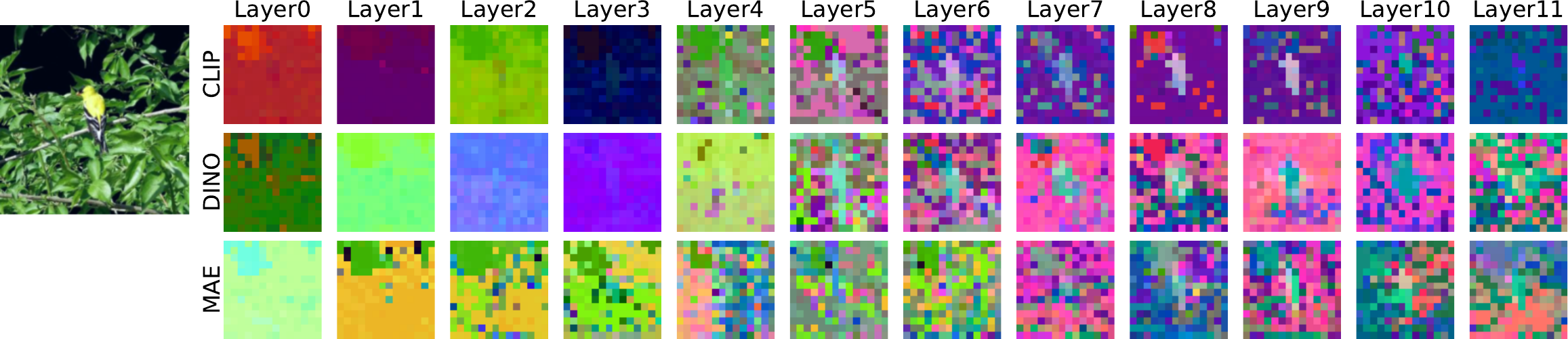}
    \includegraphics[width=\linewidth]{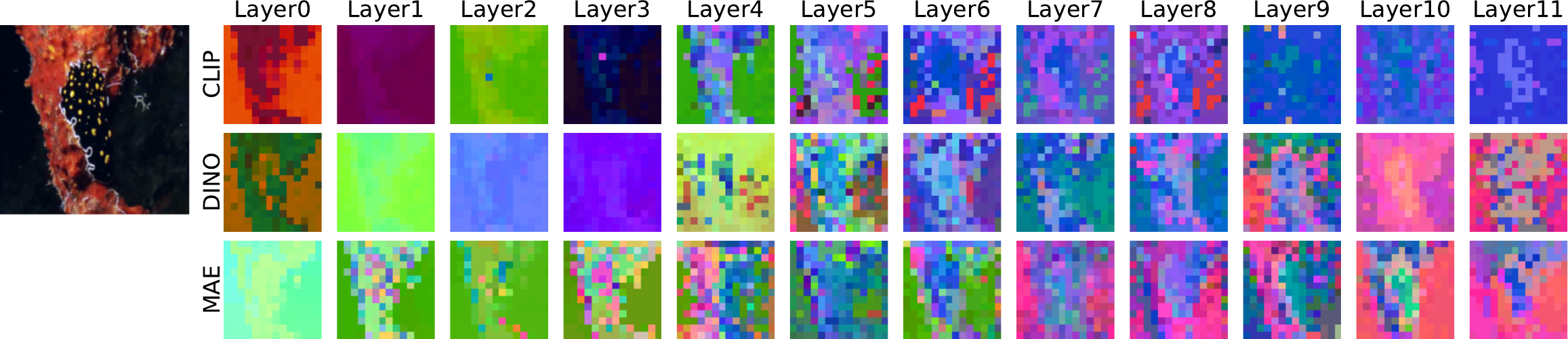}
    \includegraphics[width=\linewidth]{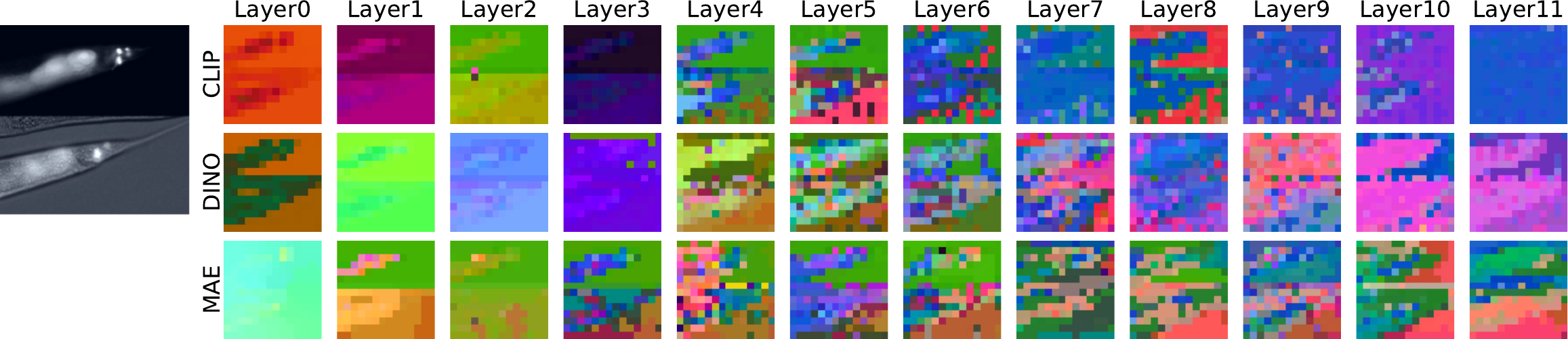}
    \caption{Spectral clustering in the universal channel aligned feature space. The image pixels are colored by our approach AlignedCut, the pixel RGB value is assigned by the 3D spectral-tSNE of the top 20 eigenvectors. The coloring is consistent across all images, layers, and models.}
    \label{fig:eig_tsne_images_app}
\end{figure}

\begin{figure}
    \centering
    \includegraphics[width=\linewidth]{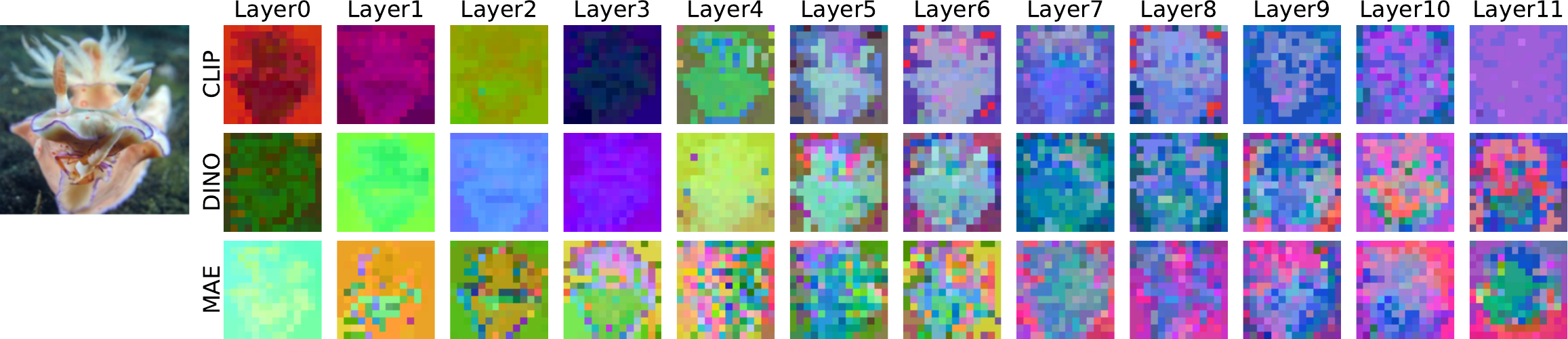}
    \includegraphics[width=\linewidth]{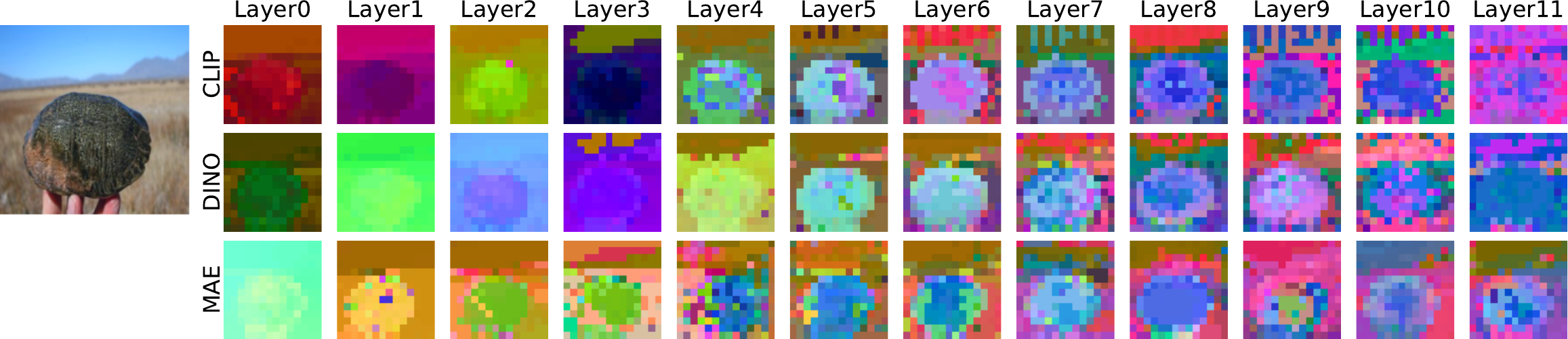}    \includegraphics[width=\linewidth]{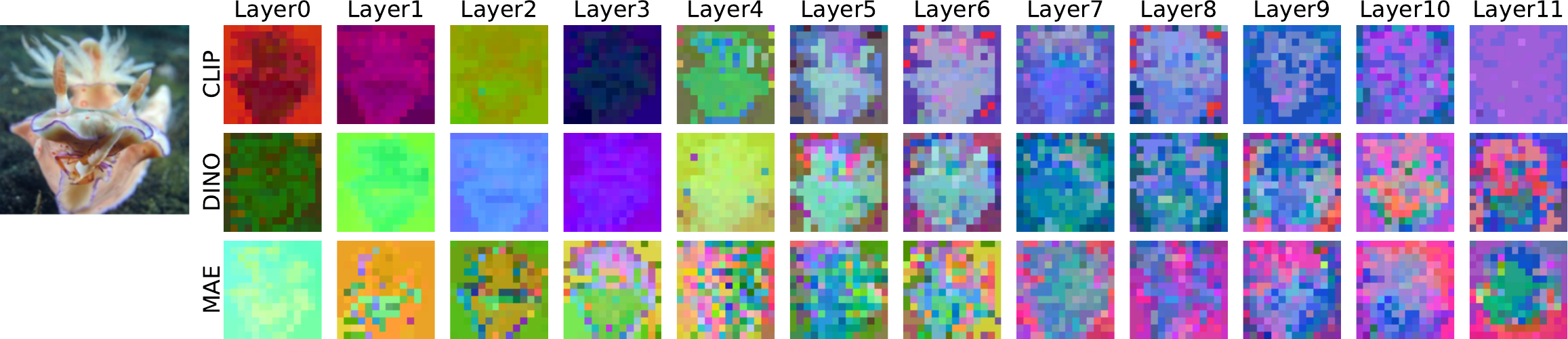}
    \includegraphics[width=\linewidth]{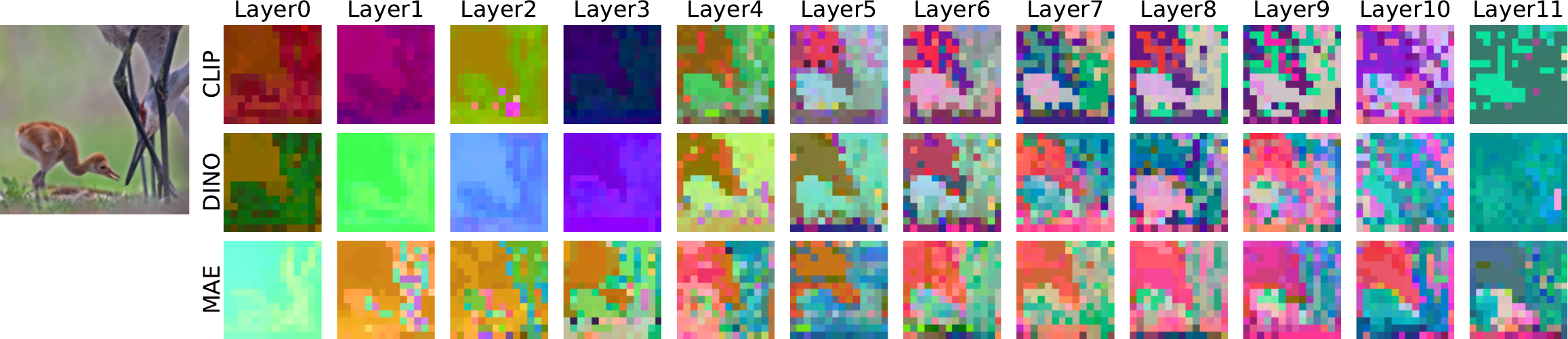}
    \includegraphics[width=\linewidth]{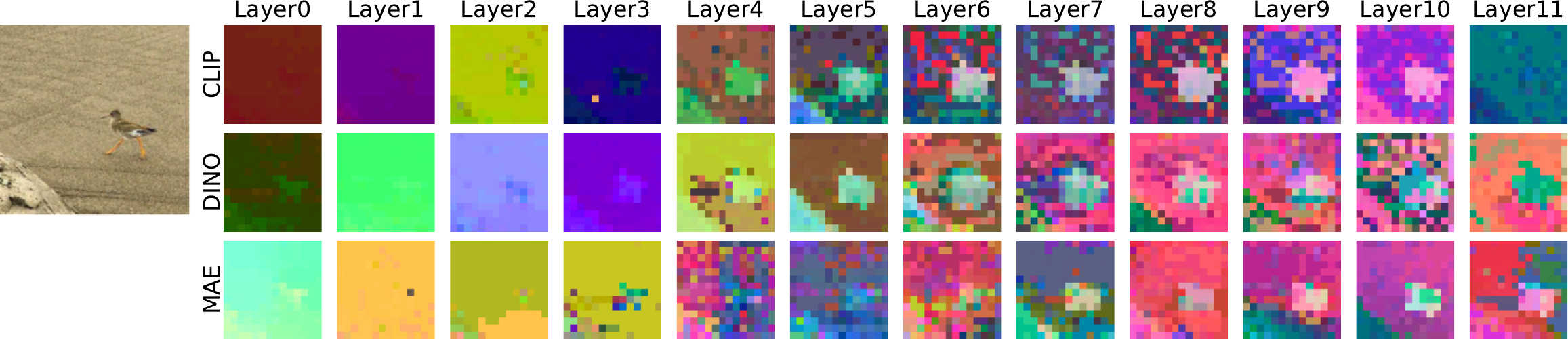}
    \includegraphics[width=\linewidth]{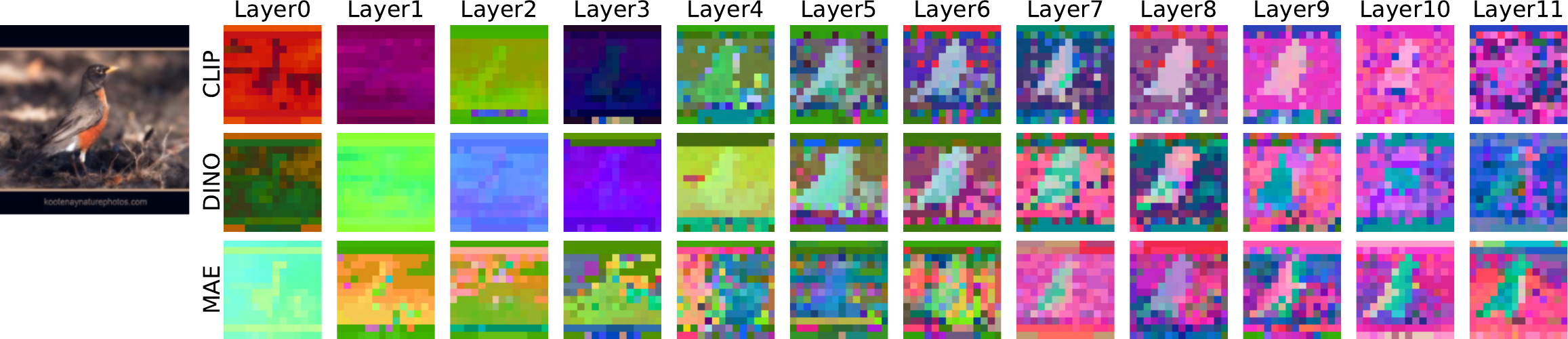}
    \caption{Spectral clustering in the universal channel aligned feature space. The image pixels are colored by our approach AlignedCut, the pixel RGB value is assigned by the 3D spectral-tSNE of the top 20 eigenvectors. The coloring is consistent across all images, layers, and models.}
    \label{fig:eig_tsne_images_app}
\end{figure}

\begin{figure}
    \centering
    \includegraphics[width=\linewidth]{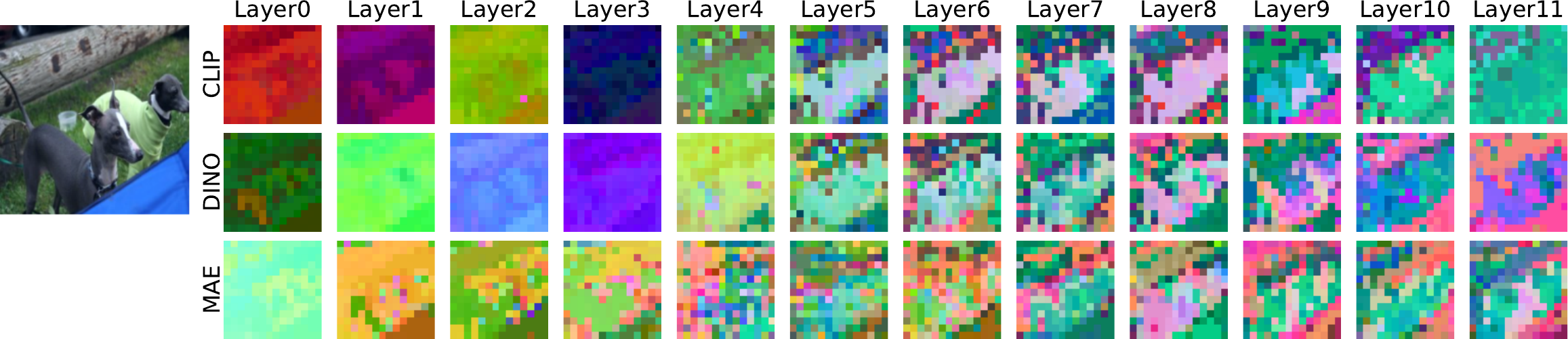}
    \includegraphics[width=\linewidth]{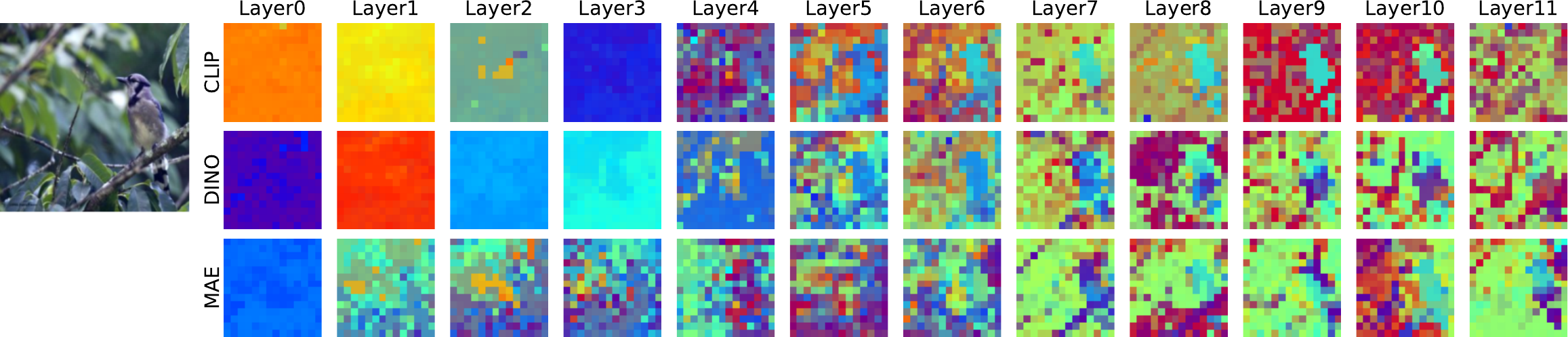}    \includegraphics[width=\linewidth]{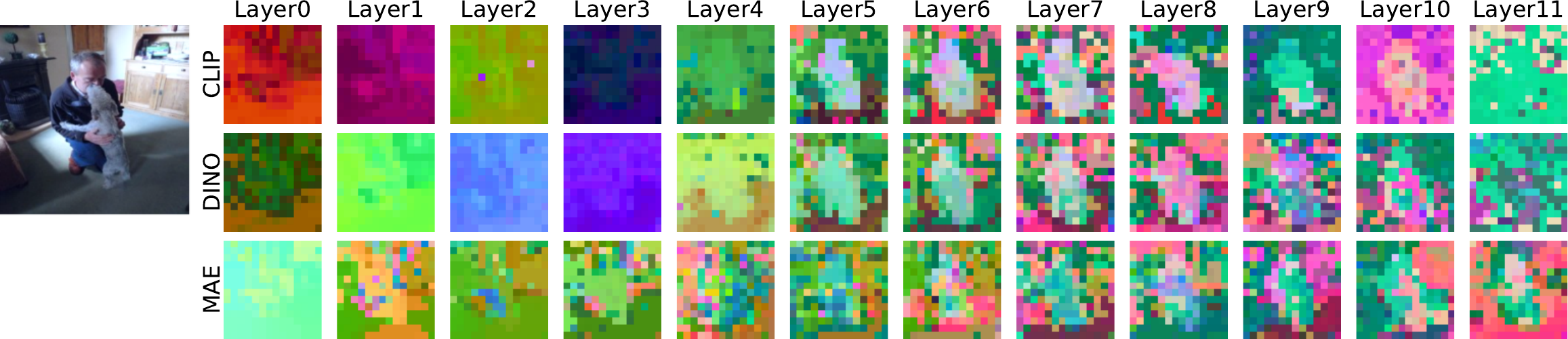}
    \includegraphics[width=\linewidth]{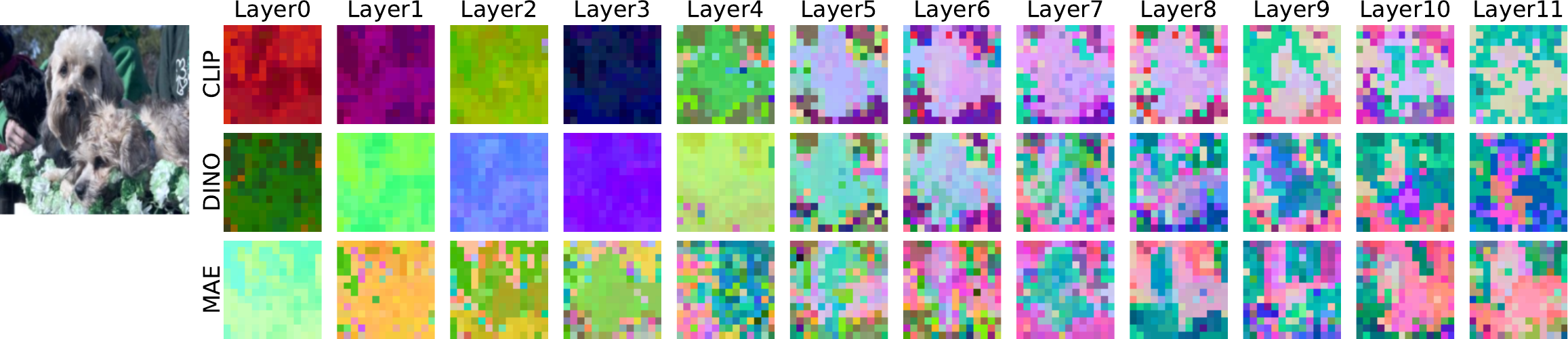}
    \includegraphics[width=\linewidth]{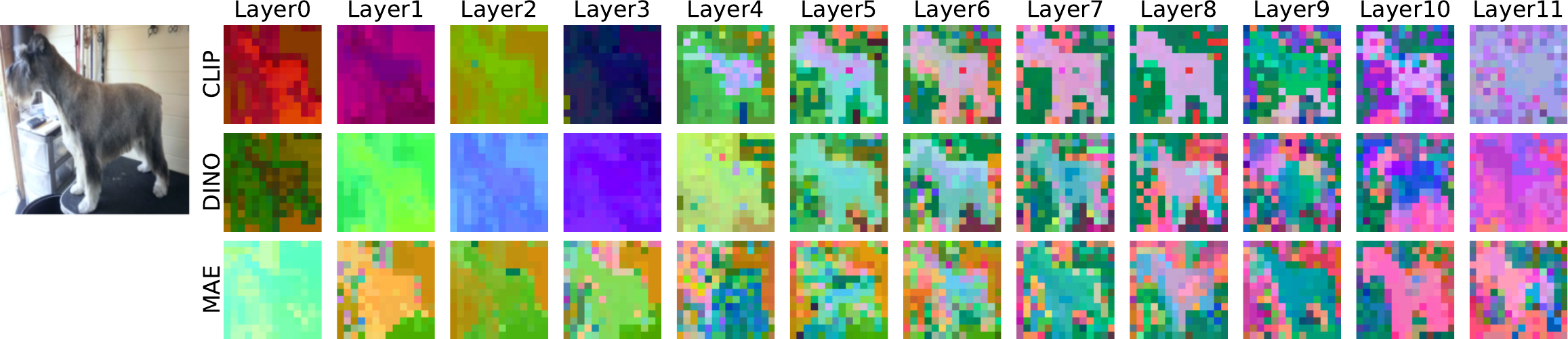}
    \includegraphics[width=\linewidth]{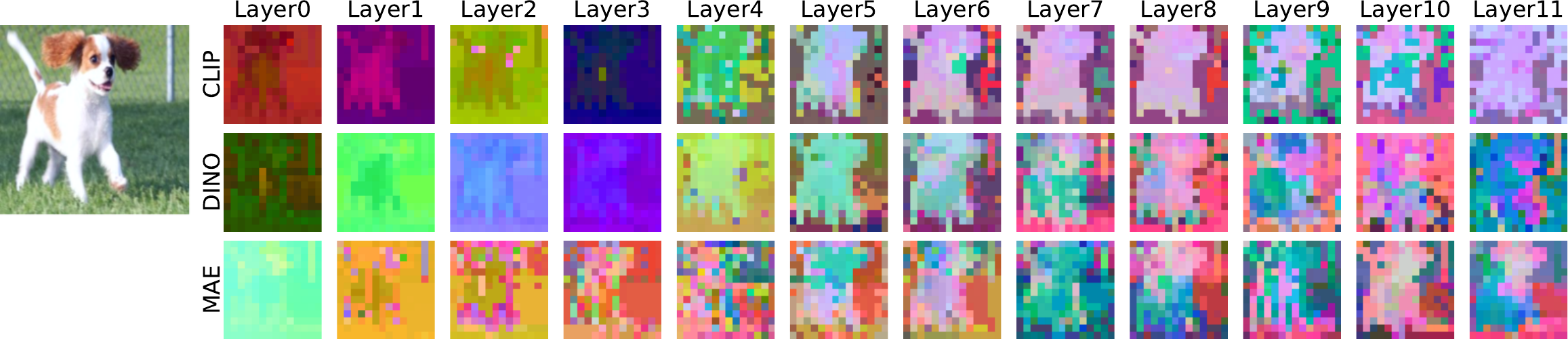}
    \caption{Spectral clustering in the universal channel aligned feature space. The image pixels are colored by our approach AlignedCut, the pixel RGB value is assigned by the 3D spectral-tSNE of the top 20 eigenvectors. The coloring is consistent across all images, layers, and models.}
    \label{fig:eig_tsne_images_app}
\end{figure}

\newpage

\clearpage
\pagestyle{fancy}
\fancyhead{}
\fancyhead[RO,LE]{\textbf{Figure-ground Channel Activation}}

\section{Figure-ground Channel Activation from All Layers and Models}
\label{sec:fgbg}

\begin{figure}[h]
    \centering
    \includegraphics[width=\linewidth]{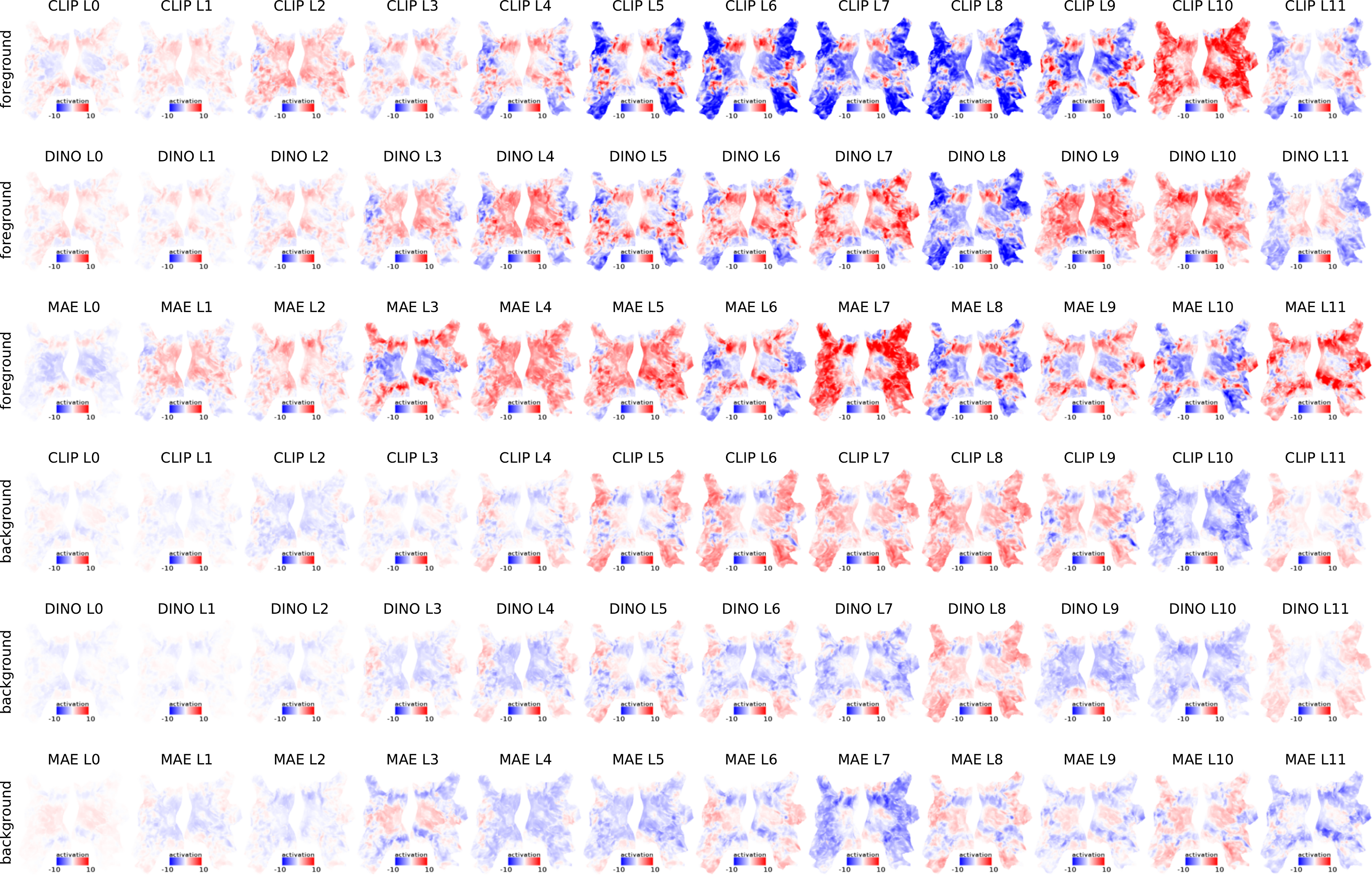}
    \caption{Mean activation of foreground or background pixels at each layer of CLIP, DINO and MAE. Channel activations are linearly transformed to the brain's space. Large absolute activation value means more consistent visual concepts.}
    \label{fig:enter-label}
\end{figure}

\newpage

\clearpage
\pagestyle{fancy}
\fancyhead{}
\fancyhead[RO,LE]{\textbf{Visual Concepts: Category-specific}}

\section{Visual Concepts: Categories }
\label{sec:cate}

\begin{figure}[h]
    \centering
    \includegraphics[width=\linewidth]{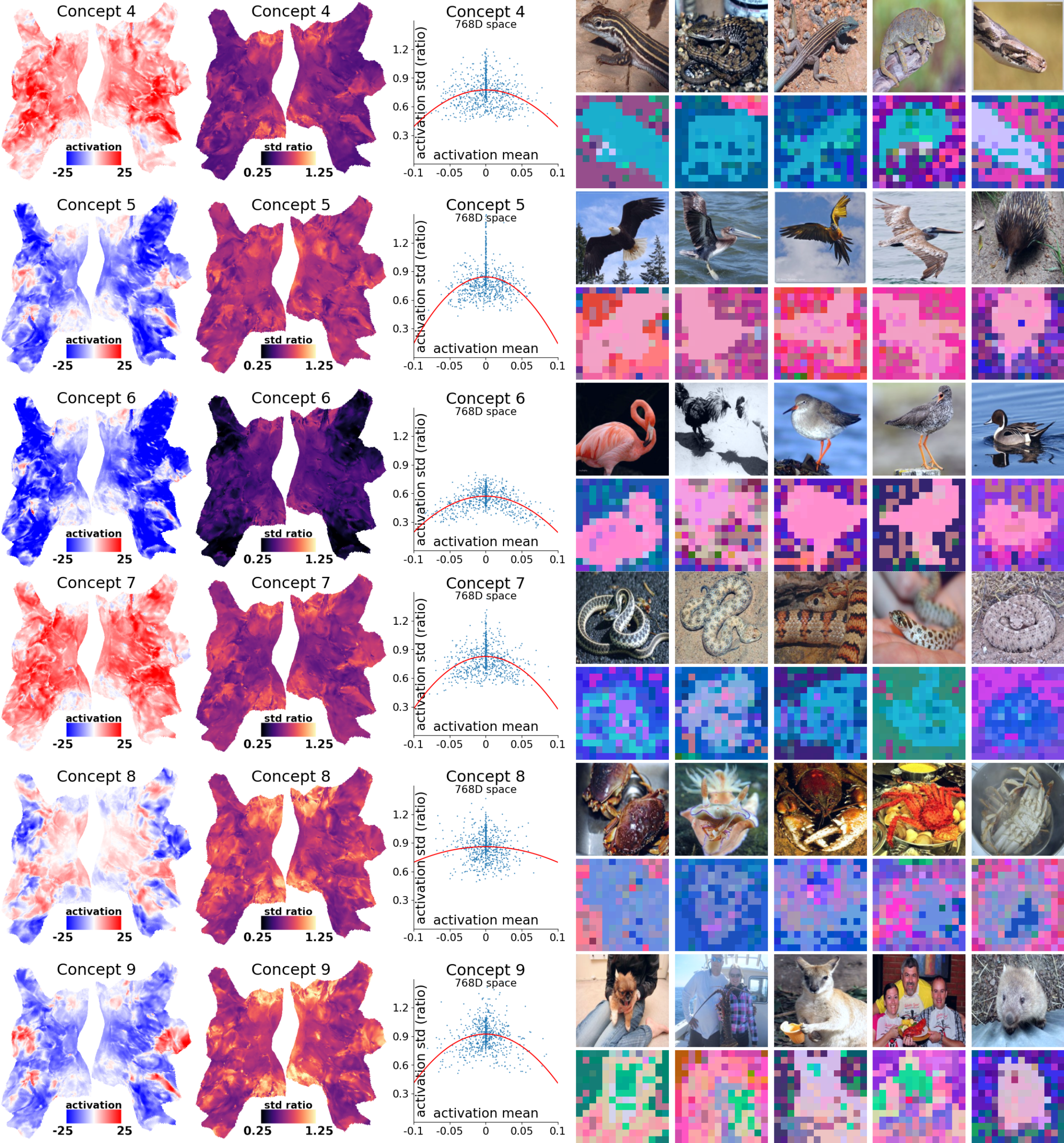}
    \caption{ Category visual concepts in CLIP Layer 9. \textbf{\textit{Left:}} Mean activation of all pixels within an Euclidean sphere centered at the visual concept in the 3D spectral-tSNE space; the concepts activate different brain regions. \textbf{\textit{Middle:}} The standard deviation negatively correlates with absolute mean activations. \textbf{\textit{Right:}} Spectral clustering, colored by 3D spectral-tSNE of the top 20 eigenvectors.
    }
    \label{fig:enter-label}
\end{figure}

\newpage

\clearpage
\pagestyle{fancy}
\fancyhead{}
\fancyhead[RO,LE]{\textbf{2D Spectral-tSNE Space Information Flow}}

\section{Layer-to-Layer Feature Computation Flow in 2D spectral-tSNE space}
\label{sec:2d}

\begin{figure}[h]
    \centering
    \includegraphics[width=\linewidth]{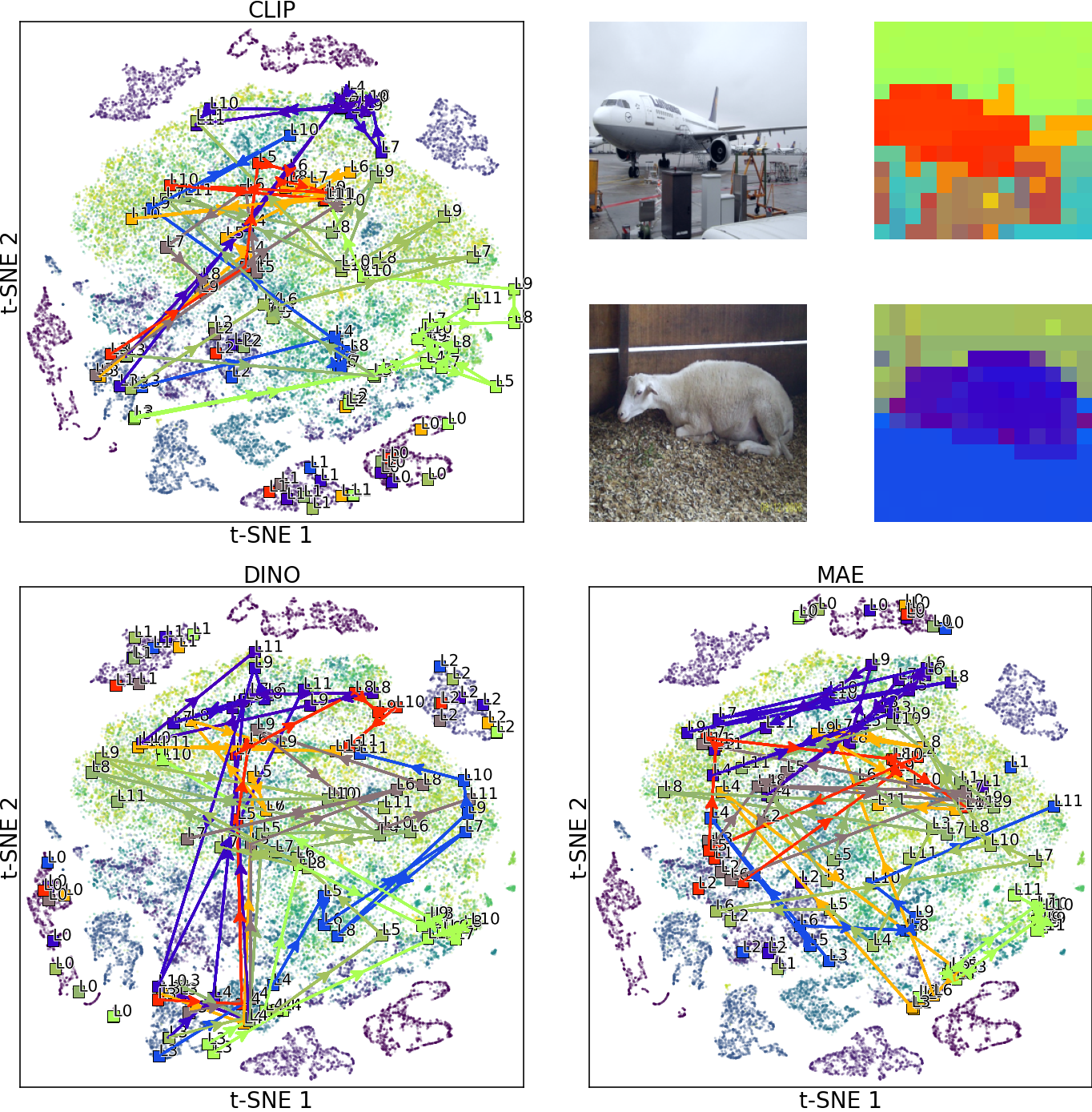}
    \caption{Trajectory of feature progression in from layer to layer, in the 2D spectral-tSNE space. Arrows displayed for 10 randomly sampled example pixels. \textbf{\textit{Top Right:}} Pixels are colored by unsupervised segmentation.}
    \label{fig:enter-label}
\end{figure}

\begin{figure}[t]
    \centering
    \includegraphics[width=\linewidth]{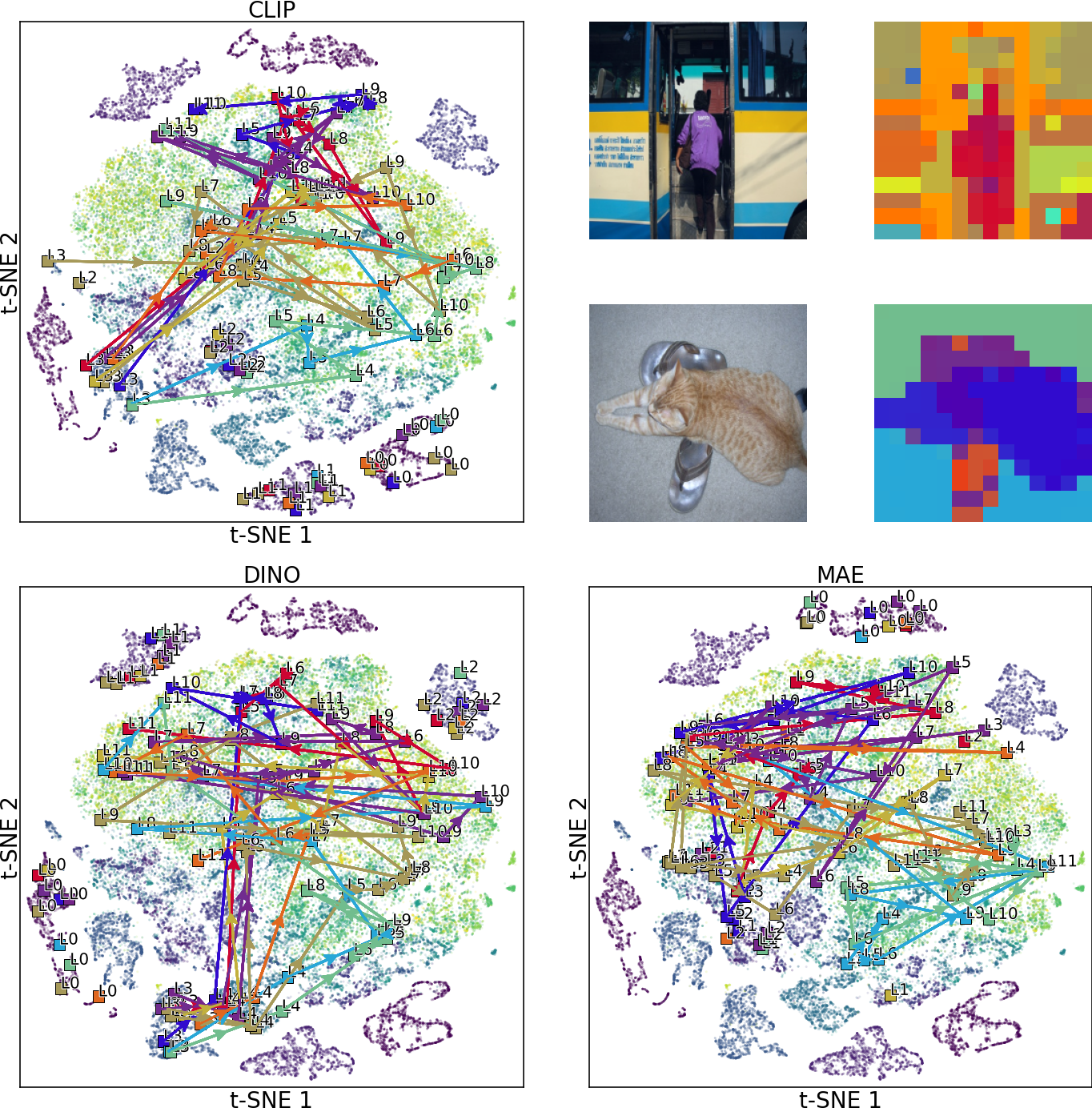}
    \caption{Trajectory of feature progression in from layer to layer, in the 2D spectral-tSNE space. Arrows displayed for 10 randomly sampled example pixels. \textbf{\textit{Top Right:}} Pixels are colored by unsupervised segmentation.}
    \label{fig:enter-label}
\end{figure}

\begin{figure}[t]
    \centering
    \includegraphics[width=\linewidth]{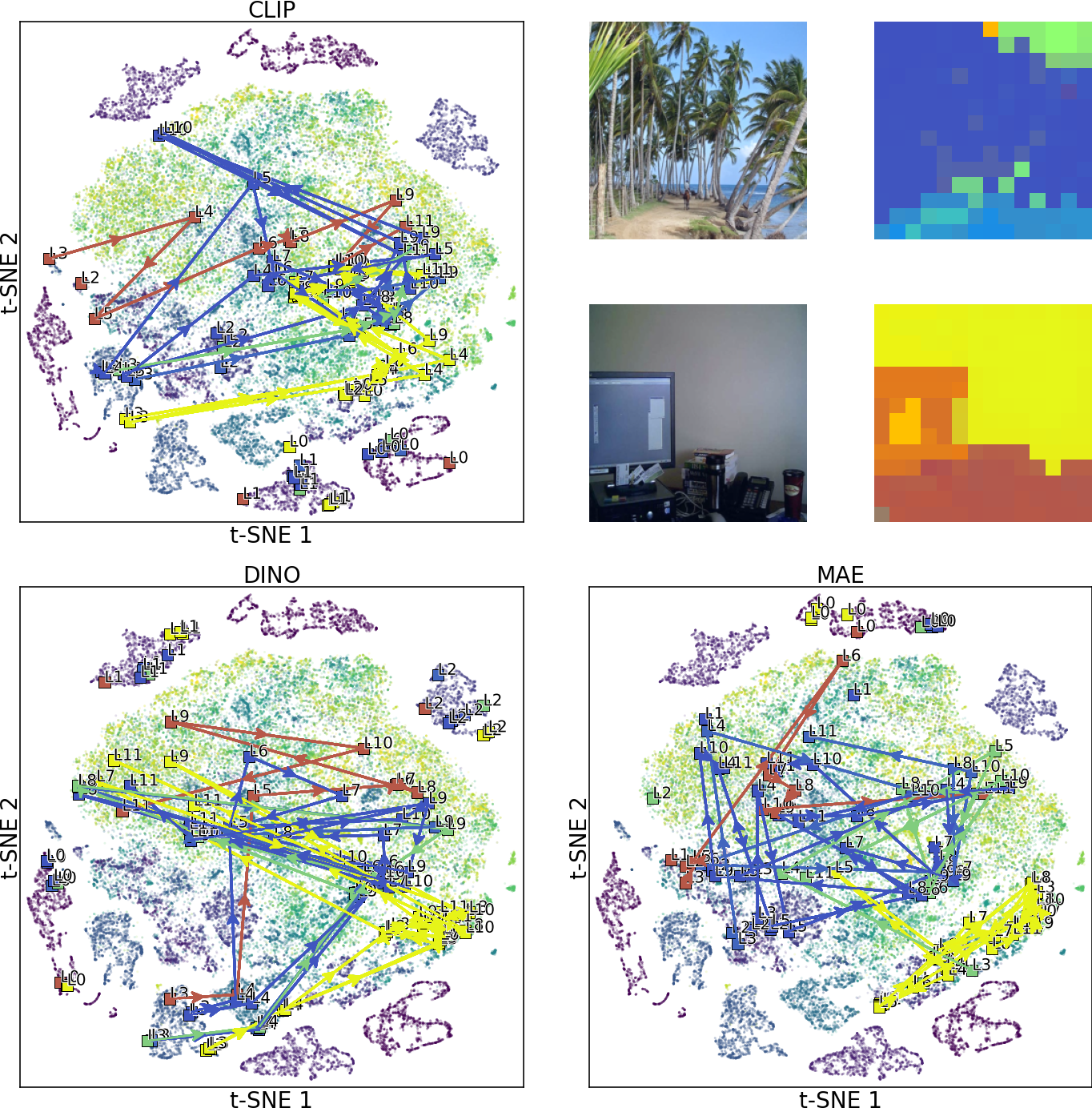}
    \caption{Trajectory of feature progression in from layer to layer, in the 2D spectral-tSNE space. Arrows displayed for 10 randomly sampled example pixels. \textbf{\textit{Top Right:}} Pixels are colored by unsupervised segmentation.}
    \label{fig:enter-label}
\end{figure}

\begin{figure}[t]
    \centering
    \includegraphics[width=\linewidth]{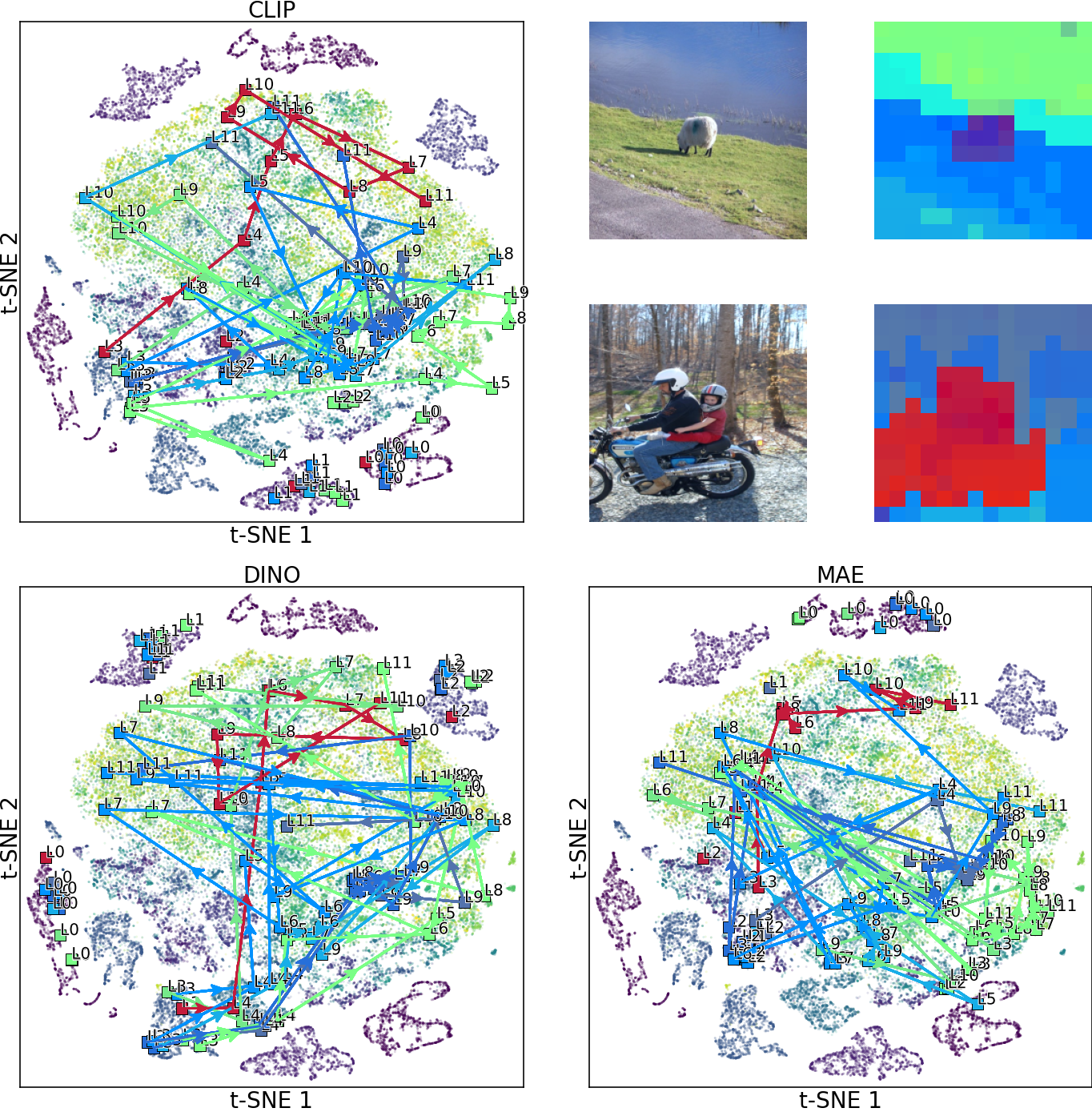}
    \caption{Trajectory of feature progression in from layer to layer, in the 2D spectral-tSNE space. Arrows displayed for 10 randomly sampled example pixels. \textbf{\textit{Top Right:}} Pixels are colored by unsupervised segmentation.}
    \label{fig:enter-label}
\end{figure}

\begin{figure}[t]
    \centering
    \includegraphics[width=\linewidth]{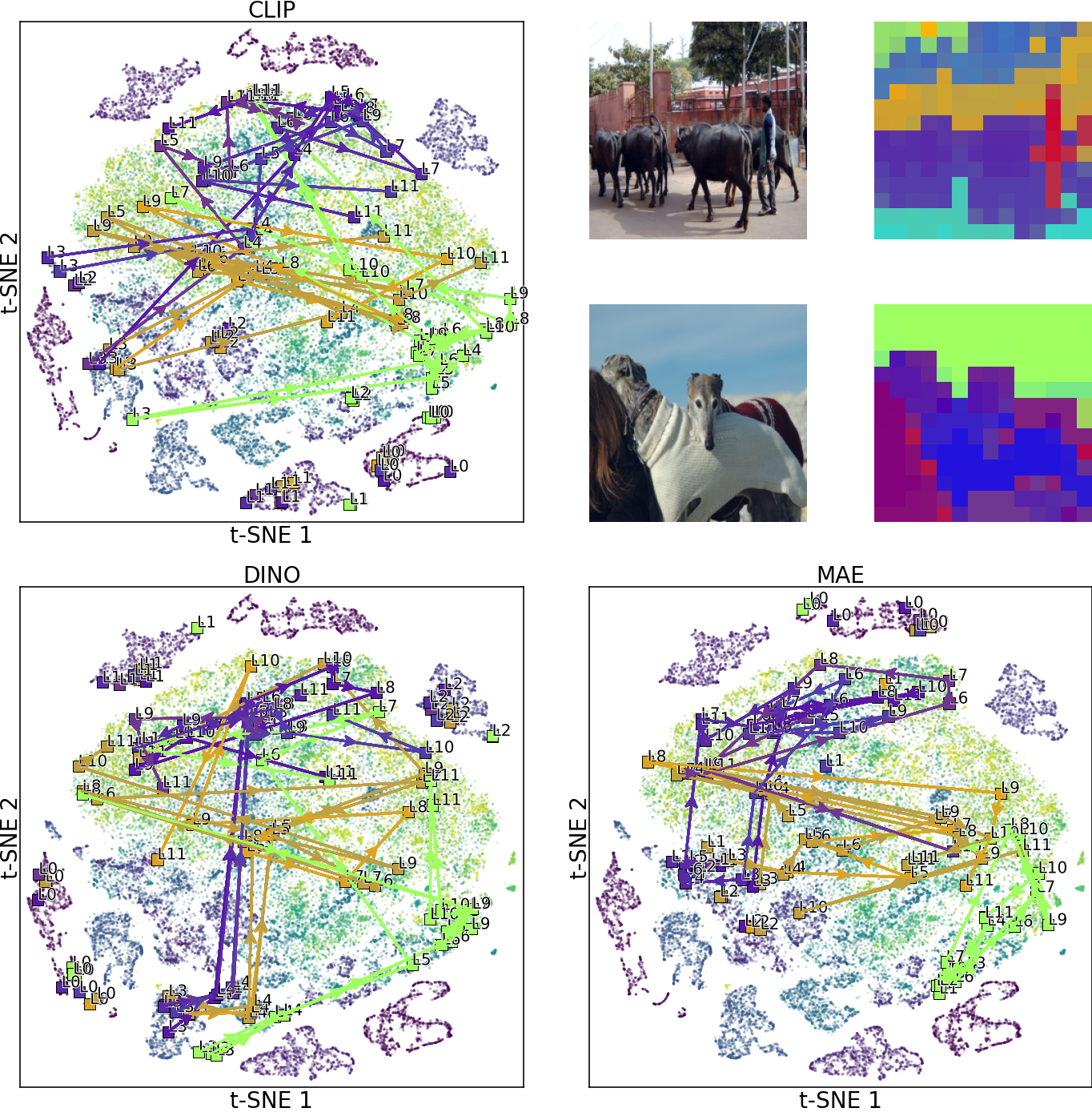}
    \caption{Trajectory of feature progression in from layer to layer, in the 2D spectral-tSNE space. Arrows displayed for 10 randomly sampled example pixels. \textbf{\textit{Top Right:}} Pixels are colored by unsupervised segmentation.}
    \label{fig:enter-label}
\end{figure}

\clearpage
\newpage

\pagestyle{plain}

\section*{NeurIPS Paper Checklist}

\begin{enumerate}

\item {\bf Claims}
    \item[] Question: Do the main claims made in the abstract and introduction accurately reflect the paper's contributions and scope?
    \item[] Answer: \answerYes{} 
    \item[] Justification: 

\item {\bf Limitations}
    \item[] Question: Does the paper discuss the limitations of the work performed by the authors?
    \item[] Answer: \answerYes{} 
    \item[] Justification: 

\item {\bf Theory Assumptions and Proofs}
    \item[] Question: For each theoretical result, does the paper provide the full set of assumptions and a complete (and correct) proof?
    \item[] Answer: \answerNA{} 
    \item[] Justification: 

    \item {\bf Experimental Result Reproducibility}
    \item[] Question: Does the paper fully disclose all the information needed to reproduce the main experimental results of the paper to the extent that it affects the main claims and/or conclusions of the paper (regardless of whether the code and data are provided or not)?
    \item[] Answer: \answerYes{} 
    \item[] Justification: Experimental details are in the appendix

\item {\bf Open access to data and code}
    \item[] Question: Does the paper provide open access to the data and code, with sufficient instructions to faithfully reproduce the main experimental results, as described in supplemental material?
    \item[] Answer: \answerNo{} 
    \item[] Justification: The data is provided as open-source from another study; our code is not yet released, it will be released upon publication.

\item {\bf Experimental Setting/Details}
    \item[] Question: Does the paper specify all the training and test details (e.g., data splits, hyperparameters, how they were chosen, type of optimizer, etc.) necessary to understand the results?
    \item[] Answer: \answerYes{} 
    \item[] Justification: Experimental details are in the appendix

\item {\bf Experiment Statistical Significance}
    \item[] Question: Does the paper report error bars suitably and correctly defined or other appropriate information about the statistical significance of the experiments?
    \item[] Answer: \answerYes{} 
    \item[] Justification: We provided standard deviation in Table 1, measured over training with 3 random seed.

\item {\bf Experiments Compute Resources}
    \item[] Question: For each experiment, does the paper provide sufficient information on the computer resources (type of compute workers, memory, time of execution) needed to reproduce the experiments?
    \item[] Answer: \answerYes{} 
    \item[] Justification: 
    
\item {\bf Code Of Ethics}
    \item[] Question: Does the research conducted in the paper conform, in every respect, with the NeurIPS Code of Ethics \url{https://neurips.cc/public/EthicsGuidelines}?
    \item[] Answer: \answerYes{}{} 
    \item[] Justification: 

\item {\bf Broader Impacts}
    \item[] Question: Does the paper discuss both potential positive societal impacts and negative societal impacts of the work performed?
    \item[] Answer: \answerNA{} 
    \item[] Justification: 
    
\item {\bf Safeguards}
    \item[] Question: Does the paper describe safeguards that have been put in place for responsible release of data or models that have a high risk for misuse (e.g., pretrained language models, image generators, or scraped datasets)?
    \item[] Answer: \answerNA{} 
    \item[] Justification: 

\item {\bf Licenses for existing assets}
    \item[] Question: Are the creators or original owners of assets (e.g., code, data, models), used in the paper, properly credited and are the license and terms of use explicitly mentioned and properly respected?
    \item[] Answer: \answerYes{} 
    \item[] Justification: 

\item {\bf New Assets}
    \item[] Question: Are new assets introduced in the paper well documented and is the documentation provided alongside the assets?
    \item[] Answer: \answerNA{} 
    \item[] Justification: 

\item {\bf Crowdsourcing and Research with Human Subjects}
    \item[] Question: For crowdsourcing experiments and research with human subjects, does the paper include the full text of instructions given to participants and screenshots, if applicable, as well as details about compensation (if any)? 
    \item[] Answer: \answerNA{} 
    \item[] Justification: 

\item {\bf Institutional Review Board (IRB) Approvals or Equivalent for Research with Human Subjects}
    \item[] Question: Does the paper describe potential risks incurred by study participants, whether such risks were disclosed to the subjects, and whether Institutional Review Board (IRB) approvals (or an equivalent approval/review based on the requirements of your country or institution) were obtained?
    \item[] Answer: \answerNA{} 
    \item[] Justification: 

\end{enumerate}


